\documentclass[twocolumn]{svjour3}          

\usepackage{cite}
\usepackage{graphicx}
\usepackage{times}
\usepackage{amsmath}
\usepackage{amssymb}
\usepackage{multirow}
\usepackage{lscape}
\usepackage{longtable}
\usepackage{adjustbox}
\usepackage{caption}
\usepackage{enumitem}
\usepackage{array}
\usepackage{colortbl}
\usepackage{hhline, soul}
\usepackage[ruled,vlined,linesnumbered]{algorithm2e}
\usepackage{makecell}
\usepackage{acronym}
\usepackage{cellspace}
\usepackage{booktabs}
\usepackage[table,xcdraw,dvidvipsnames]{xcolor}
\usepackage{nccmath}
\usepackage[edges]{forest}
\usetikzlibrary{arrows.meta, shapes.geometric, calc, shadows, shadows.blur}
\usepackage[pagebackref=true,breaklinks=true,colorlinks,bookmarks=false]{hyperref}

\newcolumntype{P}[1]{>{\raggedright\let\newline\\\arraybackslash\hspace{0pt}}m{#1}}
\newcolumntype{M}[1]{>{\centering\let\newline\\\arraybackslash\hspace{0pt}}m{#1}}
\newcolumntype{L}[1]{>{\raggedleft\let\newline\\\arraybackslash\hspace{0pt}}m{#1}}

\DeclareMathOperator{\sign}{sign}
\smartqed  
\newcommand{\etal}{\textit{et al. }}
\newcommand{\detector}{$\mathcal{D}$}
\newcommand{\hop}{HopSkipJump }
\newcommand{\image}{Tiny-ImageNet}
\definecolor{customgreen}{RGB}{24, 186, 24}

\newacro{ai}[AI]{artificial intelligence} 
\newacro{ml}[ML]{machine learning} 
\newacro{dl}[DL]{deep learning}
\newacro{sgd}[SGD]{stochastic gradient descent}
\newacro{ae}[AE]{adversarial example}
\newacro{aa}[AA]{adversarial attack}
\newacro{sfad}[SFAD]{selective and feature based adversarial detection}
\newacro{dnn}[DNN]{deep neural network}
\newacro{cnn}[CNN]{convolutional neural network}
\newacro{nn}[NN]{neural network}
\newacro{hca}[HCA]{high confidence attack}
\newacro{pgd}[PGD]{projected gradient descent}
\newacro{lid}[LID]{local intrinsic dimensionality}
\newacro{fgsm}[FGSM]{fast gradient sign method}
\newacro{nic}[NIC]{neural-network invariant checking}
\newacro{gan}[GAN]{generative adversarial network}
\newacro{dnr}[DNR]{deep neural rejection}
\newacro{mtl}[MTL]{multi-task learning} 
\newacro{rbf}[RBF]{radial basis function} 
\newacro{svm}[SVM]{support vector machine}
\newacro{fs}[FS]{feature squeezing}
\newacro{ta}[TA]{threshold attack}
\newacro{pa}[PA]{pixel attack}
\newacro{st}[ST]{spatial transformation}
\newacro{cw}[CW]{carlini-wagner}
\newacro{df}[DF]{deepFool}
\newacro{bfg}[L-BFGS]{limited memory broyden-fletcher-goldfarb-shanno}
\newacro{bim}[BIM]{basic iterative method}
\newacro{uap}[UAP]{universal adversarial perturbations}
\newacro{jsm}[JSMA]{jacobian saliency map attack}
\newacro{fa}[FA]{feature adversary attack}
\newacro{atn}[ATN]{adversarial transformation networks}
\newacro{eot}[EOT]{expectation over transformation}
\newacro{bpda}[BPDA]{backward pass differentiable approximation}
\newacro{zoo}[ZOO]{zeroth order optimization}
\newacro{sa}[SA]{square attack}
\newacro{ba}[BA]{boundary attack}
\newacro{upset}[UPSET]{universal perturbations for steering to exact targets}
\newacro{angri}[ANGRI]{antagonistic network for generating rogue images}
\newacro{cppn}[CPPN EA]{compositional pattern-producing network-encoded evolutionary algorithm}
\newacro{ba}[BA]{boundary attack}
\newacro{de}[DE]{differential evolution}
\newacro{ea}[EA]{evolutionary algorithm}
\newacro{bu}[BU]{bayesian uncertainty}
\newacro{nss}[NSS]{natural scene statistics}
\newacro{mscn}[MSCN]{mean subtracted contrast normalized}
\newacro{mmd}[MMD]{maximum mean discrepancy}
\newacro{pca}[PCA]{principal component analysis}
\newacro{kd}[KD]{kernel density}
\newacro{gda}[GDA]{gaussian discriminant analysis}
\newacro{knn}[$k$-NN]{k-nearest neighbor}
\newacro{gmm}[GMM]{Gaussian mixture model}
\newacro{fpr}[FPR]{false positive rate}
\newacro{cam}[Grad-CAM]{gradient-weighted class activation mapping}
\newacro{nlp}[NLP]{natural language processing}

\begin{document}
\title{Adversarial Example Detection for DNN Models: A Review and Experimental Comparison 
}


\author{Ahmed Aldahdooh         \and
        Wassim Hamidouche       \and
        Sid Ahmed Fezza         \and
        Olivier D\'eforges        
}


\institute{- Ahmed Aldahdooh, Wassim Hamidouche and Olivier Deforges are with University of Rennes, INSA Rennes, CNRS, IETR - UMR 6164, F-35000 Rennes, France. \email{firstname.last@insa-rennes.fr}\\
- Sid Ahmed Fezza is with National Institute of Telecommunications and ICT, Oran, Algeria. \email{sfezza@inttic.dz}
}



\date{Received: date / Accepted: date}
\maketitle

\begin{abstract}
\Ac{dl} has shown great success in many human-related tasks, which has led to its adoption in many computer vision  based applications, such as security surveillance systems, autonomous vehicles and healthcare. Such safety-critical applications have to draw their path to success deployment once they have the capability to overcome safety-critical challenges. Among these challenges are the defense against or/and the detection of the \acp{ae}. Adversaries can carefully craft small, often imperceptible, noise called perturbations to be added to the clean image to generate the \ac{ae}. The aim of \ac{ae} is to fool the \ac{dl} model which makes it a potential risk for \ac{dl} applications. Many test-time evasion attacks and countermeasures, i.e., defense or detection methods, are proposed in the literature. Moreover, few reviews and surveys were published and theoretically showed the taxonomy of the threats and the countermeasure methods with little focus in \ac{ae} detection methods. In this paper, we focus on image classification task and attempt to provide a survey for detection methods of test-time evasion attacks on neural network classifiers. A detailed discussion for such methods is provided with experimental results for eight state-of-the-art detectors under different scenarios on four datasets. We also provide potential challenges and future perspectives for this  research direction.
\end{abstract}
\acresetall

\section{Introduction} \label{sec: introduction}
\Ac{ml}, as an \ac{ai} discipline, witnessed a great success in different fields, especially in human-related tasks, such as image classification and segmentation~\cite{krizhevsky2012imagenet,simonyan2014very,ren2015faster,long2015fully}, object detection and tracking~\cite{bertinetto2016fully,danelljan2017eco}, healthcare~\cite{ker2017deep}, translation~\cite{bahdanau2014neural} and speech recognition~\cite{hannun2014deep}. Its high accuracy comes from continuous  development of \ac{ml} models, the availability of data and the increase in computational power. Image classification applications are constantly growing and deployed in medical imaging systems, autonomous cars, and safety-critical applications~\cite{kurakin2016adversarial,evtimov2017robust,gu2017badnets,papernot2017practical,melis2017deep}.

Recently and after the potential success of \acp{cnn}~\cite{o2015introduction} for image classification tasks, the focus of this survey, many \ac{dl} models are developed, such as, for instance, VGG16~\cite{simonyan2014very},\linebreak ResNet~\cite{he2016deep}, InceptionV3 \cite{szegedy2016rethinking} and MobileNet~\cite{howard2017mobilenets}. These models and others achieve high prediction accuracy on different publicly available datasets such as MNIST~\cite{lecun1998gradient}, CIFAR10~\cite{krizhevsky2009learning}, SVHN~\cite{netzer2011reading}, Tiny ImageNet~\cite{yao2015tiny}, and ImageNet~\cite{imagenet_cvpr09}. For other human tasks, many models also exist in the literature, such as R-CNN \cite{girshick2014rich}, Fast R-CNN \cite{girshick2015fast} and YOLO \cite{redmon2016you}, which are models for object detection task, while BERT \cite{devlin2019bert}, XLNet \cite{yang2020xlnet} and ALBERT\cite{lan2020albert} are models for \ac{nlp} tasks.

\begin{figure}[t!]
    \begin{center}
        \includegraphics[height=8cm, keepaspectratio]{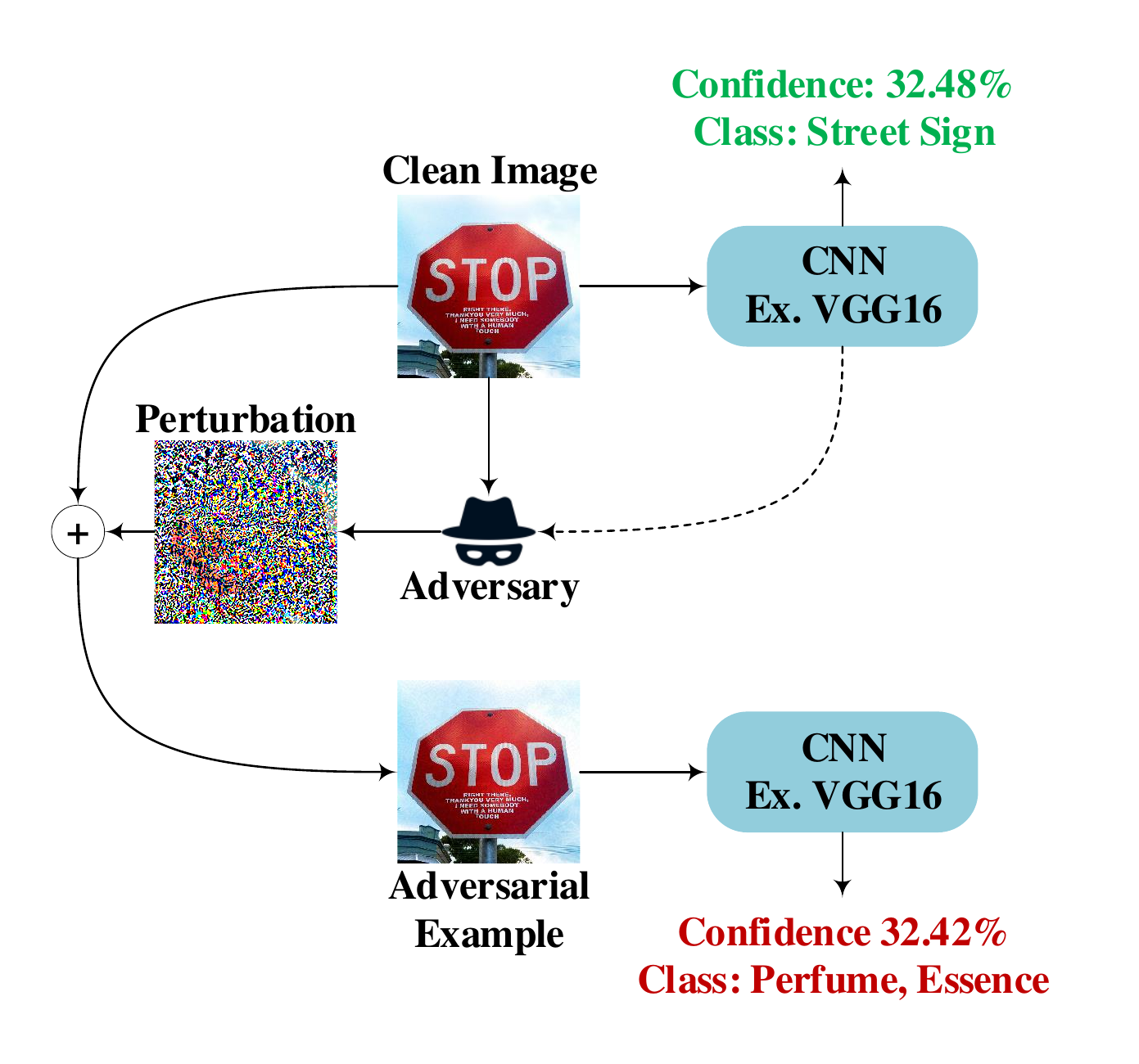}\vspace{-3mm}
    \end{center}
    \caption{The overall objective of the adversary is to fool the \ac{ml} model by adding an imperceptible noise to the original image to generate an adversarial image.} \vspace{-3mm}
    \label{fig:adv}
\end{figure}

This \ac{dl}'s bright face has been challenged by the adversaries. We can categorise the adversary's attack into two broad categories: poisoning and evasion attacks~\cite{PITROPAKIS2019100199}. In the poisoning attack, the adversary is aiming at contaminating the training data that takes place during the training time of the model. Poisoning-based backdoor attack \cite{li2020backdoor} is one of the popular ways to poison the training data. While for evasion attacks, Szegedy \etal~\cite{szegedy2013intriguing} uncovered the potential risk facing \ac{dl} models for image classification. In this paper, we review the detection methods of evasion attacks. It was shown that the adversary, in the testing time, can carefully craft small noise, called perturbation, to be added to the input of the \ac{dl} model to generate the \ac{ae}, as described in Figure~\ref{fig:adv}. The generated \ac{ae} looks perceptually like the original clean image, the perturbation is hardly perceptible for humans, while the \ac{dl} model misclassifies it. The specific objective of the adversary is to: 1) have false predictions for the input samples, 2) get high confidence for the falsely predicted samples, and/or 3) possess transferability property whereby the \acp{ae} that are designed for a specific model can fool other models. The adversarial attacks threat is very challenging since the identification of \ac{ae} and its features are hard to predict~\cite{carlini2017adversarial,ilyas2019adversarial}. 
According to the available information, the adversary can generate \acp{ae} in three different scenarios including, white box, black box and gray box attacks~\cite{akhtar2018threat,hao2020adversarial}. In white box attack scenario, the adversary knows everything about the \ac{dl} model, including model architecture and its weights, and the model inputs and outputs. Specifically, in this setting, the \ac{ae} is generated by solving an optimization problem with the guidance of the model gradients~\cite{kurakin2016adversarial,goodfellow2014explaining,moosavi2016deepfool,carlini2017towards,madry2017towards}. In black box scenario, the adversary has no knowledge about the model. Thus, by using the transferability property~\cite{papernot2016transferability} of \acp{ae} and the input samples content, the adversary can generate harmonious \ac{ae} of the input sample~\cite{chen2017zoo,engstrom2019exploring,su2019one,kotyan2019adversarial}. Finally, in the gray box scenario, the adversary has limited knowledge about the model. He has access to the training data of the model, but does not have any knowledge about the model architecture. Thus, his goal is to substitute the original model with an approximated one, then use its gradient as in white box scenario to generate \acp{ae}.

Adversarial attacks are not limited to image classification tasks, other machine learning tasks' models are also vulnerable to adversarial attacks, such as object detection \cite{xie2017adversarial,lu2017no}, \ac{nlp} \cite{zhang2020adversarial,sun2020advbert,li2019universal}, speech recognition \cite{wang2020adversarial}, physical world \cite{ren2021adversarial}, cybersecurity\cite{dasgupta2019survey} and medical imaging \cite{finlayson2018adversarial}.

Since uncovering this threat to \ac{dl} models, researchers put huge efforts to propose emerging methods to detect or to defend against \acp{ae}. Defense techniques like adversarial training~\cite{goodfellow2014explaining,madry2017towards,xie2020smooth,tramer2017ensemble}, feature denoising~\cite{xie2019feature,borkar2020defending,liao2018defense}, preprocessing~\cite{bakhti2019ddsa,mustafa2019image,prakash2018deflecting} and gradient masking~\cite{papernot2016distillation,papernot2017practical,gu2014towards,nayebi2017biologically} try to make the model robust against the attacks and let the model correctly classify the \acp{ae}. On the other hand, detecting techniques like statistical-based~\cite{grosse2017statistical}, denoiser-based \cite{meng2017magnet}, consistency-based like feature squeezing~\cite{xu2017feature}, \linebreak classification-based~\cite{grosse2016adversarial} and network invariant~\cite{ma2019nic} techniques try to predict/reject the input sample if it is adversarial before being passed to the \ac{dl} model. Besides, the brightening face of this threat is that it makes forward steps in understanding and improving deep learning~\cite{ortiz2020optimism}.

In attempts to highlight the potential challenges and to organize this research direction, few surveys have been published~\cite{akhtar2018threat,hao2020adversarial,yuan2019adversarial,chakraborty2018adversarial}. These reviews are more focusing on theoretical aspects of \acp{aa} and the countermeasures, particularly defense methods, with a lack of focus on adversarial detection techniques. Furthermore, these reviews never showed  experimental comparisons between attacks and its counter-measurements. In this review paper, we give the first insight into the theoretical and experimental aspects of test-time \acp{ae} detection techniques for computer vision image classification tasks. Therefore, our key contributions can be summarized as follows:
\begin{itemize}[noitemsep,topsep=8pt,itemsep=4pt,partopsep=4pt, parsep=4pt]\vspace{-4mm}
    \item We provide a review for state-of-the-art \acp{ae} detection methods and categorize them with respect to the knowledge of adversarial attacks and with respect to the technique that is used to distinguish clean and adversarial inputs.
    \item We provide the first experimental study for state-of-the-art \acp{ae} detection methods that are tested to detect inputs crafted using different \ac{aa} types, i.e., white-, black- and gray-box attacks, on four publicly available datasets, MNIST~\cite{lecun1998gradient}, CIFAR10~\cite{krizhevsky2009learning}, SVHN~\cite{netzer2011reading}, \linebreak and \image~\cite{yao2015tiny}. The summary of the experiments is shown in Figure \ref{fig:summary}. 
    \item We provide a detailed discussion on \acp{ae} from the point of view of their content and their impact on the detection methods.
    \item We publicly release the testing framework that can be used to reproduce the results. The framework is scalable and new detection methods can easily be included. Moreover, a benchmark website is released\footnote{Benchmark website: \url{https://aldahdooh.github.io/detectors_review/}} to promote researchers to contribute and to publish the results of their detectors against different types of the attacks. 
\end{itemize}

The rest of the paper is structured as follows. Related work is discussed in Section \ref{sec: related_work}. In  Section~\ref{sec: background}, a brief review of notations, definitions, threat models, adversarial attack algorithms and defense models are presented. Section~\ref{sec: detection_methods} is dedicated to discuss the \acp{ae} detection methods in detail. Then, we present the comparison experiments and discuss in detail the results in Section~\ref{sec: exp_setting} and Section~\ref{sec: results}, respectively. Finally, we conclude with challenges and future perspectives of this research direction in Section~\ref{sec: challenges}. 

\vspace{-3mm}
\section{Related work} \label{sec: related_work}
Different approaches to generate \acp{ae} in addition to countermeasures to deal with them have been proposed. Akhtar \etal~\cite{akhtar2018threat} published the first review covering this research field which includes works done before 2018. They classified the countermeasure methods based on where the modification is applied to the components of the model. They considered three classes as follows: 1) methods that change data, i.e., training or input data, 2) methods that change the model, and 3) methods that depend on add-on networks. Next, Yuan \etal~\cite{yuan2019adversarial} and Wang \etal~\cite{wang2019security} reviewed the defense methods and categorized them with respect to the type of action against the \acp{ae} as follows: 1) reactive methods that deal with \acp{ae} after building the \ac{dl} model and \hbox{2) proactive} methods that make \ac{dl} models robust before generating the \acp{ae}. In their work, the authors classify the detection methods as reactive. Chakraborty \etal~\cite{chakraborty2018adversarial} extensively discussed the types of attacks from different points of view, but briefly discussed the defense methods including the detection methods. In~\cite{hao2020adversarial}, the authors were the first who classified the detection methods into: 1) auxiliary models in which a subnetwork or a separate network acts as classifier to predict adversarial inputs, 2) statistical models in which statistical analyses were used to distinguish between normal and adversarial inputs and 3) prediction consistency based models that depends on the model prediction if the input or the model parameters are changed. The review of Machado \etal \cite{machado2020adversarial} treated auxiliary detection models as one taxonomy of defenses. In the review of Bulusu \etal~\cite{bulusu2020anomalous} and of Miller \etal~\cite{miller2019not,miller2020adversarial}, they classified detection methods with respect to the presence of \acp{ae} in the training process of the detector into: 1) supervised detection in which \acp{ae} are used in the training of the detector and 2) unsupervised detection in which the detector is only trained using normal training data. Serban \etal \cite{serban2020adversarial} introduced a new taxonomy of the defenses. The first category is called \textit{Guards} and the second category is called \textit{defense by design}. In the former, where the detection methods are categorised in, the defense method does not interact with the under attack and only builds precautions around it, while in the latter category, the defense acts directly on the model architecture and the training data. Carlini \etal~\cite{carlini2017adversarial} did an experimental study on ten detectors to show that all tested detectors can be broken by building new loss functions, but the work in \cite{carlini2017adversarial} did not compare the detectors' performance.

The aforementioned reviews did a great job, but detection methods are not classified and discussed in more detail, besides they are not compared to each other experimentally. In addition, considerable new detection methods have been released recently. 

\section{Adversarial attacks and defense methods} \label{sec: background}
In this section, we briefly introduce the basic concepts of \acp{aa} that target \ac{dl} models of computer vision. Firstly,  the notations that are used in the literature for \ac{dl} models and \acp{aa} are provided. Secondly, the threat models that \ac{dl} models face are presented. Finally, the state-of-the-art attacks and defenses methods are described, excluding the adversarial detection methods that will be discussed in more detail in Section~\ref{sec: detection_methods}.

\subsection{Notations and definitions} \label{sec: notation}
The \ac{dnn} and \acf{cnn} are basically a fitting function $f$ that uses its neural nodes' interconnections to extract features from labeled raw data. Let $\mathcal{X}$ be an input space, e.g., images, and $\mathcal{Y}$ a label space, e.g., classification labels. Let $\mathbb{P}(X,Y)$ be the data distribution over $\mathcal{X} \times \mathcal{Y}$. A model $f:X \rightarrow Y$, is called a prediction function. The \acp{nn} are trained, typically, using \ac{sgd} algorithm that uses the backpropagation of the error to update the model weights $\theta$. To calculate this error/loss, a loss function $ \ell : Y \times Y \rightarrow \mathbb{R}^2$ is defined. The objective for a labeled set $S_m = {(x_i, y_i)}_{i=1}^{m}  \subseteq (\mathcal{X} \times \mathcal{Y})^m$ sampled i.i.d. from $\mathbb{P}(X,Y)$, where $m$ is the number of training samples, is to reduce the empirical risk of the prediction function $f$ is $\hat{r}(f\mid S_m) \triangleq \frac{1}{m} \sum_{i=1}^{m} \ell (f(x_i),y_i)$.

The goal of the adversary, then, given the model $f$ and input sample ($x,y$) is to find an adversarial input $x^{\prime}$, such that $||x^{\prime}-x|| < \epsilon$ and $f(x) \neq f(x^{\prime})$, where $\epsilon$ is the maximum allowed perturbation and $\epsilon\in\mathcal{R}^n $. \\

\noindent Table \ref{tab:notations} and Table \ref{tab:defeinition} list respectively notations and definitions used in the literature and used in adversarial detection methods.
\begin{table}[t!]
    \centering
    \caption{List of notations.}
    \label{tab:notations}
    \begin{tabular}{|M{0.27\columnwidth}|P{0.63\columnwidth}|}
    \hline
         Notation &  Description\\ \hline \hline
         $x$& Clean input image \\ \hline
         $y$& Clean input label \\ \hline
         $x^{\prime}$& Adversarial example of $x$  \\ \hline
         $y^{\prime}$& Adversarial input label \\ \hline
         $t$& Target label of adversarial attack\\ \hline
         $f(.)$& Prediction function. It returns the probability of each class as $f(x)=softmax(z(x))$ \\ \hline
         $\theta$& Model $f$ parameters/weights \\ \hline
         $\ell(,)$& Loss function \\ \hline
         $\delta = x^{\prime}-x$& Perturbation, noise added to clean sample, or the difference between adversarial and clean samples \\ \hline
         $||\delta||$ & The similarity (distance) between $x$ and $x^{\prime}$ \\ \hline
         $\epsilon$& The maximum allowed perturbation \\ \hline
         $\nabla$& Model $f$ gradient \\ \hline
         $z$& Logits, output of the layer before $softmax$\\ \hline
         $\sigma$& Activation function \\ \hline
         $||.||_p$&  $\ell_p$-norm\\ \hline
    \end{tabular}
\end{table}

\subsection{Threat models} \label{sec: threatmodel}
Threat model refers to conditions under which the \acp{ae} are generated. It can be categorized according to many factors. In literature~\cite{akhtar2018threat,hao2020adversarial,yuan2019adversarial,chakraborty2018adversarial}, adversary knowledge, adversary goal, adversary capabilities, attack frequency, adversarial falsification, adversarial specificity and attack surface are the identified factors. We focus here in threat model that is identified by adversary knowledge and adversarial specificity:

\begin{enumerate}[leftmargin=*]
    \item Adversary knowledge
        \begin{enumerate}[label=\alph*., leftmargin=*]
            \item Adversary knowledge of baseline \ac{dl} model:
                \begin{itemize}[left=0pt]
                    \item White box attacks: the adversary knows everything about the victim model: training data, outputs and model architecture and weights. The adversary takes advantage of model information, especially the gradients, to generate the \ac{ae}.
                    \item Black box attacks: the adversary doesn't have access to the victim model configurations. He takes advantage of information acquired by querying and monitoring inputs and outputs of the victim model. 
                    \item Gray box attacks: the adversary has knowledge about training data but not the model architecture. Thus, he relies on the transferability property of the \ac{ae} and  builds a substitute model that does the same task of the victim model to generate \ac{ae}. Gray box attacks are also known as semi-white box attacks. 
                \end{itemize}
            \item Adversary knowledge of detection method~\cite{biggio2013evasion}:
                \begin{itemize}[left=0pt]
                    \item No/Zero knowledge adversary: the adversary only knows the victim model and doesn't know the detection technique, and he generates \ac{ae} using the victim model.
                    \item Perfect knowledge adversary: the adversary knows that the victim model has been secured with a detection technique and he knows the configurations (architecture, training data and detection output) of the detection mode, and uses them to generate \ac{ae}.
                    \item Limited knowledge adversary: the adversary knows the feature representation and the type of the detection technique, but doesn't have access to the detection architecture and the training data. Hence, he estimates the detection function in order to generate the \ac{ae}.  
                \end{itemize}
        \end{enumerate}
        
\begin{table}[t!]
    \centering
    \caption{List of definitions.}
    \label{tab:defeinition}
    \begin{tabular}{|M{0.27\columnwidth}|P{0.63\columnwidth}|}
    \hline
         Definition &  Description\\ \hline \hline
         \textit{Adversary}& Who generates the adversarial examples \\ \hline
         \textit{Threat model}& Conditions, scenario, or environment under which that attack is performed. Such as white box attack  \\ \hline
         \textit{Defense}& Technique to make \ac{dl} robust against attacks \\ \hline
         \textit{Detector}& Technique to predict whether the input is adversarial or not \\ \hline
         \textit{Transferability}& A property of adversarial example that shows the attack ability to fool models that aren't used to generate it \\ \hline
         \textit{White box attacks}& The adversary knows detailed information about the victim \ac{dl} model \\ \hline
         \textit{Black box attacks}& The adversary knows nothing about the victim \ac{dl} model but he can access inputs and outputs of the model\\ \hline
         \textit{Gray box attacks}& The adversary has limited knowledge about the victim \ac{dl} model, i.e., training data \\ \hline
         \textit{Targeted attacks}& Attacks that induce the victim \ac{dl} model to classify the input sample into a specific target label $t$  \\ \hline
         \textit{Untargeted attacks}& Attacks that induce the victim \ac{dl} model to classify the input sample into target label $t$ that is not equal to $y$\\ \hline
    \end{tabular}
    \vspace{-5mm}
\end{table}
    \item Adversarial specificity:
        \begin{enumerate}[label=\alph*., leftmargin=*]
            \item Targeted attacks: the adversary generates the \ac{ae} to misguide the \ac{dl} model to classify the input sample into a specific target label $t$. The adversary generates the \ac{ae} by maximizing the probability of the target label. Targeted attacks' generation is harder than untargeted attacks' generation due to the limited space to redirect the \ac{ae} to a target label $t$. Consequently, the targeted attacks are shown to have higher perturbations than untargeted attacks and have less success rates \cite{carlini2017towards,liu2016delving}.
            \item Untargeted attacks: the adversary generates the \ac{ae} to misguide the \ac{dl} model to classify the input sample into a target label $t$ that is different from the correct label $y$. The adversary generates the \ac{ae} by minimizing the probability of the correct label $y$. The adversary can also conduct the attack by generating multiple targeted attacks and then selects the one with minimum perturbation.
        \end{enumerate}
\end{enumerate}

\subsection{Adversarial attacks} \label{sec: adversarialattack}
Generally speaking, not only neural networks are vulnerable to \acp{aa} but other \ac{ml} models are facing the threat as well. Many attacks and defenses were implemented for \ac{ml} models, readers can refer to~\cite{biggio2014pattern,biggio2014security,biggio2018wild} for more information. In this work, we go through attacks that are targeting the image classification task of neural network \cite{szegedy2013intriguing}. The main categorization of \ac{aa} depends on the adversary's knowledge about the classification model, i.e., categorised by white box and black box attacks.
\subsubsection{White box attacks} \label{sec:wb_attacks}
\textbf{\acs{bfg} Attack~\cite{szegedy2013intriguing}.} \ac{bfg} attack is the first developed attack in the series. It aims at finding minimum perturbation $\delta$ such that $y^{\prime}=t$ and $x^{\prime}$ in the range of input domain. This optimization problem is solved approximately using box-constrained \ac{bfg} algorithm~\cite{liu1989limited} by introducing a loss function $\ell$ as follows:
\begin{ceqn}
    \begin{equation} \label{eq:bfg}
        \operatorname*{argmin}_{\delta} c \, ||\delta|| + \ell(x^{\prime}, t), \text{ such that } x^{\prime}\in [0,1]^{n}
    \end{equation}
\end{ceqn}
where $c$ is a regularisation parameter that we continuously search for to find minimum $\delta$ since neural networks are non-convex networks. Hence, the first term is to find minimum perturbation and the second term is to make sure that the loss value is small between $x^{\prime}$ and the target label $t$.\\\\
\textbf{\acs{fgsm} attack~\cite{goodfellow2014explaining}.} It is the first developed $L_{\infty}$ attack that uses \ac{dl} gradients to generate an \ac{ae}. \ac{fgsm} attack is a one-step gradient update algorithm that finds the perturbation direction, i.e., the sign of gradient, at each pixel of input $x$ that maximizes the loss value of the \ac{dl} model. It is expressed as follows
\begin{ceqn}
    \begin{equation} \label{eq:fgsm}
        x^{\prime} = x + \epsilon \sign(\nabla_{x} \ell(x,y)),  \text{ such that } x^{\prime}\in [0,1]^{n}
    \end{equation}
\end{ceqn}
where $\epsilon$ is a parameter to control the perturbation amount such that $||x^{\prime}-x||_\infty < \epsilon$.\\\\
\textbf{\acs{bim} attack~\cite{kurakin2016adversarial}.} It is the iterative version of the \ac{fgsm} attack. \ac{bim} attack applies \ac{fgsm} attack $k$ times. It is expressed as:
\begin{ceqn}
    \begin{equation}\label{eq:bim}
    \begin{gathered} 
        x_{i+1}^{\prime} = x_{i}^{\prime} + \alpha \sign(\nabla_{x} \ell(x_{i}^{\prime},y)),\\
        \text{ such that } x_{0}^{\prime}=x \text{ ,  } x_{i+1}^{\prime}\in [0,1]^{n} \text{ , and } i=0 \text{ to } k 
    \end{gathered} 
    \end{equation}
\end{ceqn}
where $\alpha$ is the parameter to control the $i^{th}$ iteration step size and it is $0 < \alpha < \epsilon$.\\
\textbf{\acs{pgd} attack~\cite{madry2017towards}.} It is an iterative method similar to \ac{bim} attack. Unlike \ac{bim}, in order to generate ``most-adversarial'' example, i.e., to find local maximum loss value of the model, \ac{pgd} attack starts from a random perturbation in $L_p$-ball around the input sample. Many restarts might be applied in the algorithm. $L_1$, $L_2$, and $L_\infty$ can be used to initialize the perturbation $||x^{\prime}-x||_p < \epsilon$. Recently, a budget-aware step size-free variant of \ac{pgd}~\cite{croce2020reliable} was proposed. Unlike \ac{pgd}, Auto-\ac{pgd} adds a momentum term to the gradient step, adapts the step size across iterations depending on the overall attack budget, and then restarts from the best point. \\\\
\textbf{\acs{cw} attack~\cite{carlini2017towards}.} Carlini and Wagner followed the optimization problem of \ac{bfg} (see \eqref{eq:bfg}) with few changes. Firstly, they replaced the loss function with an objective function
\begin{ceqn}
    \begin{equation} \label{eq:cw_loss}
        g(x^{\prime})=\max(\max_{i\neq t}(Z(x^{\prime})_i) - Z(x^{\prime})_t, -k),
    \end{equation}
\end{ceqn}
where $Z$ is the softmax function and $k$ is the confidence parameter. The authors also provided other six objective functions~\cite{carlini2017towards}. Hence, the optimization problem became like
\begin{ceqn}
    \begin{equation} \label{eq:cw}
    \underset{\delta}{\text{min}} ||\delta|| + c \, g(x^{\prime}), \text{ such that } x^{\prime}\in [0,1]^{n},
    \end{equation}
\end{ceqn}
Thus, minimizing $g$ helps to find $x^{\prime}$ that has a higher score for class $t$. The authors, secondly, converted the optimization from box-constrained to unconstrained problem by introducing $w$ to control the perturbation of the input sample, such that $\delta=\frac{1}{2}(\tanh(w) + 1)-x$. The \ac{cw} attack came with three variants to measure the similarity between $x^{\prime}$ and $x$ relying on $L_0$, $L_2$, and $L_\infty$. This attack is considered as one of the state-of-the-art attacks. It is firstly implemented to break distillation knowledge defense~\cite{papernot2016distillation} and it was shown it is stronger than \ac{fgsm} and \ac{bim} attacks.\\\\
\textbf{\Ac{df} attack~\cite{moosavi2016deepfool}.} Moosavi-Dezfooli \etal introduced an attack that generates smaller perturbation than \ac{fgsm} at the same fooling ratio. Given a binary affine classifier $\mathcal{F} = \{x: f(x)=0\}$, where $f(x)=w^Tx+b$, \ac{df} attack defines the orthogonal projection of $x_0$ onto $\mathcal{F}$ as the minimal perturbation that is needed to change the classifier’s decision, and it is calculated as $\delta_*=-\frac{f(x)}{||w||^2}w$. At each iteration, \ac{df} attack solves the following optimization problem
\begin{ceqn}
    \begin{equation} \label{eq:df}
    \begin{gathered} 
        \operatorname*{argmin}_{\delta_i} ||\delta_i||_2, \\ \text{ such that } f(x_i)+\nabla f(x_i)^T\delta_i =0
    \end{gathered}
    \end{equation}
\end{ceqn}
and these perturbations are accumulated to get the final perturbation.\\\\
\textbf{\acs{uap} attack~\cite{moosavi2017universal}.} It is one of the strongest attacks that generates \textit{image-agnostic universal} perturbation $v$ that can be added to any input sample and fool the \ac{dl} model with up to selected fool rate $fr$. The goal is to find $v$ that satisfies the following two constraints:
\begin{ceqn}
    \begin{equation}\label{eq:uap}
    \begin{gathered} 
        ||v||_p \leq \epsilon \\ \mathcal{P} (f(x + v) \neq f(x)) \geq 1-fr.
    \end{gathered}
    \end{equation}
\end{ceqn}
The authors used \ac{df} attack (see \eqref{eq:df}) to calculate $v$, but any other attack algorithm can be used such as \ac{pgd} or \ac{fgsm}.\\\\
\textbf{Other attacks.} In literature, there are many other attack algorithms. Papernot \etal~\cite{papernot2016limitations} proposed a $L_0$ attack named \ac{jsm}. \ac{jsm} uses \ac{dl} model gradient to calculate a Jacobian based saliency map that ranks the importance of each pixel in the image. Then, \ac{jsm} modifies a few pixels in order to fool the \ac{dl} model. \ac{fa} is proposed by Sabour \etal~\cite{sabour2015adversarial}. It is a targeted attack and alters the internal layers of the \ac{dl} model by minimizing the distance of the representation of intermediate layers instead of last layer output. In~\cite{baluja2017adversarial}, the authors used \ac{atn} to generate adversarial examples. \ac{atn} uses joint loss function, the first one $\ell_x(x, x^{\prime})$ insures the perceptual similarity between clean and adversarial samples, while the second loss function $\ell_y(f(x^{\prime}),t)$ insures that the softmax of \ac{ae} yields different prediction class than of clean sample. Other attacks are designed as a robustness property for defense and detection techniques. For instance, instead of computing the gradient over the clean samples, \ac{eot}~\cite{athalye2018synthesizing} algorithm computes the gradient over the expected transformation to the input. While \ac{bpda}~\cite{athalye2018obfuscated} attack is designed to bypass non-differentiable defenses by approximating its derivative as the derivative of the identity function. Last but not least, \ac{hca}~\cite{carlini2017adversarial} is an $L_2$ \ac{cw} attack with high confidence value $k$ (see \eqref{eq:cw_loss}), and is used to fool the detection techniques.

\subsubsection{Black box Attacks} \label{sec:bb_attacks}
\textbf{\acs{zoo} attack~\cite{chen2017zoo}.} Since gradients have to be estimated to generate the \ac{ae}, \ac{zoo} attack monitors the changes in softmax output $f(x)$, i.e., prediction confidence, when input sample is tuned. It uses symmetric difference quotient to estimate the gradient and Hessian using
\begin{ceqn}
    \begin{equation} \label{eq:zoo}
    \begin{gathered}
        \frac{\partial f(x)}{\partial x_i} \approx	\frac{f(x+he_i)-f(x-he_i)}{2h}, \\
        \frac{\partial^2 f(x)}{\partial x_i^2} \approx	\frac{f(x+he_i)-2f(x)+f(x-he_i)}{h^2}, \\
    \end{gathered}
    \end{equation}
\end{ceqn}
where $h$ is a small constant and $e_i$ is a standard basis vector with only the $i$-th component as 1.\\\\
\textbf{Pixel Attacks ~\cite{su2019one}.} The series starts with One-Pixel Attack in which the algorithm changes only one pixel to fool the \ac{dl} model. It uses the \ac{de}, one of \acs{ea}, to solve this optimization problem.
\begin{ceqn}
    \begin{equation} \label{eq:opa}
        \underset{\delta}{\text{max }} f_{adv}(x+\delta) \text{ , such that } ||\delta||_0 \leq d 
    \end{equation}
\end{ceqn}
where $d$ is a small number and equal to one in case of one-pixel. Kotyan \etal~\cite{kotyan2019adversarial} generalized one-pixel attack algorithm, they proposed two variants: \ac{pa} and \ac{ta}. \ac{pa} alters more than one pixel ($d>1$ in \eqref{eq:opa}) while \ac{ta} uses $L_\infty$-norm to solve the optimization problem in \eqref{eq:opa}. \\\\
\textbf{\acs{st} attack~\cite{engstrom2019exploring}.} Engstrom \etal showed that \ac{dl} models are vulnerable to translation and rotation changes of input samples. They proposed the attack in order to make the \ac{dl} models more robust using data augmentation during the training. \ac{st} attack solves the optimization problem 
\begin{ceqn}
    \begin{equation} \label{eq:st}
        \underset{\delta u,\delta v, \theta}{\text{max }} \ell(f(x^{\prime}), y) \text{ , for } x^{\prime}=T(x;\delta u,\delta v, \theta) 
    \end{equation}
\end{ceqn}
where $T, \delta u,\delta v \text{ and }\theta$ are, the transform function,\linebreak $x$-coordinate translation, $y$-coordinate translation and angle rotation, respectively.\\\\
\textbf{\acs{sa} attack~\cite{andriushchenko2020square}.} Via random search strategy and at each iteration of the algorithm, \ac{sa} selects colored $\epsilon$-bounded localized square shaped updates at random positions in order to generate perturbation $\delta$ that satisfies the optimization problem
\begin{ceqn}
    \begin{equation} \label{eq:sa}
        \underset{x^{\prime} \in [0,1]^n}{\text{min }} \ell(f(x^{\prime}), y) \text{ , such that } ||\delta||_p \leq \epsilon
    \end{equation}
\end{ceqn}
where $\ell(f(x^{\prime}), y)=f_y(x^{\prime})-\max_{k\neq y}f_k(x^{\prime})$. $f_y(x^\prime)$ and $f_k(x^\prime)$ are the prediction probability scores of $x^\prime$ for $y$ and $k$ classes, respectively.  The algorithm has two $L_p$ variants ($L_2$ and $L_\infty$).\\\\
\textbf{Other attacks.} Some other black box attacks are proposed in the literature. For instance,  \ac{ba}~\cite{brendel2017decision} is a black box attack that starts from largely perturbed adversarial example $\delta$ and moves towards the clean input class boundary by minimizing the $||\delta||_2$, while staying adversarial. Another boundary-decision based attack that depends on estimating gradient-based direction was proposed in~\cite{chen2020hopskipjumpattack}, it is known as \hop Attack. It achieves competitive performance compared to \ac{ba}~\cite{brendel2017decision}. \ac{upset} and \ac{angri} are two algorithms proposed by Sarkar \etal~\cite{sarkar2017upset}. The former generates one universal perturbation for each class in the dataset using a residual network, while the latter generates image-specific perturbation using dense network. In~\cite{nguyen2015deep}, Nguyen \etal proposed \ac{cppn} and showed that it is possible to generate unrecognizable images to humans but the \ac{dl} model predicts them with very high confidence. This does not fulfill the definition of the adversarial attacks although it fools the \ac{dl} model.

\subsection{Defense methods} \label{sec: defense}
Hardening the \ac{nn} models to avoid adversarial attacks is the aim of the defense techniques. In order to harden the \ac{nn} models and defend against the attacks, adversarial training approaches~\cite{goodfellow2014explaining,madry2017towards,xie2020smooth,tramer2017ensemble} include \acp{ae} in the training process. In~\cite{goodfellow2014explaining}, \ac{fgsm} attacks are added to the training process on MNIST dataset. While in~\cite{madry2017towards,xie2020smooth}, \acp{ae} generated using \ac{pgd} attacks are \linebreak added to the datasets. These models are not robust against transferred perturbations, hence, Tramer \etal~\cite{tramer2017ensemble} introduced ensemble adversarial training in which the transferred perturbations are included. The main limitations of adversarial training are: 1) it requires previous knowledge about the attacks and hence it is not robust against new/unknown attacks and 2) adversarial training still not robust against black box attacks~\cite{tramer2017ensemble}.

Feature denoising~\cite{xie2019feature,borkar2020defending,liao2018defense} is another approach for defending against \acp{aa}. In which, during the inference time, the input sample features are denoised  after some/all model layers. In~\cite{xie2019feature}, layer outputs are denoised using non-local means filters. Borkar \etal \cite{borkar2020defending} introduced selective feature regeneration as a denoising process. While in~\cite{liao2018defense}, a high-level representation guided denoiser (HGD) is proposed. It uses a loss function defined as the difference between the target model’s outputs activated by the clean image and denoised image. Feature denoising doesn't change the fact that the hardened model is still differentiable which makes it not robust to white box attacks. Moreover, it is time consuming since it requires end-to-end training.

In pre-processing approaches~\cite{mustafa2019image,prakash2018deflecting,bakhti2019ddsa}, the input sample is denoised first by removing added perturbations, then the denoised input  is passed to the base \ac{nn} model. In~\cite{prakash2018deflecting}, the pixel deflection algorithm changes the input sample to be much like natural image statistics by corrupting the input image by redistributing pixel values. While in~\cite{mustafa2019image}, the image restoration process uses wavelet denoising and super resolution techniques. Still, like any image denoising algorithms, this technique while removing perturbations other distortions will be added to image content. Moreover, it \linebreak doesn't stand against expectation over transformation (EOT) attacks \cite{athalye2018synthesizing}. 

Finally, the gradient masking techniques~\cite{papernot2017practical,papernot2016distillation,gu2014towards,nayebi2017biologically} try to train the \ac{nn} model to have a gradient close to 0 so that the model is less sensitive to small perturbations in the input sample. This technique yields robust models against white box attacks but not against black box attacks. For instance, Papernot \etal~\cite{papernot2016distillation} used the knowledge distillation concept and proposed ``defensive distillation" in which the output (smoothed labels) of the \ac{nn} model is used to train the \ac{nn} model. Then, they hide the model gradient by replacing the Softmax layer with a harder version. Some works show that this technique can be broken~\cite{papernot2017practical,carlini2017towards,carlini2016defensive}.

\section{Adversarial example detection methods} \label{sec: detection_methods}
\vspace{-2mm}
Despite the argument that \acp{ae} detection methods are  defense methods or not, we believe that the two have different specific goals while agreeing on a larger goal of defeating attack. Defense methods aim at classifying clean samples and their adversarial version with the same prediction class, while detection methods aim at classifying the input whether it is adversarial or not. As emphasized in~\cite{carlini2017adversarial}, no defenses have been able to classify adversarial examples correctly, and some research efforts are turned to design detection methods. Although the detection methods are found to be vulnerable to well-crafted attacks \cite{carlini2017adversarial}, the detection method might be an added value to the system even if a robust defense classifier is used. For instance, baseline classifier output may not agree with the robust classifier, and you need to know if this because the input was an \ac{ae} or not.

\begin{figure*}[!ht]
    \begin{center}
        \includegraphics[width=\linewidth, keepaspectratio]{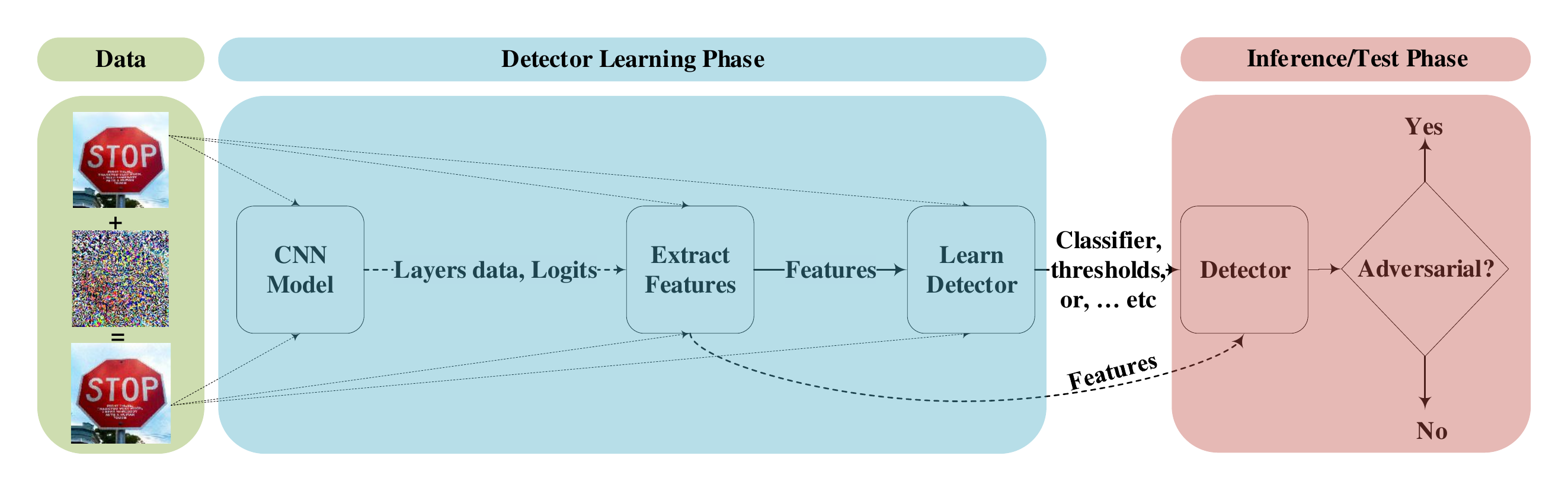}\vspace{-5mm}
    \end{center}
    \caption{Abstract overview of learning a detector for a victim model. On one side, Data side, the adversary generates \ac{ae} using adversarial attack algorithm with the help of knowledge that he got about the victim model. On the other side, in the Detector Learning Phase, the defender trains the detector using the information that he got from the victim model, training data, and/or attacked/adversarial data. In Inference/Test Phase, the adversary replaces the clean input with the adversarial one, then, the detector checks the input and recommend if it is clean or adversarial which should be rejected. }
    \label{fig:detectors_abstract}
\end{figure*}

\begin{figure*}[!ht]
    \centering
    \resizebox{\textwidth}{!}{%
    \begin{forest}
    forked edges,
    for tree={draw,align=center,edge={-latex},fill=white,blur shadow,
        where level=1{
          for descendants={%
           grow'=0,
           folder,
           l sep'+=2.5pt,
           }, 
        }{}
    }
    [\acp{ae} Detection Methods
     [Supervised
      [Network invariant
        [SafetyNet\\\cite{lu2017safetynet}]
        [Histogram\\\cite{pertigkiozoglou2018detecting}]
        [\acp{ae}\\evolution\\\cite{carrara2018adversarial}]
        [Dynamic\\Adversary\\Training\\\cite{metzen2017detecting}]
        [RAID\\\cite{eniser2020raid}]
      ]
      [Auxiliary model
        [Model\\Uncertainty\\\cite{feinman2017detecting,smith2018understanding}, draw=red!80, fill=red!10,]
        [Softmax\\based\\\cite{hendrycks2016baseline,pertigkiozoglou2018detecting}\\\cite{aigrain2019detecting,monteiro2019generalizable}]
        [Raw \acp{ae}\\\cite{gong2017adversarial,grosse2017statistical}\\\cite{hosseini2017blocking,pertigkiozoglou2018detecting}]
        [Gradient\\based\\\cite{lust2020gran}]
        [\acs{nss}\\\cite{kherchouche2020detection}, draw=red!80, fill=red!10,]
        [E\&R\\\cite{zuo2020exploiting}]
      ]
      [Statistical
        [\acs{mmd}\\\cite{grosse2017statistical}]
        [\acs{pca}\\\cite{li2017adversarial}]
        [\acs{kd}\\\cite{feinman2017detecting}, draw=red!80, fill=red!10,]
        [\acs{lid}\\\cite{ma2018characterizing}, draw=red!80, fill=red!10,]
        [\acs{knn}\\\cite{cohen2020detecting}]
        [Mahalanobis\\\cite{lee2018simple}]
      ]
     ]
     [Unsupervised
      [Network invariant
        [NIC\\\cite{ma2019nic}, draw=red!80, fill=red!10,]
      ]
      [Object-based
        [UnMask\\\cite{freitas2020unmask}]
      ]
      [Denoiser
        [PixelDefend\\\cite{song2017pixeldefend}]
        [Magnet\\\cite{meng2017magnet}, draw=red!80, fill=red!10,]
      ]
      [Statistical
        [Softmax\\based\\\cite{hendrycks2016baseline}]
        [\acs{pca}\\\cite{hendrycks2016early}]
        [\acs{gmm}\\\cite{zheng2018robust}]
      ]
      [Feature Squeezing
        [Bit-Depth\\and\\Smoothing\\\cite{xu2017feature}, draw=red!80, fill=red!10,]
        [Adaptive\\Noise\\Reduction\\\cite{liang2017detecting}]
      ]
      [Auxiliary model
        [\acs{knn}\\classifier\\\cite{carrara2017detecting}]
        [Reverse\\Cross-Entropy\\\cite{pang2018towards}]
        [Uncertainty\\\cite{sheikholeslami2019minimum}]
        [\acs{dnr}\\\cite{sotgiu2020deep}, draw=red!80, fill=red!10,]
        [\acs{sfad}\\\cite{aldahdooh2021selective}, draw=red!80, fill=red!10,]
      ]
     ]
    ]
    \end{forest}
    }
    \vspace{2mm}
    \caption{Categories of Adversarial Examples Detection Methods. The highlighted red rectangles are the detection methods considered in the experimental study.}
    \label{fig:cat_dete}
\end{figure*}
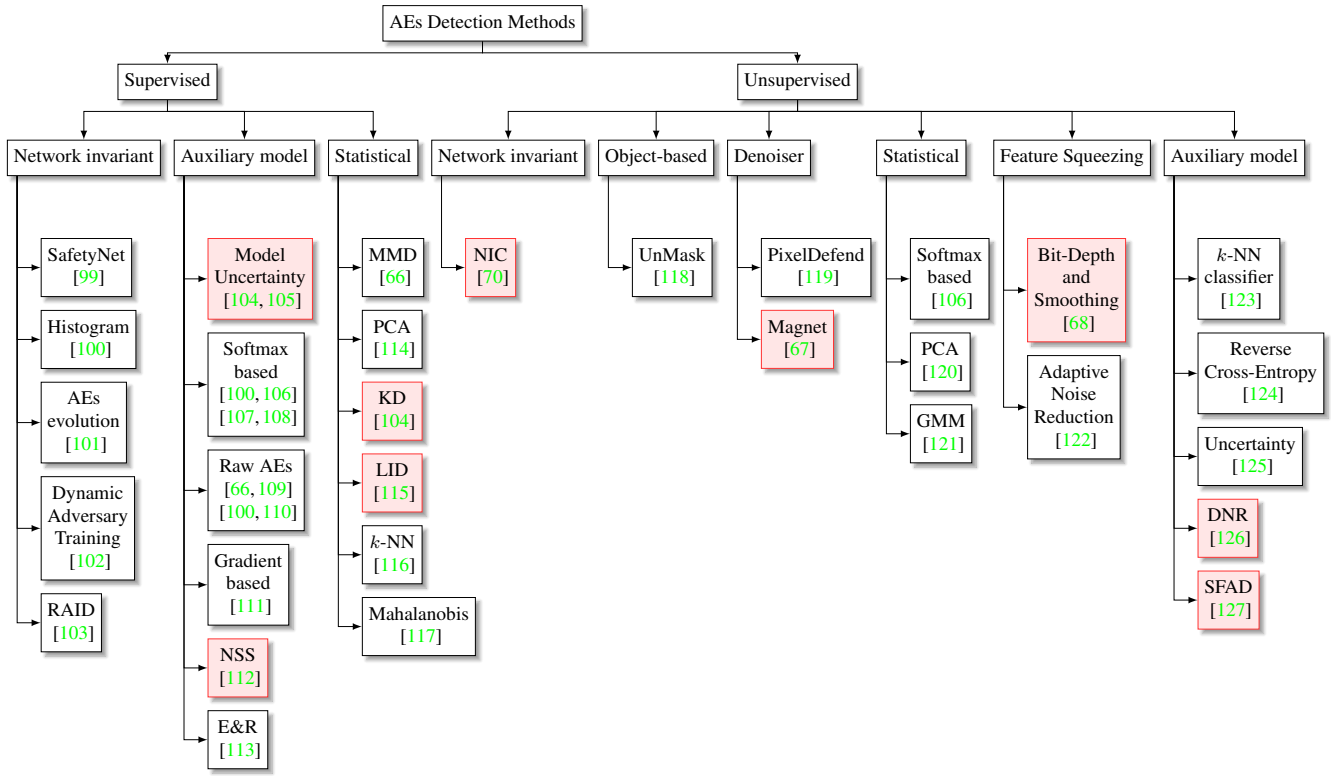

Hence, in this section, \acp{ae} detection methods will be discussed in detail. 
Figure \ref{fig:detectors_abstract} shows the abstract overview of how \acp{ae} detectors work. Detectors are considered as 3$^{rd}$ party entities that reject adversarial inputs and let clean inputs pass to the \textit{victim} \ac{dl} model. As will be discussed in this section, detectors differ in two factors; 1) using knowledge of adversarial attacks or not, and 2) the technique that is used to distinguish clean and adversarial inputs. Thus, we firstly categorize the detector methods with respect to the former factor and then to the latter one as illustrated in Figure \ref{fig:cat_dete}. In order to assess detector's performance, we consider the following criteria:
\begin{itemize}
\item \textbf{Detection rate:} It is the accuracy of the detector and it is measured by the number of successful\footnote{Successful \acp{ae} are the attacked samples that are able to fool the learning model, while the failed \acp{ae} are the attacked samples that are not able to fool the learning model. } \acp{ae} that are predicted by the detector and divided by the total number of successful \acp{ae}. The higher the better.

\item \textbf{\Ac{fpr}:} It is a very important criteria, and it is dedicated to know to what extent the detector \detector \ treats the clean inputs as adversarial ones. It is measured by calculating the number of clean inputs that are detected as adversarial inputs divided by the total number of clean inputs. The lower the better.

\item \textbf{Complexity:} It is the needed time to train the detector \detector. Some industries have sufficient hardware capability to run detectors with high computational complexity, but in the event that they have new data or need to include new attacks, it is inappropriate to train very complex models many times.

\item \textbf{Overhead:} It is related to the detector \detector \  architecture and the extra parameter size required to deploy the detector. The less the better, to be suitable platforms with limited memory and computation resources such mobile devices.

\item \textbf{Inference time latency:} It is the response time of the detector \detector \ to tell if the input is adversarial or not. To be appropriate for real-time applications, the less the better.
\end{itemize}
Table \ref{tab:compare_stateoftheart} summarizes the detection methods and highlights the main characteristic of each in terms of performance reported in their original papers and review papers. We rank the detection accuracy with up to five stars, since it is not fair to compare it with real numbers since, they are tested on different victim models, datasets, and attacks.


\begin{table*}[!ht]
    \centering
    \caption{Summary of state-of-the-art detection methods and the corresponding detection accuracy performance as reported in their original paper. \textbf{M}=MINIST, \textbf{C}=CIFAR-10, \textbf{S}=SVHN, \textbf{I}=ImageNet and \textbf{TI}=\image. Detection rate in average: $\bigstar\bigstar\bigstar\bigstar\bigstar$=90-100, $\bigstar \bigstar \bigstar \bigstar $=80-89, $\bigstar \bigstar \bigstar $=70-79, $\bigstar \bigstar $=50-69, and $\bigstar $=0-49}
    \label{tab:compare_stateoftheart}
    \resizebox{\textwidth}{!}{%
        \begin{tabular}{|m{0.02\textwidth}|m{0.14\textwidth}|m{0.15\textwidth}|m{0.22\textwidth}|m{0.30\textwidth}|}
        \hline
             \rule{0pt}{35pt}\rotatebox{90}{Category} & \rule{0pt}{13pt}Sub Category & \rule{0pt}{13pt}Model & \rule{0pt}{13pt}Tested against & \rule{0pt}{13pt}Performance Notes \\ \hline
             \multirow{27}{*}{\rotatebox{90}{Supervised}} & \multirow{11}{0.14\textwidth}{Auxiliary model} & Uncertainty~\cite{feinman2017detecting} & \ac{fgsm}, \ac{bim}, \ac{cw}, \ac{jsm} & M($\bigstar \bigstar \bigstar \bigstar $), C($\bigstar \bigstar $), S($\bigstar \bigstar $), Circumventable \cite{carlini2017adversarial}  \\ \cline{3-5}
             & & Softmax \cite{pertigkiozoglou2018detecting} & \ac{bim}, \ac{df} & M($\bigstar \bigstar \bigstar \bigstar \bigstar $)  \\ \cline{3-5}
             & & Softmax \cite{aigrain2019detecting} & \ac{fgsm}, \ac{bim}, \ac{df} & M($\bigstar \bigstar \bigstar \bigstar \bigstar $), C($\bigstar \bigstar \bigstar \bigstar \bigstar $)  \\ \cline{3-5}
             & & Softmax \cite{monteiro2019generalizable} & \ac{fgsm}, \ac{bim}, \ac{jsm}, \ac{df} & M($\bigstar \bigstar \bigstar \bigstar \bigstar $), C($\bigstar \bigstar \bigstar \bigstar \bigstar $)  \\ \cline{3-5}
             & & Raw \acp{ae} \cite{gong2017adversarial} & \ac{fgsm}, \ac{bim} & M($\bigstar \bigstar \bigstar \bigstar \bigstar $), C($\bigstar \bigstar \bigstar \bigstar \bigstar $), S($\bigstar \bigstar \bigstar \bigstar \bigstar $), Circumventable \cite{carlini2017adversarial}  \\ \cline{3-5}
             & & Raw \acp{ae} \cite{grosse2017statistical} &\ac{fgsm}, \ac{jsm}, & M($\bigstar \bigstar \bigstar \bigstar \bigstar $), Circumventable and bad performance for CIFAR-10 \cite{carlini2017adversarial}  \\ \cline{3-5}
             & & Raw \acp{ae} \cite{pertigkiozoglou2018detecting} & \ac{bim}, \ac{df} & M($\bigstar \bigstar \bigstar \bigstar $)  \\ \cline{3-5}
             & & \acs{nss} \cite{kherchouche2020detection} &\ac{fgsm}, \ac{bim}, \ac{cw}, \ac{df} & M($\bigstar \bigstar \bigstar \bigstar \bigstar $), C($\bigstar \bigstar \bigstar \bigstar \bigstar $), I($\bigstar \bigstar \bigstar \bigstar \bigstar $)  \\ \cline{3-5}
             & & Gradient \cite{lust2020gran} &\ac{fgsm}, \ac{bim}, \ac{jsm}, \ac{cw} & M($\bigstar \bigstar \bigstar \bigstar \bigstar $), C($\bigstar \bigstar \bigstar \bigstar \bigstar $), S($\bigstar \bigstar \bigstar \bigstar \bigstar $)  \\ \cline{3-5}
             & & E\&R \cite{zuo2020exploiting} &\ac{cw}, \ac{df} & C($\bigstar \bigstar \bigstar \bigstar \bigstar $), I($\bigstar \bigstar \bigstar \bigstar \bigstar $)  \\ \cline{2-5}
             & \multirow{7}{0.14\textwidth}{Statistical}  & \acs{mmd} \cite{grosse2017statistical} & \ac{fgsm}, \ac{jsm} & MNIST($\bigstar \bigstar \bigstar \bigstar \bigstar $), Circumventable \cite{carlini2017adversarial}  \\ \cline{3-5}
             & & \acs{pca} \cite{li2017adversarial} &\ac{bfg}, & I($\bigstar \bigstar \bigstar \bigstar $), Circumventable \cite{carlini2017adversarial} and bad performance for M and C  \\ \cline{3-5}
             & & \acs{kd} \cite{feinman2017detecting} & \ac{fgsm}, \ac{bim}, \ac{cw}, \ac{jsm} & M($\bigstar \bigstar \bigstar \bigstar $), C($\bigstar \bigstar \bigstar \bigstar $), S($\bigstar \bigstar $), Circumventable \cite{carlini2017adversarial}  \\ \cline{3-5}
             & & \acs{lid} \cite{ma2018characterizing} &\ac{fgsm}, \ac{bim}, \ac{jsm} & M($\bigstar \bigstar \bigstar \bigstar \bigstar $), C($\bigstar \bigstar \bigstar \bigstar \bigstar $), S($\bigstar \bigstar \bigstar \bigstar \bigstar $), Circumventable \cite{athalye2018obfuscated}  \\ \cline{3-5}
             & & Mahalanobis \cite{lee2018simple} &\ac{fgsm} & C($\bigstar \bigstar \bigstar \bigstar \bigstar $), S($\bigstar \bigstar \bigstar \bigstar \bigstar $) \\ \cline{3-5}
             & & \acs{knn} \cite{cohen2020detecting} &\ac{fgsm}, \ac{jsm}, \ac{df}, \ac{pgd}, \ac{cw} & C($\bigstar \bigstar \bigstar \bigstar \bigstar $), S($\bigstar \bigstar \bigstar \bigstar \bigstar $)  \\ \cline{2-5}
             & \multirow{6}{0.14\textwidth}{Network invariant}  & Safetynet \cite{lu2017safetynet} & \ac{fgsm}, \ac{bim}, \ac{df} & C($\bigstar \bigstar \bigstar \bigstar $), I($\bigstar \bigstar \bigstar $)  \\ \cline{3-5}
             & & Histogram \cite{pertigkiozoglou2018detecting} &\ac{bim}, \ac{df} & M($\bigstar \bigstar \bigstar \bigstar $)  \\ \cline{3-5}
             & & Dynamic Adversary Training \cite{metzen2017detecting} & \ac{fgsm}, \ac{bim}, \ac{df} & C($\bigstar \bigstar \bigstar \bigstar $), I($\bigstar \bigstar \bigstar $),   Circumventable \cite{carlini2017adversarial} \\ \cline{3-5}
             & & \acp{ae} evolution \cite{carrara2018adversarial} & \ac{bfg}, \ac{fgsm}, \ac{bim}, \ac{pgd}& I($\bigstar \bigstar \bigstar \bigstar $) \\ \cline{3-5}
             & & RAID \cite{eniser2020raid} &\ac{fgsm}, \ac{bim}, \ac{pgd}, \ac{df}, \ac{cw}, \ac{jsm} & M($\bigstar \bigstar \bigstar \bigstar $), C($\bigstar \bigstar \bigstar \bigstar \bigstar $)  \\ \cline{1-5}
             \multirow{17}{*}{\rotatebox{90}{Unsupervised}} & \multirow{6}{0.14\textwidth}{Auxiliary model} & \acs{knn} \cite{carrara2017detecting} & \ac{bfg}, \ac{fgsm}, & I($\bigstar \bigstar \bigstar \bigstar $)  \\ \cline{3-5}
             & & Reverse Cross-Entropy \cite{pang2018towards} & \ac{fgsm}, \ac{bim}, \ac{cw}, \ac{jsm} & M($\bigstar \bigstar \bigstar \bigstar \bigstar $), C($\bigstar \bigstar \bigstar \bigstar \bigstar $)  \\ \cline{3-5}
             & & Uncertainty \cite{sheikholeslami2019minimum} &\ac{fgsm}, \ac{bim}, \ac{cw} & C($\bigstar \bigstar \bigstar \bigstar $)  \\ \cline{3-5}
             & & \acs{dnr} \cite{sotgiu2020deep} & optimized-\ac{pgd}, & M($\bigstar \bigstar \bigstar $), C($\bigstar $)  \\ \cline{3-5}
             & & SFAD \cite{aldahdooh2021selective} &\ac{fgsm}, \ac{pgd}, \ac{cw}, \ac{df} & M($\bigstar \bigstar \bigstar \bigstar \bigstar $), C($\bigstar \bigstar \bigstar $)  \\ \cline{2-5}
             & \multirow{2}{0.14\textwidth}{Statistical} & \acs{pca} \cite{hendrycks2016early}  &\ac{fgsm}, \ac{bim} & M($\bigstar \bigstar \bigstar \bigstar \bigstar $), Circumventable \cite{carlini2017adversarial} and not effective for CIFAR-10 \\ \cline{3-5}
             & & \acs{gmm} \cite{zheng2018robust} &\ac{fgsm}, & MNIST($\bigstar \bigstar \bigstar \bigstar \bigstar $) \\ \cline{2-5}
             & \multirow{2}{0.1\textwidth}{Denoiser} & PixelDefend \cite{song2017pixeldefend} &\ac{fgsm}, \ac{bim}, \ac{df}, \ac{cw}& C($\bigstar \bigstar \bigstar \bigstar $), Circumventable \cite{athalye2018obfuscated}  \\ \cline{3-5}
             & & MagNet \cite{meng2017magnet} &\ac{fgsm}, \ac{bim}, \ac{df}, \ac{cw} & M($\bigstar \bigstar \bigstar \bigstar \bigstar $), C($\bigstar \bigstar \bigstar \bigstar $), Circumventable \cite{carlini2017magnet}  \\ \cline{2-5}
             & \multirow{3}{0.14\textwidth}{Feature Squeezing} & Bit-Depth  and  Smoothing \cite{xu2017feature} &\ac{fgsm}, \ac{bim}, \ac{df}, \ac{cw}, \ac{jsm} & M($\bigstar \bigstar \bigstar \bigstar \bigstar $), C($\bigstar \bigstar $), I($\bigstar \bigstar \bigstar $)  \\ \cline{3-5}
             & & Adaptive Noise Reduction \cite{liang2017detecting} & \ac{fgsm}, \ac{cw}, \ac{df} & M($\bigstar \bigstar \bigstar \bigstar \bigstar $), I($\bigstar \bigstar \bigstar \bigstar \bigstar $)  \\ \cline{2-5}
             & Network invariant & NIC \cite{ma2019nic} &\ac{fgsm}, \ac{bim}, \ac{df}, \ac{cw}, \ac{jsm} & M($\bigstar \bigstar \bigstar \bigstar \bigstar $), C($\bigstar \bigstar \bigstar \bigstar \bigstar $), I($\bigstar \bigstar \bigstar \bigstar \bigstar $)  \\ \cline{2-5}
             & Object-Based & UnMask \cite{freitas2020unmask} & \ac{bim}, \ac{pgd} & I($\bigstar \bigstar \bigstar \bigstar $)  \\ \hline
        \end{tabular}
   }
\end{table*}

\subsection{Supervised detection} \label{sec: supervised_detectors}
In supervised detection, the defender considers \acp{ae} generated by one or more adversarial attack algorithms in designing and training the detector \  \detector. It is believed that \acp{ae} have distinguishable features that make them different from clean inputs~\cite{ilyas2019adversarial}, hence, defenders take this advantage to build a robust detector \detector. To accomplish this, many approaches have been presented in the literature. 

\subsubsection{Auxiliary model approach}\label{sec: super_aux}
In this approach, models exploit features that can be extracted by monitoring the clean and adversarial samples behaviors. Then, either classifiers or thresholds are built and calculated. 

\textbf{Model uncertainty.} Defenders are using \ac{dl} models uncertainty of clean and adversarial inputs. The uncertainty is usually measured by adding randomness to the model using Dropout~\cite{srivastava2014dropout} technique. The idea is that with many dropouts, clean input class prediction remains correct, while it is not with \acp{ae}. Uncertainty values are used as features to build a binary classifier as a detector \detector. Feinman \etal~\cite{feinman2017detecting} proposed \ac{bu} metric, which uses Monte Carlo dropout to estimate the uncertainty, to detect those \acp{ae} that are near the classes manifold, while Smith \etal~\cite{smith2018understanding} used mutual information method for such task. 

\textbf{Softmax/logits-based.} Hendrycks \etal~\cite{hendrycks2016baseline} showed that softmax prediction probabilities can be used to detect abnormality, they append a decoder to reconstruct clean input from the softmax and trained it jointly with the baseline classifier. Then, they train a classifier, a detector \detector, using the reconstructed input, logits and confidence scores for clean and \acp{ae} inputs. In one of the methods that were proposed in~\cite{pertigkiozoglou2018detecting}, Pertigkiozoglou \etal used model vector features, i.e., confidence outputs, to calculate regularized vector features. The baseline classifier is retrained by adding this regularized vector features to the last layer of the classifier. The detector \detector \ considers an input as \ac{ae} if there is no match between baseline classifier and the retrained classifier. Aigrain \etal~\cite{aigrain2019detecting} built a simple \ac{nn} detector \detector \ which takes the baseline model logits of clean and \acp{ae} as inputs to build a binary classifier. Finally, following the hypothesis that different models make different mistakes when presented with the same attack inputs, Monteiro \etal~\cite{monteiro2019generalizable} proposed a bi-model mismatch detection. The detector \detector \  is a binary \ac{rbf}-\ac{svm} classifier. Its inputs are the output of two baseline classifiers of clean and \acp{ae}. 

\textbf{Raw \acp{ae}-based classifier.} Gong \etal~\cite{gong2017adversarial} trained a binary classifier, detector \detector, that is completely separated from the baseline classifier and takes as input the clean and adversarial images. In~\cite{grosse2017statistical,hosseini2017blocking}, the authors retrained the baseline classifier with a new added class, i.e., adversarial class. Hosseini \etal employed adversarial training and the used training labels are performed using label smoothing~\cite{szegedy2015rethinking}. In one of the methods that were proposed in~\cite{pertigkiozoglou2018detecting}, the authors took advantage of the \ac{dl} model input's parts that are ignored by the model to detect the \acp{ae}. They iteratively perturbed the input, clean or adversarial, and if the probability of the predicted input class is less than the threshold, then the input is declared as adversarial.

\textbf{\Ac{nss}.} \ac{nss} has been used in many areas of image processing, especially in image quality estimation, since it has been proved that statistics of natural images are different from those of manipulated images. Kherchouche \etal~\cite{kherchouche2020detection} followed this assumption and built a binary classifier that takes as input features parameters of the Generalized Gaussian Distribution (GGD) and Asymmetric Generalized Distribution (AGGD) computed from the \ac{mscn} coefficients \cite{mittal2012no} of clean images and \ac{pgd}-based \acp{ae}.

\textbf{Gradient based.} Lust \etal~\cite{lust2020gran} proposed a detector \detector \ named GraN. At each layer, they calculated the gradient norm of a smoothed input, clean and adversarial, with respect to the predicted class of the baseline classifier. Then, they train a binary classifier \detector \ to detect \acp{ae} in inference time.

\textbf{Erase\&restore (E\&R)~\cite{zuo2020exploiting}}. In this model, Zuo \etal  proposed a binary classifier, a detector \detector, to train clean and $L_2$-norm adversarial samples after processing. Firstly, the input samples are processed by erasing some pixels and restoring them in an inpainting process. Secondly, the confidence probability is calculated using the baseline classifier. Finally, the processed confidence probability is then passed to the binary classifier. The detector \detector \ announces an input as adversarial if the binary classifier says so.

\subsubsection{Statistical approach} \label{sec: super_stat}
In this approach, different statistical properties of clean and \acp{ae} inputs are calculated and then used to build the detector. These properties are more related to in- or out- of training data distribution/manifolds. The following statistical approaches are used in the literature:

\textbf{\Ac{mmd}}. Grosse \etal~\cite{grosse2017statistical} employed a statistical test, called \ac{mmd}~\cite{gretton2012kernel}, to distinguish adversarial examples from the model’s training data. It is model-agnostic and kernel-based two-sample test. To answer the hypothesis test assumption, the detector \detector \ firstly computes the \ac{mmd} between clean and \acp{ae} samples, $a=MMD(x,x^{\prime})$. Then, shuffle
the elements of $x$ and $x^{\prime}$ into two new sets $y_1$ and $y_2$, and compute $b=MMD(y_1,y_2)$. Finally, conclude that $x$ and $x^{\prime}$ are drawn from different distributions and reject the hypothesis if $a < b$.

\textbf{\Ac{pca}.} The work in~\cite{li2017adversarial} built cascade classifiers. Each \ac{svm} classifier corresponds to one layer. It is trained using clean and \acp{ae} samples. The input of the \ac{svm} is the \ac{pca} of each layer output. The detector \detector \ announces an input as clean if all classifiers say so.

\textbf{\Acf{kd}.} It was shown that \acp{ae} subspaces usually have lower density than clean samples especially if the input sample is far from a class manifold. Feinman \etal~\cite{feinman2017detecting} proposed \ac{kd} estimation for each class in the training data and then trained a binary classifier, detector \detector, using densities and uncertainties features of clean, noisy, and \acp{ae}. 

\textbf{\Ac{lid}.} As an alternative measure to \ac{kd}, Ma \etal in~\cite{ma2018characterizing} used \ac{lid} to calculate the distance distribution of the input sample to its neighbors to assess the space-filling capability of the region surrounding that input sample.

\textbf{Mahalanobis-based.} As an alternative measure to \ac{kd} and \ac{lid}, Lee \etal~\cite{lee2018simple} proposed Mahalanobis distance-based score to detect out-of-distribution and adversarial input samples. This confidence score is based on an induced generative classifier under \ac{gda} that actually replaces the softmax classifier.

\textbf{\Ac{knn}.} The work in~\cite{cohen2020detecting} firstly measured the impact/influence of every training sample on the validation set data and then found the most supportive training samples for any given validation example. Then, at each layer, using the \ac{dl} layers representative output, a \ac{knn} model is fitted to rank these supporting training samples. These features are extracted from clean and \acp{ae} to train a detector \detector. Recently, Mao \etal~\cite{mao2020learning} proposed Neighbor Context Encoder (NCE) detector. It used transformer~\cite{vaswani2017attention} to train a classifier with $k$ nearest neighbors to represent the surrounded subspace of the
detected sample.

\subsubsection{Network invariant approach}
It is believed that the clean and the adversarial samples yield different feature maps and different activation values for the network layers. Analysing this network invariant violation is the core components for many detection methods.

\textbf{Safetynet~\cite{lu2017safetynet}.} SafetyNet states the hypothesis ``Adversarial attacks work by producing different patterns of activation in late stage ReLUs to those produced by natural examples''. Hence, SafetyNet quantizes the last ReLU activation layer of the model and builds a binary \ac{svm} \ac{rbf} classifier.

\textbf{Dynamic adversary training~\cite{metzen2017detecting}}. Metzen \etal presented dynamic adversary training to harden the detector in which the classifier was trained with \acp{ae}. The detector \detector \ is augmented to the pre-trained classifier at a specific layer output. It takes layer's representative output for clean samples and for on fly generated \acp{ae} as input to build a binary classifier. 

\textbf{Histogram-based~\cite{pertigkiozoglou2018detecting}}. Pertigkiozoglou \etal observed that for \acp{ae} there is an increase in the values of some peaks of clean output while there is a decrease in the values on the rest of the points of the output. Hence, they built a binary \ac{svm} classifier which takes as inputs the histogram of the first convolutional layer output of the baseline classifier for clean and \acp{ae}.

\textbf{\acp{ae} evolution~\cite{carrara2018adversarial}}. Carrara \etal hypothesized that intermediate representations of \acp{ae} follow a different evolution with respect to clean inputs. The detector \detector \  encodes the relative positions of internal activations of points that represent the dense parts of the feature space. The detector is a binary classifier built on top of the pre-trained network and takes as inputs the encoded relative positions of internal activations of points that represent the dense parts of the feature space for \acp{ae} and clean inputs. 

\textbf{RAID~\cite{eniser2020raid}}. Eniser \etal built a binary classifier  that takes as inputs the differences in neuron activation values between clean and \acp{ae} inputs. In order to make the adaptive attacks much harder, the authors also provided an extension to RAID called Pooled-RAID. This latter aims at training a pool of detectors, each trained with a randomly selected number of neurons. In the test time, the Pooled-RAID selects randomly one detection classifier from the pool to test if the input is adversarial or not.

\subsection{Unsupervised detection}
The main limitation of supervised detection methods is that  they require prior knowledge about the attacks and hence they might not be robust against new/unknown attacks. In unsupervised detection, the defender considers only the clean training data in designing and training the detector \detector. It is also known as inconsistency prediction models since it depends on the fact that \acp{ae} might not fool every \ac{nn} model.\\ 
Basically, unsupervised detectors aim at reducing the limited input feature space available to adversaries and to accomplish this goal, many approaches have been presented in the literature. 
\subsubsection{Auxiliary model approach} \label{sec:unsuper_auxil}
Unlike auxiliary models of supervised detection, unsupervised models exploit features that can be concluded by monitoring only the clean samples behaviors. Then, either classifiers or thresholds are built and calculated. 

\textbf{\Ac{knn} classifier~\cite{carrara2017detecting}}. Carrara \etal used the output of one of the intermediate \ac{dl} model layers to build a \ac{knn} classifier. The output of this classifier is not used for the detection, but it is used to score the predicted class of baseline classifier. The detector \detector \ announces the input is adversarial if this score is less than a specified threshold. They also provided a process to use the \ac{pca} of the output of one of the intermediate \ac{dl} model layers to reduce the feature dimension.

\textbf{Reverse cross-entropy~\cite{pang2018towards}.} Pang \etal proposed a training procedure and a threshold-based detector. Firstly, the baseline classifier is retrained with a reverse cross-entropy loss function in order to better learn latent representations that will distinguish clean inputs and \acp{ae}. Then, for each class, a kernel density is estimated and, then, the threshold is calculated. Finally, the detector \detector announces an input as adversarial if its density score is less than the calculated threshold. The authors in~\cite{pang2018towards} also introduced an alternative estimation of Kernel density called Non-maximal entropy but they found that detection using kernel density estimation gives better results than non-ME Non-maximal entropy in most of the cases.

\textbf{Uncertainty-based.} Following \ac{bu} assumption that \acp{ae} distances from in-distribution data make the \ac{dl} model uncertainty differs from clean data, Sheikholeslami \etal~\cite{sheikholeslami2019minimum} proposed to introduce randomness for randomly sampled hidden units of each layer of \ac{dl} model. Then, the uncertainty is estimated for in-distribution training data and a mutual information based threshold is identified. They provided a layer-wise minimum variance solver to estimate the uncertainty. At inference time, the input image overall uncertainty is estimated using the hidden layers outputs. Detector \detector \ announces the input sample as adversarial if its mutual information is larger than the threshold.

\textbf{\Ac{dnr}~\cite{sotgiu2020deep}}. Sotgiu \etal proposed to use the $N$-last representative layers outputs of the baseline classifiers to build $N$-\ac{svm} classifiers with \ac{rbf} kernel. The output of these classifiers, i.e., the confidence probabilities, are combined to build the last classification task classifier which is an \ac{svm}-\ac{rbf} classifier. The detector \detector \ announces an adversarial input as adversarial if the maximum confidence probability is less than a predefined threshold.

\textbf{Selective detection~\cite{aldahdooh2021selective}.} Aldahdooh \etal proposed \ac{sfad} technique. They use the recent uncertainty method called SelectiveNet~\cite{selective2019} and integrated three detection modules. The first is the selective detection module, which is a threshold-based detection derived from uncertainty of clean training data using SelectiveNet. The second is confidence detection module, which is threshold-based detection derived from softmax probabilities of clean training data from \ac{sfad}'s classifiers. \ac{sfad}'s classifiers analyse the representative data of last $N$-layers as a key point to present robust features of input data using autoencoding, up/down sampling, bottleneck, and noise blocks. The last module is ensemble prediction, which is a mismatch-based prediction between the detector and the baseline \ac{dl} classifiers. 

\subsubsection{Statistical approach}
In this approach, different statistical properties of only clean inputs are calculated and then used to build the detector. These properties are more related to in- or out- of training data distribution/manifolds. The following statistical approaches are used in the literature:

\textbf{Softmax distribution~\cite{hendrycks2016baseline}.} Hendrycks \etal found that maximum/predicted class probability of in-distribution samples are higher than of out-of-distribution. This information is used and Kullback-Leibler divergence~\cite{kullback1951information} is computed between in-distribution and clean input samples to determine the threshold. 

\textbf{\ac{pca}~\cite{hendrycks2016early}}. Hendrycks \etal observed that the later \ac{pca} components variance of \acp{ae} is larger than those of clean inputs, hence, they proposed a detector \detector \ to declare the input as adversarial if the later \ac{pca} components variance is above the threshold.

\textbf{\ac{gmm}~\cite{zheng2018robust}}. Zheng \etal proposed a detection method called I-defender, referred here as ``intrinsic''. It explores the distributions of \ac{dl} model hidden states of the clean training data. I-defender uses \ac{gmm} to approximate the intrinsic hidden state distribution of each class. I-defender chooses to only model the state of the fully connected hidden layers and then a threshold for each class is calculated. The detector \detector \ announces the input sample as adversarial if its hidden state distribution probability is less than the predicted class's threshold. On the other hand, the work in~\cite{miller2019not} works under the assumption that the adversarial input 1) has atypically low likelihood compared to the density model of predicted class and is called ``too atypical'', and 2) has high likelihood for a class other than the class of clean input and is called ``too typical''. Hence, for each case, a two-class posterior is evaluated, i.e., one with respect to the density estimation and one with respect to the \ac{dl} model. The final score for ``too atypical'' and ``too typical'' are calculated using the Kullback-Leibler divergence. The detector \detector \  declares an input as adversarial if the score is larger than the predefined threshold.\\

\subsubsection{Denoiser approach}
To prevent the adversary from estimating the location of \acp{ae} accurately, one can make the input gradient very small or irregularly large. This phenomenon is known as exploding/vanishing gradients. One method to do that is to denoise or reconstruct \acp{ae} to maximize the ability to project the \acp{ae} to the training data manifold. The main limitation of using denoiser is that it is not guaranteed to remove all the noise to produce highly denoised inputs, and it might introduce extra distortion. Besides, it is not effective in denoising the $L_0$ attacks, since $L_0$ attacks target a few pixels and these pixels might not be denoised by the denoiser.

\textbf{PixelDefend~\cite{song2017pixeldefend}}. Generative models such as PixelCNN \cite{van2016conditional} explode the gradient by applying cumulative product of partial derivatives from each layer. PixelDefend detection~\cite{song2017pixeldefend} utilised PixelCNN to build its detector. Firstly, PixelDefend reconstructs/purifies the clean training data using PixelCNN and then computes the prediction probabilities using baseline classifier. It is found that reconstructed images have higher probabilities under in-distribution of training data. Then, the probability density of the training samples are computed. The detector \detector \ works by, firstly, computing the probability density of tested input. Secondly, this density is ranked with training data densities. Finally, the rank can be used as a test statistic, and $p$-value is calculated to determine if the input sample belongs to the in-distribution of training data or it is adversarial.

\textbf{Magnet~\cite{meng2017magnet}}. Magnet trains denoisers in clean training data to reconstruct the input samples. Magnet proposed two ways to detect \acp{ae}. The first one assumes that the reconstruction error will be small for clean images and large in \acp{ae} and hence, it calculates the reconstruction error as a score. The second way measures the distances between the predictions of input samples and their denoised/filtered versions. The detector \detector \ announces the input sample as adversarial if the score exceeds a predefined threshold.

\subsubsection{Feature Squeezing approach}
This approach aims at squeezing out unnecessarily features of input samples to destroy perturbations. This process will limit the features space available for the adversary but if the squeezer is not built efficiently, it may enlarge the perturbation.

\textbf{Bit-depth and smoothing~\cite{xu2017feature}}. Xu \etal squeezes the input samples by projecting/transforming it to produce new samples. They used color bit-depth reduction, local smoothing using median filter and non-local smoothing filter using non-local mean denoiser. The detector \detector \ considers the input as adversarial if the distance between predicted original input and the squeezed version exceeds the identified threshold. 

\textbf{Adaptive noise reduction~\cite{liang2017detecting}.} Liang \etal on the other hand, squeezes the input samples using scalar quantization and smoothing spatial filter. They used the image entropy as a metric to implement the adaptive noise reduction. The detector \detector \  considers the input as adversarial if the class of original input is different from the squeezed version.

\subsubsection{Network invariant approach}
Unlike the network invariant approach of supervised detection, here, the detector aims at observing behaviors of clean training data only in the intermediate \ac{dl} model layers. The recent work of Ma \etal\cite{ma2019nic} showed that if the two attack channels, the provenance channel and the activation value distribution channel, are monitored, then the \acp{ae} can be detected. Ma \etal\cite{ma2019nic} proposed a \ac{nic} method that builds a set of models for individual layers to describe the provenance and the activation value distribution channels. The provenance channel describes the instability of activated neurons set in the next layer when small changes are present in the input sample, while the activation value distribution channel describes the changes with the activation values of a layer. To train the invariant models, the authors used One-Class Classification (OCC) problem as a way to model in-distribution training data. The detector \detector \ is a joint OCC classifier that joins all invariant models' outputs. It announces the input sample as adversarial if the detector classifier declares the input is out-of-distribution. 

\subsubsection{Object-based approach}
In this approach, the aim is to extract object-based features from the input sample and compare them with training data of the same prediction label. UnMask is a method proposed by Freitas \etal~\cite{freitas2020unmask} that works as follows: firstly, assume the adversary altered a bicycle image to be predicted as a bird. UnMask first extracts object-based low-level features from the attacked image ``the bicycle'' and compares them with object-based low-level features of ``the bird''. Then, if there is a small overlap, the detector \detector \ will announce the input as adversarial. Also, Unmask continues ``as a defense'' to find which class in the training data classes has the highest overlap with the predicted one to announce the correct class.

\section{Experiment settings} \label{sec: exp_setting}
\subsection{Datasets}
In this work, we evaluate the detection methods on the following four datasets:

\textbf{MNIST~\cite{lecun1998gradient}}. It is a handwriting digit recognition dataset for digits from 0 to 9. It contains 70000 gray images/samples, 60000 for training and 10000 for testing.

\textbf{SVHN~\cite{netzer2011reading}}. It is a real street view house numbers recognition dataset. The numbers are cropped in digits of ten classes. It contains 99289 RGB images/samples, 73257 digits for training and 26032 digits for testing.

\textbf{CIFAR-10~\cite{krizhevsky2009learning}}. It is a collection of images that is usually used in computer vision tasks. It is $32\times 32$ RGB images of ten classes: airplanes, cars, birds, cats, deer, dogs, frogs, horses, ships, and trucks. It contains 60000 images, 50000 for training and 10000 for testing.

\textbf{Tiny ImageNet~\cite{yao2015tiny}}. It is a tiny version of ImageNet~\cite{imagenet_cvpr09} dataset. It contains $64\times 64$ RGB images and includes 200 classes. It is composed of 110,000 images, 100000 for training and 10000 for testing.

\subsection{Baseline ``Victim'' classifiers}
In order to evaluate the detection methods, we built and trained four baseline \textit{victim} models, one for each dataset.

\textbf{MNIST}. We built and trained a 6-layer CNN classifier for this dataset. It achieves state of the art results of 98.73\% accuracy. We follow the architecture that is described in \cite{sotgiu2020deep} and  shown in Table \ref{tab:mnist_arch}.

\textbf{SVHN}. We built and trained a 6-layer CNN classifier, similar to MNIST, for the SVHN dataset. It achieves state of the art results of 94.99\% accuracy. We use similar architecture of the MNIST and we only changed the number of neurons of the dense layers as shown in Table \ref{tab:svhn_arch}.

\textbf{CIFAR10}. Since the CIFAR10 dataset is not a complex task, we did not use complex \ac{cnn} architecture to avoid the phenomena of the \ac{cnn} not using saliency regions of clean images in predicting the correct class\cite{madry2017towards}. We follow the architecture that is described in \cite{sotgiu2020deep} and  shown in Table \ref{tab:cifar10_arch}. An 8-layer CNN classifier was built and trained for CIFAR10 dataset. It achieves accuracy of 89.11\%. 

\textbf{\image}. We use a classifier relying on DenseNet201 \cite{huang2017densely}, one of the state-of-the-art classifiers for image classification. We started with the DenseNet201 weights of ImageNet and then the model was fine-tuned for a 200-class classification task.  It achieves 65\% classification accuracy.

\begin{table}[t!]
\caption{MNIST baseline classifier architecture.}
\label{tab:mnist_arch}
\begin{tabular}{m{0.02\columnwidth}|m{0.56\columnwidth}|M{0.26\columnwidth}}
\hline
     \# & Layer & Description \\ \hline \hline
     1 & Conv2D + ReLU & 32 filters ($3\times3$) \\ \hline
     2 & Conv2D + ReLU + Max Pooling($2\times2$) & 32 filters ($3\times3$) \\ \hline
     3 & Conv2D + ReLU & 64 filters ($3\times3$) \\ \hline
     4 & Conv2D + ReLU + Max Pooling($2\times2$) & 64 filters ($3\times3$) \\ \hline 
     5 & Dense + ReLU + Dropout ($p=0.3$) & 256 units \\ \hline
     6 & Dense + ReLU & 256 units \\ \hline
     7 & Dense + Softmax & 10 classes \\ \hline
\end{tabular}
\end{table}

\begin{table}[t!]
\caption{SVHN baseline classifier architecture.}
\label{tab:svhn_arch}
\begin{tabular}{m{0.02\columnwidth}|m{0.56\columnwidth}|M{0.26\columnwidth}}
\hline
     \# & Layer & Description \\ \hline \hline
     1 & Conv2D + ReLU & 32 filters ($3\times3$) \\ \hline
     2 & Conv2D + ReLU + Max Pooling($2\times2$) & 32 filters ($3\times3$) \\ \hline
     3 & Conv2D + ReLU & 64 filters ($3\times3$) \\ \hline
     4 & Conv2D + ReLU + Max Pooling($2\times2$) & 64 filters ($3\times3$) \\ \hline 
     5 & Dense + ReLU + Dropout ($p=0.3$) & 512 units \\ \hline
     6 & Dense + ReLU & 128 units \\ \hline
     7 & Dense + Softmax & 10 classes \\ \hline
\end{tabular}
\end{table}

\begin{table}[t!]
\caption{CIFAR10 baseline classifier architecture.}
\label{tab:cifar10_arch}
\begin{tabular}{m{0.02\columnwidth}|m{0.56\columnwidth}|M{0.26\columnwidth}}
\hline
     \# & Layer & Description \\ \hline \hline
     1 & Conv2D + BatchNorm + ReLU & 64 filters ($3\times3$) \\ \hline
     2 & Conv2D + BatchNorm + ReLU + Max Pooling($2\times2$) + Dropout ($p=0.1$) & 64 filters ($3\times3$) \\ \hline
     4  & Conv2D + BatchNorm + ReLU & 128 filters ($3\times3$) \\ \hline
     5 & Conv2D + BatchNorm + ReLU + Max Pooling($2\times2$) + Dropout ($p=0.2$) & 128 filters ($3\times3$) \\ \hline 
     6 & Conv2D + BatchNorm + ReLU & 256 filters ($3\times3$) \\ \hline
     7 & Conv2D + BatchNorm + ReLU + Max Pooling($2\times2$) + Dropout ($p=0.3$) & 256 filters ($3\times3$) \\ \hline
     8 & Conv2D + BatchNorm + ReLU + Max Pooling($2\times2$) + Dropout ($p=0.4$) & 512 filters ($3\times3$) \\ \hline
     9 & Dense  & 512 units \\ \hline
     10 & Dense + Softmax & 10 classes \\ \hline
\end{tabular}
\end{table}

\begin{table}[t!]
\caption{Baseline classifiers' accuracies on normal clean testing data and attacked($\epsilon$) data.}
\label{tab:baseline_acc}
\resizebox{\linewidth}{!}{
\begin{tabular}{M{0.1\linewidth}|P{0.25\linewidth}|M{0.12\linewidth}|M{0.11\linewidth}|M{0.11\linewidth}|M{0.15\linewidth}|}

\cline{2-6}
     \multirow{2}{*}{} & \multirow{2}{*}{Attack($\epsilon$)}& \multicolumn{4}{c|}{Datasets} \\ \cline{3-6}
     & & MNIST& CIFAR& SVHN&Tiny\newline ImageNet \\ \hline \hline
     Clean\newline Data& - & 98.73 & 89.11 & 94.98 & 64.48 \\ \hline \hline
     \multirow{31}{*}{\shortstack{White\\box}}&\ac{fgsm}(8)& - & 14.45 & 15.06 & 12.14 \\ \cline{2-6}
     &\ac{fgsm}(16)& - & 13.66 & 5.91 & 8.11\\ \cline{2-6}
     &\ac{fgsm}(32)& 76.97 & 11.25 & - & -\\ \cline{2-6}
     &\ac{fgsm}(64)& 13.76 & - & - & -\\ \cline{2-6}
     &\ac{fgsm}(80)& 8.64 & - & - & -\\ \cline{2-6}
     &\ac{bim}(8)& - & 1.9 & 1.25 & 0.3\\ \cline{2-6}
     &\ac{bim}(16)& - & 0.61 & 0 & 0\\ \cline{2-6}
     &\ac{bim}(32)& 21.84 & - & - & -\\ \cline{2-6}
     &\ac{bim}(64)& 0 & - & - & -\\ \cline{2-6}
     &\ac{bim}(80)& 0 & - & - & -\\ \cline{2-6}
     &\ac{pgd}-$L_1$(5)& - & 43.45 & - & -\\ \cline{2-6}
     &\ac{pgd}-$L_1$(10)& 65.95 & 10.56 & - & - \\ \cline{2-6}
     &\ac{pgd}-$L_1$(15)& 25.74 & 5.27 & 17.59 & 44.7 \\ \cline{2-6}
     &\ac{pgd}-$L_1$(20)& 4.95 & - & 7.97 & 31.34 \\ \cline{2-6}
     &\ac{pgd}-$L_1$(25)& - & - &  3.73 & 21.97\\ \cline{2-6}
     &\ac{pgd}-$L_2$(0.25)& - & 13.97 & - & -\\ \cline{2-6}
     &\ac{pgd}-$L_2$(0.3125)& - & 8.19 & 35.5 & -\\ \cline{2-6}
     &\ac{pgd}-$L_2$(0.5)& - & 5.52 & 13.26 & 8.46\\ \cline{2-6}
     &\ac{pgd}-$L_2$(1)& 70.54 & - & 0.8 & 1.34\\ \cline{2-6}
     &\ac{pgd}-$L_2$(1.5)& 18.89 & - & - & -\\ \cline{2-6}
     &\ac{pgd}-$L_2$(2)& 0.79 & - & - & -\\ \cline{2-6}
     &\ac{pgd}-$L_\infty$(8)& - & 0.78 & 0.8 & 0.02\\ \cline{2-6}
     &\ac{pgd}-$L_\infty$(16)& - & 0.28 & 0 & 0\\ \cline{2-6}
     &\ac{pgd}-$L_\infty$(32)& 19.05 & - & - & -\\ \cline{2-6}
     &\ac{pgd}-$L_\infty$(64)& 0 & - & - & -\\ \cline{2-6}
     &\ac{cw}-$L_\infty$& 38.98 & 20.95 & 23.73 & 16.64\\ \cline{2-6}
     &\ac{cw}-\ac{hca}(8)& - & 46.51 & 47.06 & 39.47\\ \cline{2-6}
     &\ac{cw}-\ac{hca}(16)& - & 18.96 & 29.06 & 17.51\\ \cline{2-6}
     &\ac{cw}-\ac{hca}(80)& 43.36 & - & - & -\\ \cline{2-6}
     &\ac{cw}-\ac{hca}(128)& 8.64 & - & - & -\\ \cline{2-6}
     &\ac{df}& 4.96 & 4.8 & 6.12 & 0.52\\ \cline{2-6}
     &\ac{jsm}& 0 & 0 & 0 & 0.3\\ \cline{1-6} \cline{1-6}
     \multirow{4}{*}{\shortstack{Black\\box}}&\ac{sa}& 4.66 & 0 & 0.7 & 0.22 \\ \cline{2-6}
     &\hop & 0 & 0 & 0 & 0 \\ \cline{2-6}
     &\ac{st}& 22.04 & 52.57& 17.0 & 52.28\\ \cline{2-6}
     &\ac{pa}& 7.7 & 7.9& 9.8 & 0.5\\ \cline{1-6}
\end{tabular}
}\vspace{-4mm}
\end{table}
\subsection{Threat Model and Attacks}
Here, we define the environment that the adversary faces to generate the \acp{ae}. It is assumed that the adversary has zero-knowledge about the detection methods. Then, he might generate, using available information on the victim model, white box attacks, black box attacks and gray box attacks. We use the ART~\cite{art2018} library to generate the attacks under all tested datasets. 

\textbf{White box attacks}. Different $L_p$-norm attacks are used to test the detection methods. \ac{jsm} is used to generate $L_0$ attacks (only 1500 samples for \image dataset). For $L_1$ attacks, $L_1$ \ac{pgd} attack is used. For $L_2$ attacks, \ac{pgd}, \ac{cw}/\ac{hca} and \ac{df} $L_2$ attacks are used. For $L_\infty$ attacks, \ac{fgsm}, \ac{bim}, \ac{pgd} and \ac{cw} $L_\infty$ attacks are considered. For \ac{fgsm}, \ac{bim} and \ac{pgd} attacks, the $\epsilon=\{8,$ $16,$  $32,$ $64,$ $80,$ $128\}$ is set to each dataset as shown in Table~\ref{tab:baseline_acc}. For \ac{cw} attack, 200 iterations and zero confidence setting are used.

\textbf{Black box attacks}. \ac{pa}~\cite{su2019one}, \ac{sa}~\cite{andriushchenko2020square}, \hop~\cite{chen2020hopskipjumpattack} and \ac{st}~\cite{engstrom2019exploring} black box attacks are generated in the testing process. The translation and rotation values of \ac{st} attack are set to 10 and 60 for MNIST and SVHN, and to 8 and 30 for CIFAR and \image, respectively. For \ac{sa} attack, the epsilon ($\epsilon$) is set to $32-80$ out of 255. For \hop attack, untargeted and unmasked attack is considered, besides, 40 and 100 are set for iterations steps and maximum evaluations, respectively. For \ac{pa} attacks, only 1000 \acp{ae} for each dataset are generated.

\textbf{Gray box attacks}. In order to evaluate the detection methods against gray box attacks, we built surrogate models of the baseline classifiers. For MNIST, SVHN and CIFAR10, surrogate classifiers are similar to victim classifiers with only one change that is a Dropout layer is added before the last Dense layer. For MNIST, the classification accuracy is 99.32\%, for SVHN the classification accuracy is 95.48\%, and for CIFAR10 the classification accuracy is 93.35\%. For \image, ResNet50V2~\cite{he2016identity} classifier is fine-tuned and it achieves accuracy of 51.7\%.  Then, the white box attacks are generated under the surrogate classifiers. 

\textbf{Untargeted Attack.} All the tested attacks in this work are untargeted attacks. It was shown that untargeted attacks 1) have less perturbations than targeted attacks 2) have better success rates, and 3) possess stronger transferability capability~\cite{carlini2017towards,liu2016delving}.  

Table~\ref{tab:baseline_acc} shows the baseline classifiers' accuracy to the clean training data and the tested attacked data.

\textbf{Robust attacks}. As shown in \cite{carlini2017adversarial,athalye2018obfuscated}, detectors and defenses can be bypassed using different strategies such as; strong attacks, unknown attacks, or circumventing the detectors, especially for differential-based\linebreak classifiers/detectors. In this work, we considered the method that is described in \cite{athalye2018obfuscated} to break \ac{lid} detector. The goal is to generate high confidence attacks \cite{carlini2017adversarial} that minimize the $L_\infty$-norm under the zero-knowledge of the deployed defense.

\subsection{Detection Methods}
In our experiments, we retest all the following detection methods under the same environment. We choose to assess the performance of 8 different detection methods from different categories that are popular in the research communities and their source code are publicly available. We build a testing environment that combines all the tested methods and is publicly available\footnote{The code is available at: \url{https://github.com/aldahdooh/detectors_review}}. In this subsection, a technical description for each detector is provided with their configurations and hyper-parameters.


\subsubsection{\ac{kd}+\ac{bu}~\cite{feinman2017detecting}:}
\textbf{The detector:} For given clean and adversarial examples, noisy samples are crafted to be treated as clear samples. Then,  Bayesian uncertainty scores is computed using Monte Carlo Dropout estimation for clean, noisy, and adversarial samples. Kernel density is estimated for each class using Gaussian estimation with specific bandwidth and then, kernel density scores are computed for clean, noisy and adversarial samples. Finally, a logistic regression classifier is trained using uncertainty and density scores.\\

\noindent\textbf{Settings:} The following settings are considered: 
\begin{itemize}

    \item To generate noisy samples: Gaussian noise is added to each pixel with a \textit{scale} that is set to be equal to the \hbox{$L_p$--norm} of the adversarial perturbation. For $L_0$ attacks, an equal number of pixels that were altered in the adversarial example are flipped randomly. 
    \item For kernel density estimation: the bandwidth parameter is tuned and set to be 1.26, 0.26, 1 and 0.26 for MNIST, CIFAR, SVHN and \image \ datasets, respectively. 
    \item 70\% of testing data is used to train the detector and 30\% of testing data is used to test the detector. 
    \item The adversarial examples for both training and testing were generated by applying one of the attacks at a time. 
\end{itemize}

\subsubsection{\ac{lid}~\cite{ma2018characterizing}:}
\textbf{The detector:} For given clean and adversarial examples, noisy samples are crafted to be treated as clear samples. Then, for each sample the LID score is calculated for $k$-neighbor samples. Finally, a logistic regression classifier is trained using LID scores.\\

\noindent\textbf{Settings:} The following settings are considered: 
\begin{itemize}

    \item To generate noisy samples: same procedures of \ac{kd}+\ac{bu} are applied. 
    \item The $k$-neighbor is set to 20 for MNIST and SVHN, and 30 for CIFAR and \image .
    \item  70\% of testing data is used to train the detector and 30\% of testing data is used to test the detector. 
    \item The adversarial examples for both training and testing were generated by applying one of the attacks at a time.
\end{itemize}

\subsubsection{\ac{nss}~\cite{kherchouche2020detection}:}
\textbf{The detector:} For given clean samples, \acp{ae} are generated using \ac{pgd} attack with different perturbation radius $\epsilon$. For the clean and the adversarial examples, 18 features from the asymmetric generalized distribution (AGGD) are then computed using mean subtracted contrast normalized (MSCN) coefficients. Finally,  a binary \ac{svm} classifier is trained using these 18 features.\\

\noindent\textbf{Settings:} The following settings are considered: 
\begin{itemize}

    \item To generate the \acp{ae} using \ac{pgd} attack: we divide the clean testing data into six groups and the \ac{pgd}-based \acp{ae} for each group are generated using $\epsilon=\{0.03125,$ $0.0625,$ $0.125,$ $0.25,$ $0.3125,$ $0.5\}$, respectively. 
    \item For the \ac{svm} classifier, the regularization parameter $C$ and the kernel coefficient $g$/\textit{gamma} of the classifier are tuned using grid search.
\end{itemize}

\subsubsection{\acs{fs}~\cite{xu2017feature}:}
\textbf{The detector:} Once squeezers are defined, squeezed clean samples are generated. The maximum $L_1$ or $L_2$ distance is computed between the prediction probabilities of clean and squeezed samples. Finally, the threshold value with a specific false positive rate is computed.\\

\noindent\textbf{Settings:} The following settings are considered: 
\begin{itemize}

    \item Squeezers: 1) Color bit depth reduction. 2) Median smoothing filter 3) Non-local mean denoiser. 
    \begin{itemize}
    \item Color bit depth reduction: 1-bit for MNIST and 5-bit for SVHN, CIFAR, and \image 
    \item Median smoothing filter: $2\times 2$ size filter for all datasets. 
    \item Non-local mean denoiser: search window = 13, block size = 3,  and filter strength for luminance component = 2 for all datasets except MNIST. For MNIST dataset, the non-local mean denoiser is not used. 
    \end{itemize}
    \item Detector training: 50\% of test data is used for detector training. 
    \item False positive rate: 5\%
\end{itemize}

\subsubsection{MagNet~\cite{meng2017magnet}:}
\textbf{The detector:} Here, we demonstrate the detection process only of MagNet without the defense process. For given clean training samples, one or more autoencoders are trained. For a given clean validation data, calculate the $L_1$ reconstruction error using the autoencoders. Then, for each autoencoder, calculate the threshold value from the calculated reconstruction errors with a specific false positive rate.\\

\noindent\textbf{Settings:} The following settings are considered: 
\begin{itemize}
    \item Detector Autoencoders: two detectors are used
    \begin{itemize}
    \item The first autoencoder structure: [Conv2D($3\times 3$), average pooling, Conv2D($3\times 3$), Conv2D($3\times 3$), up sampling, Conv2D($3\times 3$)]. 
    \item The second autoencoder structure: [Conv2D($3\times 3$), Conv2D($3\times 3$)]. 
    \end{itemize}
    \item 5000 samples from clean training samples are dedicated for validation process. 
    \item False positive rate: 1\% for MNIST and 5\% for other datasets. 
    \item We report only the results of the detector without taking into consideration the defense part, i.e., classification accuracy after applying the reformer. Please note that the original paper report the overall performance of the detection and the defense  
\end{itemize}

\subsubsection{\ac{dnr}~\cite{sotgiu2020deep}:}
\textbf{The detector:} For given clean training samples, train three image classification classifiers using \ac{rbf}-\ac{svm}. Each classifier receives, as input, the feature map(s) of a specific baseline classifier layer(s).  Train a fourth image classification classifier using \ac{rbf}-\ac{svm}. The classifier takes, as input, the prediction probabilities of the three classifiers trained in the first step. Given clean testing samples, get the maximum prediction probabilities and then calculate the threshold value from prediction probabilities for a given false positive rate.\\

\noindent\textbf{Settings:} The following settings are considered: 
\begin{itemize}

    \item \resizebox{\linewidth}{!}{%
            \begin{tabular}{|m{0.24\linewidth}|m{0.14\linewidth}|m{0.14\linewidth}|m{0.35\linewidth}|}
                  \hline
                  Input of & MNIST, SVHN & CIFAR & \image  \\\hline
                  $1^{st}$ classifier & Layer 4 & Layer 7 & Layer pool4\_bn \\\hline
                  $2^{nd}$ classifier & Layer 5 & Layer 8 & Layer conv5\_block17\_0\_bn \\\hline
                  $3^{rd}$ classifier & Layer 6 & Layer 9 & Layer bn \\ \hline
             \end{tabular} 
             }%
    \item See Tables \ref{tab:mnist_arch}-\ref{tab:cifar10_arch} for Layer numbers. 
    \item For the \ac{svm} classifiers, the regularization parameter $C$ is set to 1 and the kernel coefficient $\gamma$ is set to \textit{scale}, where $scale=\frac{1}{F.V}$, $F$ is the number of features, and $V$ is the variance of the inputs
    \item False positive rate: 10\% 
\end{itemize}

\subsubsection{\ac{sfad}~\cite{aldahdooh2021selective}:}
\textbf{The detector:} For given clean training samples, three image classification classifiers are trained using SelectiveNet as described in Section \ref{sec:unsuper_auxil}. Each classifier receives, as input, the feature map(s) of a specific baseline classifier layer(s). The feature maps are processed during the training using autoencoding, up/down sampling, bottleneck, and noise blocks.  A fourth image classification classifier is trained using SelectiveNet. The classifier takes, as input, the prediction probabilities of the three classifiers trained in the first step. Given clean testing samples, get the maximum prediction probabilities, and the selective probabilities. Finally, threshold values are computed from probabilities of the three classifiers for a given false positive rate.\\

\noindent\textbf{Settings:} The following settings are considered: 
\begin{itemize}

    \item Classifiers' inputs are the same as \ac{dnr} detector. 
    \item The \{coverage, coverage threshold\} for SelectiveNet classifiers are set to: 
    \item \resizebox{\linewidth}{!}{%
            \begin{tabular}{|m{0.3\linewidth}|m{0.2\linewidth}|m{0.18\linewidth}|m{0.22\linewidth}|}
                  \hline
                  SelectiveNet & MNIST,  & CIFAR, SVHN & \image  \\\hline
                  $1^{st}$, $2^{nd}$, and $3^{rd}$ classifiers & \{1, 0.995\} & \{0.9, 0.9\} & \{0.8, 0.5\} \\\hline
                  $4^{th}$ classifier & \{1, 0.7\} & \{0.9, 0.7\} & \{0.8, 0.5\} \\\hline
            \end{tabular}  
            }%
    \item False positive rate: 10\% 
\end{itemize}

\subsubsection{\ac{nic}~\cite{ma2019nic}:}
\textbf{The detector:} For given clean training samples and for each layer in the baseline model, get the feature \linebreak map/`provenance invariant (PI)'. Train PI classifier, \linebreak OneClassSVM classifier, for each layer using layer feature's map.  Get the prediction probabilities/'activation values invariant (VI)' for the current layer and the next layer.  Train VI classifier, OneClassSVM classifier, for each layer using the prediction probabilities. Finally,  train \ac{nic} classifier, \linebreak OneClassSVM classifier, using the decision values of all PIs and VIs classifiers.\\

\noindent\textbf{Settings:} The following settings are considered: 
\begin{itemize}
    \item For provenance invariant channel classifiers, we use the first 5000 PCA components as features if the layer has more than 5000 features and we use $\nu=0.01$ and $\gamma=1$ for the OneClassSVM classifiers of all layers. 
    \item For the activation value invariant channel and for the final \ac{nic} classifiers, we use $\nu=0.1$ and $\gamma=scale$, where $scale=\frac{1}{F.V}$, $F$ is the number of features, and $V$ is the variance of the inputs, for the OneClassSVM classifiers for all layers.  
\end{itemize}

\subsection{Performance measures}
As discussed in Section \ref{sec: detection_methods}, there are many criteria to assess the performance of detectors. In our experiment, we use detection rate (DR) and false positive rate (FPR) as two main performance evaluations. Other performance evaluation measures, like complexity (CM), overhead (OV) and inference time latency (INF) will be discussed as well. 

\section{Results and discussions} \label{sec: results}

\begin{figure*}[t!]
    \begin{center}
         \includegraphics[width=\linewidth, keepaspectratio]{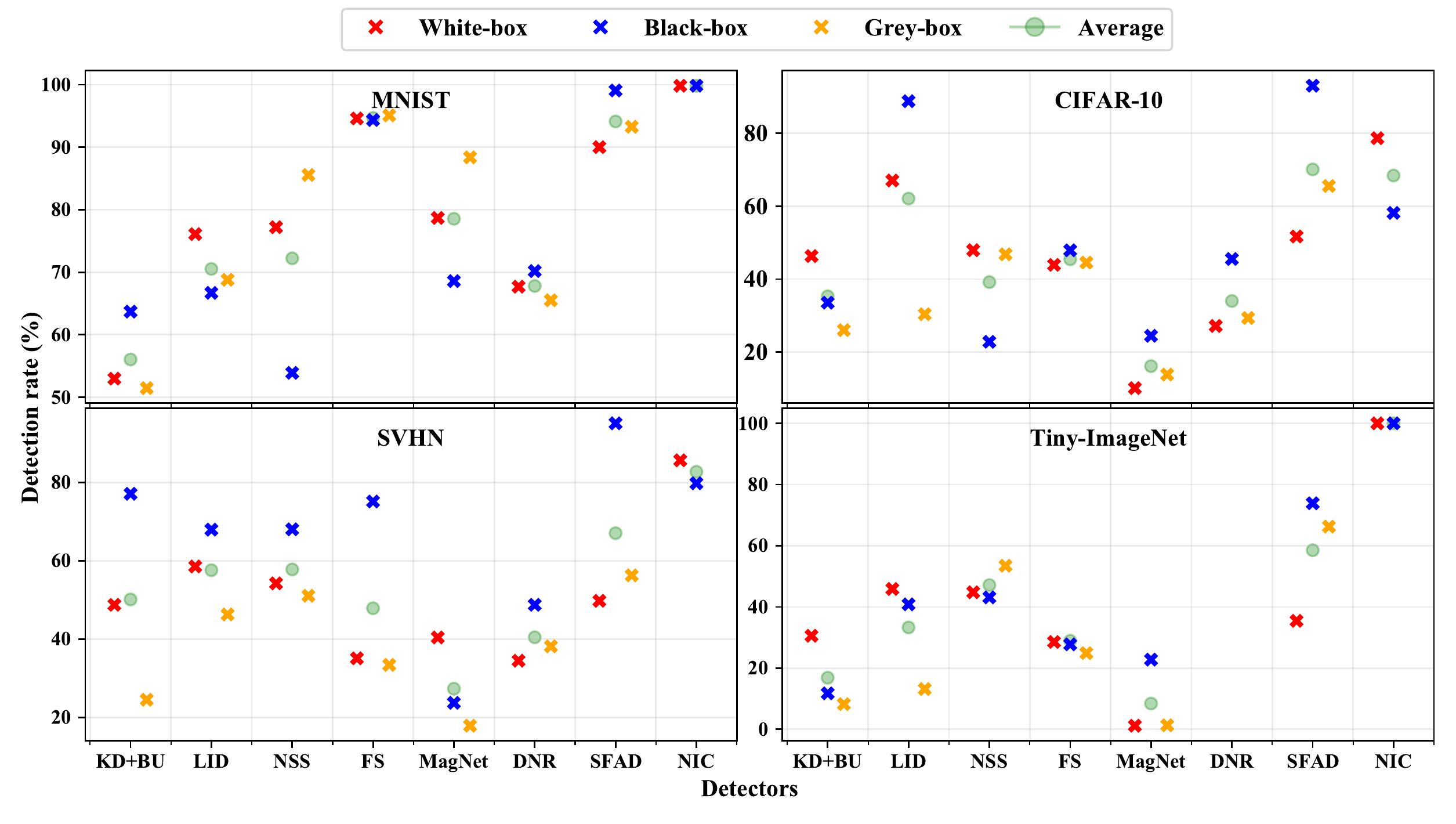}\vspace{-4mm}
    \end{center}
    \caption{The average detection rate of eight detectors assessed against white-, black- and gray-box attacks scenarios. The green points represent the average over all scenarios.} \vspace{-3mm}
    \label{fig:summary}
\end{figure*}

In this section, we evaluate the performance of the detection methods on different datasets against different types of successful attack scenarios, white, black and gray box attacks. Tables \ref{tab:white_mnist}, \ref{tab:white_cifar}, \ref{tab:white_svhn} and \ref{tab:white_tiny} show detection rate and \ac{fpr} of the tested detectors on MNIST, CIFAR, SVHN and Tiny-ImagNet datasets, respectively, against white box attacks. Tables \ref{tab:black_mnist}, \ref{tab:black_cifar}, \ref{tab:black_svhn} and \ref{tab:black_tiny} show results for black box attacks, while Tables \ref{tab:gray_mnist}, \ref{tab:gray_cifar}, \ref{tab:gray_svhn} and \ref{tab:gray_tiny} show results for gray box attacks. The summary of all the experiments is shown in Figure \ref{fig:summary}. 

In the following subsections, for each detection method, the performance results will be discussed for each attack scenario.  

\subsection{\ac{kd}+\ac{bu}~\cite{feinman2017detecting}}
\textbf{White box attacks}. In general, the detector jointly combines density and uncertainty estimations. Basically, it is a two-feature binary classifier. \acp{ae} with low density probability and high uncertainty will be easily detected and that is obvious for \ac{bim} and \ac{pgd}-$L_\infty$ attacks on CIFAR, SVHN and \image, but not for other attacks in which the detector has medium to poor performance. The detector is performing well against \ac{jsm} attacks on simple datasets like MNIST and SVHN, while it is not performing well on CIFAR and \image . Besides, for \image \ dataset the detector fails to learn against \ac{pgd}-$L_1$, \ac{pgd}-$L_2(0.5)$ and \ac{df}. One reason for this failure is due to the density bandwidth which is not appropriate for the \image \ dataset.  
Moreover, it seems that the model uncertainty of \acp{ae} is less than that of clean samples, hence the uncertainty measure, i.e., Dropout technique, is not good enough to split between them. Applying better uncertainty measures will definitely enhance the detector. On the other hand, relying on density estimation requires clean enough data and it is not appropriate for small noisy data. In terms of \ac{fpr}, the detector reaches \ac{fpr} in the interval between 0.0-17.09\% depending on the \acp{ae} used in the training process.

\textbf{Black box attacks}. The detector is not very effective for detecting black box attacks. We can interpret the low accuracy of \ac{st} attack since the baseline classifiers are trained with data augmentation and hence, low uncertainty values are estimated for \ac{st} attack. \hop is a boundary-decision based attack, we run it with 40 steps, and we expect from the \ac{kd}+\ac{bu} detector to have effective performance against \hop attack. We get that for SVHN and MNIST, but not for CIFAR and \image, this is due to the high dimension of \image \ and the fact that the uncertainty of the attack remains low for clean samples after many iterations of the attack. Latter reason applies to the \ac{sa} attack as well. The detector has an acceptable performance against \ac{pa} attack on MNIST and SVHN, while has poor performance on CIFAR and \image .

\textbf{Gray box attacks}. The detector performance against gray box attacks is comparable to the performance of white box attacks except, in general, for \ac{bim}, \ac{pgd}-$L_\infty$ and \ac{cw} in which it became worse compared to white box attacks scenario. It means that the transferable attacks keep their characteristics of having high density probability and low uncertainty values and hence, the detector's uncertainty and density estimations are not good enough.

\begin{table*}[!htb]
\caption{Detection rates (DR\%) and false positive rate (FPR\%) for the detectors against tested white box attacks($\epsilon$) on MNIST. Top 3 are colored with \textcolor{red}{red}, \textcolor{blue}{blue} and \textcolor{customgreen}{green},  respectively. }
\label{tab:white_mnist}
\resizebox{\linewidth}{!}{
\begin{tabular}{|P{0.12\linewidth}|M{0.039\linewidth}|M{0.039\linewidth}||M{0.039\linewidth}|M{0.039\linewidth}||M{0.039\linewidth}|M{0.039\linewidth}|||M{0.039\linewidth}|M{0.039\linewidth}||M{0.039\linewidth}|M{0.039\linewidth}||M{0.039\linewidth}|M{0.039\linewidth}||M{0.039\linewidth}|M{0.039\linewidth}||M{0.039\linewidth}|M{0.039\linewidth}|}
\hline
     \multirow{3}{*}{Attack}& \multicolumn{6}{c|||}{Supervised Detectors} & \multicolumn{10}{c|}{Unsupervised Detectors}\\ \cline{2-17}
     &\multicolumn{2}{c||}{\ac{kd}+\ac{bu}~\cite{feinman2017detecting}}& \multicolumn{2}{c||}{\ac{lid}~\cite{ma2018characterizing}} & \multicolumn{2}{c|||}{\ac{nss}~\cite{kherchouche2020detection}}& \multicolumn{2}{c||}{\acs{fs}~\cite{xu2017feature}}& \multicolumn{2}{c||}{MagNet~\cite{meng2017magnet}} & \multicolumn{2}{c||}{\ac{dnr}~\cite{sotgiu2020deep}} & \multicolumn{2}{c||}{\ac{sfad}~\cite{aldahdooh2021selective}} & \multicolumn{2}{c|}{\ac{nic}~\cite{ma2019nic}} \\ \cline{2-17}
     &DR & \ac{fpr} & DR & \ac{fpr} & DR & \ac{fpr} & DR & \ac{fpr} & DR & \ac{fpr} & DR & \ac{fpr} & DR & \ac{fpr} & DR & \ac{fpr} \\ \hline \hline
     \ac{fgsm}(32)& 85.54&3.46&81.66&1.41&\textcolor{red}{100.0}&0.0&97.8&5.27&\textcolor{blue}{100.0}&0.2&59.28&10.01&97.76&10.79&\textcolor{customgreen}{100}&10.12\\ \hline
     \ac{fgsm}(64)& 53.8&0.64&77.09&0.07&\textcolor{red}{100.0}&0.0&98.06&5.27&\textcolor{blue}{100.0}&0.2&87.81&10.01&98.74&10.79&\textcolor{customgreen}{100}&10.12\\ \hline
     \ac{fgsm}(80)& 49.03&0.17&73.64&0.07&\textcolor{red}{100.0}&0.0&98.02&5.27&\textcolor{blue}{100.0}&0.2&91.91&10.01&99.47&10.79&\textcolor{customgreen}{100}&10.12\\ \hline
     \ac{bim}(32)& 58.66&3.77&80.06&0.94&\textcolor{red}{100.0}&0.0&99.18&5.27&\textcolor{blue}{100.0}&0.2&67.19&10.01&93.81&10.79&\textcolor{customgreen}{99.46}&10.12\\ \hline
     \ac{bim}(64)& 48.2&5.52&74.71&0.4&\textcolor{red}{100.0}&0.0&95.08&5.27&\textcolor{blue}{100.0}&0.2&51.24&10.01&70.34&10.79&\textcolor{customgreen}{99.99}&10.12\\ \hline
     \ac{bim}(80)& 80.76&2.09&77.63&0.57&\textcolor{red}{100.0}&0.0&89.73&5.27&\textcolor{blue}{100.0}&0.2&62.99&10.01&66.53&10.79&\textcolor{customgreen}{100}&10.12\\\hline
     \ac{pgd}-$L_1$(10)& 71.34&3.03&77.71&2.49&65.32&0.0&\textcolor{blue}{97.8}&5.27&5.0&0.2&57.56&10.01&\textcolor{customgreen}{95.66}&10.79&\textcolor{red}{100}&10.12\\ \hline
     \ac{pgd}-$L_1$(15)& 52.41&3.4&73.87&2.19&\textcolor{customgreen}{88.61}&0.0&\textcolor{blue}{94.56}&5.27&51.51&0.2&56.57&10.01&88.3&10.79&\textcolor{red}{99.99}&10.12\\ \hline
     \ac{pgd}-$L_1$(20)& 28.16&3.46&65.96&1.61&\textcolor{blue}{98.05}&0.0&88.1&5.27&\textcolor{customgreen}{94.55}&0.2&49.47&10.01&78.18&10.79&\textcolor{red}{98.77}&10.12\\ \hline
     \ac{pgd}-$L_2$(1)& 73.87&2.69&81.27&2.93&62.36&0.0&\textcolor{blue}{98.07}&5.27&11.0&0.2&54.52&10.01&\textcolor{customgreen}{96.42}&10.79&\textcolor{red}{100}&10.12\\ \hline
     \ac{pgd}-$L_2$(1.5)& 0.04&0.0&59.34&3.46&88.51&0.0&\textcolor{blue}{96.07}&5.27&60.79&0.2&57.83&10.01&\textcolor{customgreen}{89.61}&10.79&\textcolor{red}{100}&10.12\\ \hline
     \ac{pgd}-$L_2$(2)& 0.71&0.07&57.41&1.45&\textcolor{blue}{98.63}&0.0&85.58&5.27&\textcolor{customgreen}{93.31}&0.2&48.21&10.01&75.44&10.79&\textcolor{red}{100}&10.12\\ \hline
     \ac{pgd}-$L_\infty$(32)& 55.32&3.97&79.04&0.98&\textcolor{red}{100.0}&0.0&99.2&5.27&\textcolor{blue}{100.0}&0.2&67.13&10.01&93.47&10.79&\textcolor{customgreen}{100}&10.12\\ \hline
     \ac{pgd}-$L_\infty$(64)& 49.71&5.55&75.01&0.4&\textcolor{red}{100.0}&0.0&95.18&5.27&\textcolor{blue}{100.0}&0.2&51.28&10.01&70.18&10.79&\textcolor{customgreen}{100}&10.12\\ \hline
     \ac{cw}-$L_\infty$& 42.77&0.71&64.43&3.94&2.47&0.0&\textcolor{blue}{98.41}&5.27&40.56&0.2&57.98&10.01&\textcolor{customgreen}{98.24}&10.79&\textcolor{red}{100}&10.12\\ \hline
     \ac{cw}-\ac{hca}(80)& 32.52&1.78&69.17&0.17&58.62&0.0&\textcolor{blue}{100.0}&5.27&\textcolor{red}{100.0}&0.2&79.79&10.01&98.71&10.79&\textcolor{customgreen}{100}&10.12\\ \hline
     \ac{cw}-\ac{hca}(128)& 86.0&3.13&99.85&0.0&5.9&0.0&99.98&5.27&\textcolor{red}{100.0}&0.2&\textcolor{blue}{100.0}&10.01&100.0&10.79&\textcolor{customgreen}{100}&10.12\\ \hline
     \ac{df}& 48.97&0.37&93.3&0.1&\textcolor{customgreen}{98.5}&0.0&66.96&5.27&96.99&0.2&95.6&10.01&\textcolor{blue}{99.58}&10.79&\textcolor{red}{100}&10.12\\ \hline
     \color{red}\ac{jsm}& 88.56&0.3&84.73&0.27&0.01&0&\textcolor{blue}{99.88}&5.27&41.47&0.2&80.48&10.01&\textcolor{customgreen}{99.88}&10.79&\textcolor{red}{100}&10.12\\ \hline\hline
     \color{red}\textbf{Average}& \textbf{52.97}&2.32&\textbf{76.10}&1.23&\textbf{77.21}&0&\textcolor{blue}{\textbf{94.61}}&5.27&\textbf{78.69}&0.2&\textbf{67.2}&10.01&\textcolor{customgreen}{\textbf{90.02}}&10.79&\textcolor{red}{\textbf{99.84}}&10.12\\ \hline 
\end{tabular}
}
\end{table*}

\begin{table*}[!htb]
\caption{Detection rates (DR\%) and false positive rate (FPR\%) for the detectors against tested white box attacks($\epsilon$) on CIFAR-10. Top 3 are colored with \textcolor{red}{red}, \textcolor{blue}{blue} and \textcolor{customgreen}{green},  respectively. }
\label{tab:white_cifar}
\resizebox{\linewidth}{!}{
\begin{tabular}{|P{0.12\linewidth}|M{0.039\linewidth}|M{0.039\linewidth}||M{0.039\linewidth}|M{0.039\linewidth}||M{0.039\linewidth}|M{0.039\linewidth}|||M{0.039\linewidth}|M{0.039\linewidth}||M{0.039\linewidth}|M{0.039\linewidth}||M{0.039\linewidth}|M{0.039\linewidth}||M{0.039\linewidth}|M{0.039\linewidth}||M{0.039\linewidth}|M{0.039\linewidth}|}
\hline
     \multirow{3}{*}{Attack}& \multicolumn{6}{c|||}{Supervised Detectors} & \multicolumn{10}{c|}{Unsupervised Detectors}\\ \cline{2-17}
     &\multicolumn{2}{c||}{\ac{kd}+\ac{bu}~\cite{feinman2017detecting}}& \multicolumn{2}{c||}{\ac{lid}~\cite{ma2018characterizing}} & \multicolumn{2}{c|||}{\ac{nss}~\cite{kherchouche2020detection}}& \multicolumn{2}{c||}{\acs{fs}~\cite{xu2017feature}}& \multicolumn{2}{c||}{MagNet~\cite{meng2017magnet}} & \multicolumn{2}{c||}{\ac{dnr}~\cite{sotgiu2020deep}} & \multicolumn{2}{c||}{\ac{sfad}~\cite{aldahdooh2021selective}} & \multicolumn{2}{c|}{\ac{nic}~\cite{ma2019nic}} \\ \cline{2-17}
     &DR & \ac{fpr} & DR & \ac{fpr} & DR & \ac{fpr} & DR & \ac{fpr} & DR & \ac{fpr} & DR & \ac{fpr} & DR & \ac{fpr} & DR & \ac{fpr} \\ \hline \hline
     \ac{fgsm}(8)& 35.03&7.3&\textcolor{customgreen}{53.0}&3.84&\textcolor{red}{87.59}&6.56&29.33&5.07&0.72&0.77&32.09&10.01&\textcolor{blue}{67.94}&10.9&43.64&10.08\\ \hline
     \ac{fgsm}(16)& 33.23&4.5&\textcolor{blue}{81.23}&1.44&\textcolor{red}{99.94}&6.56&35.34&5.07&3.11&0.77&31.35&10.01&\textcolor{customgreen}{79.9}&10.9&58.48&10.08\\ \hline
     \ac{fgsm}(32)& 0.08&0.04&\textcolor{customgreen}{94.23}&0.11&\textcolor{blue}{99.67}&6.56&32.83&5.07&\textcolor{red}{100.0}&0.77&27.24&10.01&92.58&10.9&87.32&10.08\\ \hline
     \ac{bim}(8)& \textcolor{customgreen}{84.47}&2.18&\textcolor{blue}{88.05}&3.65&52.16&6.56&8.74&5.07&0.56&0.77&4.27&10.01&18.12&10.9&\textcolor{red}{99.95}&10.08\\ \hline
     \ac{bim}(16)& \textcolor{blue}{99.55}&0.07&\textcolor{customgreen}{98.55}&0.44&87.74&6.56&0.34&5.07&0.69&0.77&17.07&10.01&45.35&10.9&\textcolor{red}{100}&10.08\\ \hline
     \ac{pgd}-$L_1$(5)& 51.96&7.12&0.0&0.0&5.32&6.56&\textcolor{red}{75.61}&5.07&0.4&0.77&38.66&10.01&\textcolor{blue}{66.06}&10.9&\textcolor{customgreen}{56.12}&10.08\\ \hline
     \ac{pgd}-$L_1$(10)& 9.67&1.81&\textcolor{customgreen}{48.18}&24.71&8.02&6.56&\textcolor{blue}{70.7}&5.07&0.61&0.77&28.92&10.01&30.34&10.9&\textcolor{red}{74.57}&10.08\\ \hline
     \ac{pgd}-$L_1$(15)& 34.48&10.14&\textcolor{blue}{69.43}&21.1&11.38&6.56&\textcolor{customgreen}{56.61}&5.07&0.68&0.77&18.07&10.01&13.7&10.9&\textcolor{red}{92.32}&10.08\\ \hline
     \ac{pgd}-$L_2$(0.25)& \textcolor{customgreen}{36.99}&6.79&30.53&16.78&7.38&6.56&\textcolor{red}{73.59}&5.07&0.55&0.77&30.49&10.01&34.95&10.9&\textcolor{blue}{72.18}&10.08\\ \hline
     \ac{pgd}-$L_2$(0.3125)& 0.12&0.11&\textcolor{customgreen}{51.83}&23.39&8.75&6.56&\textcolor{blue}{67.14}&5.07&0.62&0.77&26.12&10.01&24.08&10.9&\textcolor{red}{89.1}&10.08\\ \hline
     \ac{pgd}-$L_2$(0.5)& \textcolor{customgreen}{55.01}&9.26&\textcolor{blue}{77.97}&17.71&13.72&6.56&45.36&5.07&0.7&0.77&10.65&10.01&10.95&10.9&\textcolor{red}{97.21}&10.08\\ \hline
     \ac{pgd}-$L_\infty$(8)& \textcolor{customgreen}{92.27}&0.96&\textcolor{blue}{94.39}&1.81&57.06&6.56&8.2&5.07&0.57&0.77&11.34&10.01&29.49&10.9&\textcolor{red}{100}&10.08\\ \hline
     \ac{pgd}-$L_\infty$(16)& \textcolor{blue}{99.89}&0.0&\textcolor{customgreen}{99.22}&0.26&93.24&6.56&0.2&5.07&0.66&0.77&25.11&10.01&52.9&10.9&\textcolor{red}{100}&10.08\\ \hline
     \ac{cw}-$L_\infty$& 21.12&4.54&\textcolor{blue}{64.52}&20.58&27.48&6.56&56.18&5.07&13.23&0.77&44.15&10.01&\textcolor{red}{87.68}&10.9&\textcolor{customgreen}{61.68}&10.08\\ \hline
     \ac{cw}-\ac{hca}(8)& 37.29&6.34&44.59&15.01&40.94&6.56&\textcolor{blue}{68.33}&5.07&0.61&0.77&34.91&10.01&\textcolor{customgreen}{57.76}&10.9&\textcolor{red}{75.18}&10.08\\ \hline
     \ac{cw}-\ac{hca}(16)& 29.69&3.95&\textcolor{blue}{65.46}&19.25&\textcolor{customgreen}{65.12}&6.56&44.28&5.07&0.44&0.77&27.79&10.01&33.94&10.9&\textcolor{red}{71.39}&10.08\\ \hline
     \ac{df}& 54.02&1.44&\textcolor{customgreen}{63.57}&6.12&50.15&6.56&39.18&5.07&57.33&0.77&30.2&10.01&\textcolor{red}{89.57}&10.9&\textcolor{blue}{84.91}&10.08\\ \hline
     \color{red}\ac{jsm}& 58.95&4.32&\textcolor{blue}{82.26}&3.69&47.11&6.56&\textcolor{customgreen}{78.18}&5.07&0.5&0.77&53.32&10.01&\textcolor{red}{95.02}&10.9&51.64&10.08\\ \hline\hline
     \textbf{Average}& \textbf{46.32}&3.94&\textcolor{blue}{\textbf{67.06}}&9.99&\textbf{47.93}&6.56&\textbf{43.9}&5.07&\textbf{10.11}&0.77&\textbf{27.15}&10.01&\textcolor{customgreen}{\textbf{51.69}}&10.9&\textcolor{red}{\textbf{78.65}}&10.08\\ \hline
\end{tabular}
}
\end{table*}

\begin{table*}[!htb]
\caption{Detection rates (DR\%) and false positive rate (FPR\%) for the detectors against tested white box attacks($\epsilon$) on SVHN. Top 3 are colored with \textcolor{red}{red}, \textcolor{blue}{blue} and \textcolor{customgreen}{green},  respectively. }
\label{tab:white_svhn}
\resizebox{\linewidth}{!}{
\begin{tabular}{|P{0.12\linewidth}|M{0.039\linewidth}|M{0.039\linewidth}||M{0.039\linewidth}|M{0.039\linewidth}||M{0.039\linewidth}|M{0.039\linewidth}|||M{0.039\linewidth}|M{0.039\linewidth}||M{0.039\linewidth}|M{0.039\linewidth}||M{0.039\linewidth}|M{0.039\linewidth}||M{0.039\linewidth}|M{0.039\linewidth}||M{0.039\linewidth}|M{0.039\linewidth}|}
\hline
     \multirow{3}{*}{Attack}& \multicolumn{6}{c|||}{Supervised Detectors} & \multicolumn{10}{c|}{Unsupervised Detectors}\\ \cline{2-17}
     &\multicolumn{2}{c||}{\ac{kd}+\ac{bu}~\cite{feinman2017detecting}}& \multicolumn{2}{c||}{\ac{lid}~\cite{ma2018characterizing}} & \multicolumn{2}{c|||}{\ac{nss}~\cite{kherchouche2020detection}}& \multicolumn{2}{c||}{\acs{fs}~\cite{xu2017feature}}& \multicolumn{2}{c||}{MagNet~\cite{meng2017magnet}} & \multicolumn{2}{c||}{\ac{dnr}~\cite{sotgiu2020deep}} & \multicolumn{2}{c||}{\ac{sfad}~\cite{aldahdooh2021selective}} & \multicolumn{2}{c|}{\ac{nic}~\cite{ma2019nic}} \\ \cline{2-17}
     &DR & \ac{fpr} & DR & \ac{fpr} & DR & \ac{fpr} & DR & \ac{fpr} & DR & \ac{fpr} & DR & \ac{fpr} & DR & \ac{fpr} & DR & \ac{fpr} \\ \hline \hline
     \ac{fgsm}(8)& 46.75&10.02&\textcolor{customgreen}{72.18}&7.37&\textcolor{red}{98.95}&0.54&47.5&5.1&8.57&0.49&45.45&10.0&\textcolor{blue}{81.26}&11.02&67.35&9.99\\ \hline
     \ac{fgsm}(16)& 44.3&9.28&\textcolor{blue}{89.79}&3.47&\textcolor{red}{99.85}&0.54&51.88&5.1&18.75&0.49&50.63&10.0&\textcolor{customgreen}{88.57}&11.02&59.86&9.99\\ \hline
     \ac{bim}(8)& 49.49&11.01&52.38&11.01&\textcolor{blue}{92.08}&0.54&11.71&5.1&\textcolor{customgreen}{54.29}&0.49&24.8&10.0&26.07&11.02&\textcolor{red}{92.91}&9.99\\ \hline
     \ac{bim}(16)& \textcolor{customgreen}{93.64}&2.79&86.64&5.4&\textcolor{blue}{99.85}&0.54&0.73&5.1&88.08&0.49&14.74&10.0&14.22&11.02&\textcolor{red}{99.96}&9.99\\ \hline
     \ac{pgd}-$L_1$(15)& 9.84&6.97&43.03&19.99&0.48&0.54&\textcolor{customgreen}{43.32}&5.1&20.43&0.49&36.9&10.0&\textcolor{blue}{46.9}&11.02&\textcolor{red}{91.99}&9.99\\ \hline
     \ac{pgd}-$L_1$(20)& 22.24&14.22&\textcolor{blue}{48.8}&19.75&0.59&0.54&30.79&5.1&32.03&0.49&34.64&10.0&\textcolor{customgreen}{37.62}&11.02&\textcolor{red}{88.7}&9.99\\ \hline
     \ac{pgd}-$L_1$(25)& \textcolor{customgreen}{42.67}&17.09&\textcolor{blue}{53.67}&18.86&0.78&0.54&21.62&5.1&41.71&0.49&30.16&10.0&31.06&11.02&\textcolor{red}{93.45}&9.99\\ \hline
     \ac{pgd}-$L_2$(0.3125)& 20.4&5.26&22.88&9.13&0.38&0.54&\textcolor{customgreen}{59.33}&5.1&7.73&0.49&37.34&10.0&\textcolor{blue}{60.69}&11.02&\textcolor{red}{71.26}&9.99\\ \hline
     \ac{pgd}-$L_2$(0.5)& 10.99&9.26&\textcolor{blue}{46.48}&20.56&0.53&0.54&37.59&5.1&24.86&0.49&35.41&10.0&\textcolor{customgreen}{42.13}&11.02&\textcolor{red}{83.26}&9.99\\ \hline
     \ac{pgd}-$L_2$(1)& \textcolor{blue}{71.9}&13.07&\textcolor{customgreen}{64.86}&16.04&2.34&0.54&9.89&5.1&59.34&0.49&19.98&10.0&22.69&11.02&\textcolor{red}{99.45}&9.99\\ \hline
     \ac{pgd}-$L_\infty$(8)& \textcolor{customgreen}{61.21}&10.25&55.83&10.64&\textcolor{red}{95.54}&0.54&10.35&5.1&65.74&0.49&21.83&10.0&24.08&11.02&\textcolor{blue}{92.98}&9.99\\ \hline
     \ac{pgd}-$L_\infty$(16)& \textcolor{customgreen}{95.75}&1.82&89.77&4.13&\textcolor{blue}{99.95}&0.54&0.53&5.1&92.95&0.49&16.19&10.0&13.96&11.02&\textcolor{red}{99.99}&9.99\\ \hline
     \ac{cw}-$L_\infty$& 55.14&8.67&43.37&11.96&11.02&0.54&\textcolor{customgreen}{67.01}&5.1&8.3&0.49&46.19&10.0&\textcolor{red}{87.09}&11.02&\textcolor{blue}{85.83}&9.99\\ \hline
     \ac{cw}-\ac{hca}(8)& 15.3&4.73&24.9&7.07&\textcolor{blue}{79.08}&0.54&32.45&5.1&35.43&0.49&33.04&10.0&\textcolor{customgreen}{49.68}&11.02&\textcolor{red}{81.92}&9.99\\ \hline
     \ac{cw}-\ac{hca}(16)& 47.89&6.87&53.27&9.47&\textcolor{blue}{91.47}&0.54&16.07&5.1&\textcolor{customgreen}{68.6}&0.49&23.87&10.0&33.46&11.02&\textcolor{red}{93.84}&9.99\\ \hline
     \ac{df}& 58.47&7.58&64.74&2.3&58.8&0.54&62.33&5.1&44.98&0.49&\textcolor{customgreen}{66.7}&10.0&\textcolor{red}{89.55}&11.02&\textcolor{blue}{83.25}&9.99\\ \hline
     \color{red}\ac{jsm}& \textcolor{customgreen}{82.68}&6.68&82.56&4.3&\textcolor{blue}{90.21}&0.54&93.08&5.1&15.06&0.49&47.96&10.0&\textcolor{red}{96.96}&11.02&70.4&9.99\\ \hline\hline
     \textbf{Average}& \textbf{48.74}&8.56&\textcolor{blue}{\textbf{58.54}}&10.76&\textcolor{customgreen}{\textbf{54.23}}&0.54&\textbf{35.07}&5.1&\textbf{40.4}&0.49&\textbf{34.46}&10&\textbf{49.76}&11.02&\textcolor{red}{\textbf{85.67}}&9.99\\ \hline
\end{tabular}
}
\end{table*}

\begin{table*}[!htb]
\caption{Detection rates (DR\%) and false positive rate (FPR\%) for the detectors against tested white box attacks($\epsilon$) on \image. Top 3 are colored with \textcolor{red}{red}, \textcolor{blue}{blue} and \textcolor{customgreen}{green},  respectively. }
\label{tab:white_tiny}
\resizebox{\linewidth}{!}{
\begin{tabular}{|P{0.12\linewidth}|M{0.039\linewidth}|M{0.039\linewidth}||M{0.039\linewidth}|M{0.039\linewidth}||M{0.039\linewidth}|M{0.039\linewidth}|||M{0.039\linewidth}|M{0.039\linewidth}||M{0.039\linewidth}|M{0.039\linewidth}||M{0.039\linewidth}|M{0.039\linewidth}||M{0.039\linewidth}|M{0.039\linewidth}||M{0.039\linewidth}|M{0.039\linewidth}|}
\hline
     \multirow{3}{*}{Attack}& \multicolumn{6}{c|||}{Supervised Detectors} & \multicolumn{10}{c|}{Unsupervised Detectors}\\ \cline{2-17}
     &\multicolumn{2}{c||}{\ac{kd}+\ac{bu}~\cite{feinman2017detecting}}& \multicolumn{2}{c||}{\ac{lid}~\cite{ma2018characterizing}} & \multicolumn{2}{c|||}{\ac{nss}~\cite{kherchouche2020detection}}& \multicolumn{2}{c||}{\acs{fs}~\cite{xu2017feature}}& \multicolumn{2}{c||}{MagNet~\cite{meng2017magnet}} & \multicolumn{2}{c||}{\ac{dnr}~\cite{sotgiu2020deep}} & \multicolumn{2}{c||}{\ac{sfad}~\cite{aldahdooh2021selective}} & \multicolumn{2}{c|}{\ac{nic}~\cite{ma2019nic}} \\ \cline{2-17}
     &DR & \ac{fpr} & DR & \ac{fpr} & DR & \ac{fpr} & DR & \ac{fpr} & DR & \ac{fpr} & DR & \ac{fpr} & DR & \ac{fpr} & DR & \ac{fpr} \\ \hline \hline
     \ac{fgsm}(8)& 4.76&1.64&0.0&0.0&\textcolor{blue}{83.71}&21.81&23.04&5.33&0.56&0.9&-&-&\textcolor{customgreen}{50.08}&16.38&\textcolor{red}{100}&10.09\\ \hline
     \ac{fgsm}(16)& 0.0&0.0&27.89&6.52&\textcolor{blue}{97.01}&21.81&23.88&5.33&1.16&0.9&-&-&\textcolor{customgreen}{57.74}&16.38&\textcolor{red}{100}&10.09\\ \hline
     \ac{bim}(8)& \textcolor{customgreen}{56.43}&9.75&\textcolor{blue}{65.74}&9.29&33.13&21.81&9.14&5.33&0.65&0.9&-&-&9.23&16.38&\textcolor{red}{100}&10.09\\ \hline
     \ac{bim}(16)& \textcolor{customgreen}{85.93}&2.57&\textcolor{blue}{90.09}&3.75&59.46&21.81&1.92&5.33&0.65&0.9&-&-&6.81&16.38&\textcolor{red}{100}&10.09\\ \hline
     \ac{bim}(32)& \textcolor{customgreen}{96.66}&0.21&\textcolor{blue}{96.77}&1.13&92.88&21.81&0.5&5.33&0.81&0.9&-&-&5.63&16.38&\textcolor{red}{100}&10.09\\ \hline
     \ac{pgd}-$L_1$(15)& 0.0&0.0&0.0&0.0&19.69&21.81&\textcolor{blue}{50.68}&5.33&0.7&0.9&-&-&\textcolor{customgreen}{44.2}&16.38&\textcolor{red}{100}&10.09\\ \hline
     \ac{pgd}-$L_1$(20)& 0.0&0.0&0.0&0.0&19.83&21.81&\textcolor{blue}{54.48}&5.33&0.63&0.9&-&-&\textcolor{customgreen}{37.29}&16.38&\textcolor{red}{100}&10.09\\ \hline
     \ac{pgd}-$L_1$(25)& 0.0&0.0&\textcolor{customgreen}{54.93}&34.09&20.23&21.81&\textcolor{blue}{55.64}&5.33&0.6&0.9&-&-&30.99&16.38&\textcolor{red}{100}&10.09\\ \hline
     \ac{pgd}-$L_2$(0.5)& 0.0&0.0&\textcolor{blue}{63.45}&29.11&21.38&21.81&\textcolor{customgreen}{51.83}&5.33&0.63&0.9&-&-&20.09&16.38&\textcolor{red}{100}&10.09\\ \hline
     \ac{pgd}-$L_2$(1)& \textcolor{customgreen}{40.73}&3.49&\textcolor{blue}{75.43}&14.99&23.55&21.81&28.31&5.33&0.77&0.9&-&-&10.96&16.38&\textcolor{red}{100}&10.09\\ \hline
     \ac{pgd}-$L_\infty$(8)& \textcolor{customgreen}{80.75}&4.93&\textcolor{blue}{91.48}&2.82&58.23&21.81&9.74&5.33&0.64&0.9&-&-&7.62&16.38&\textcolor{red}{100}&10.09\\ \hline
     \ac{pgd}-$L_\infty$(16)& \textcolor{customgreen}{96.71}&0.41&\textcolor{blue}{97.54}&0.87&83.65&21.81&1.99&5.33&0.64&0.9&-&-&5.91&16.38&\textcolor{red}{100}&10.09\\ \hline
     \ac{cw}-$L_\infty$& 0.0&0.0&0.0&0.0&\textcolor{customgreen}{32.35}&21.81&24.78&5.33&7.93&0.9&-&-&\textcolor{blue}{68.13}&16.38&\textcolor{red}{100}&10.09\\ \hline
     \ac{cw}-\ac{hca}(8)& 21.46&3.49&31.89&6.62&38.02&21.81&\textcolor{blue}{44.76}&5.33&0.44&0.9&-&-&\textcolor{customgreen}{42.89}&16.38&\textcolor{red}{100}&10.09\\ \hline
     \ac{cw}-\ac{hca}(16)& 35.34&5.44&\textcolor{blue}{53.87}&10.73&34.35&21.81&\textcolor{customgreen}{39.01}&5.33&0.19&0.9&-&-&33.45&16.38&\textcolor{red}{100}&10.09\\ \hline
     \ac{df}& 0.0&0.0&30.88&17.73&27.2&22.68&\textcolor{customgreen}{36.96}&5.33&1.28&0.9&-&-&\textcolor{blue}{72.32}&16.38&\textcolor{red}{100}&10.09\\ \hline
     \color{red}\ac{jsm}& 1.11&0.28&0&0&16.98&22.68&\textcolor{customgreen}{28.69}&5.33&1.88&0.9&-&-&\textcolor{blue}{99.48}&16.38&\textcolor{red}{100}&10.09\\ \hline\hline
     \textbf{Average}& \textbf{30.58}&1.89&\textcolor{blue}{\textbf{45.88}}&8.10&\textcolor{customgreen}{\textbf{44.80}}&21.86&\textbf{28.55}&5.33&\textbf{1.19}&0.9&\textbf{-}&-&\textbf{35.46}&16.38&\textcolor{red}{\textbf{100}}&10.09\\ \hline
\end{tabular}
}\vspace{3mm}
\end{table*}

\subsection{\ac{lid}~\cite{ma2018characterizing}}
\textbf{White box attacks}. Like \ac{kd}+\ac{bu}, \ac{lid} has better performance for \ac{bim} and \ac{pgd}-$L_\infty$ attacks, while \ac{lid} is much better than \ac{kd}+\ac{bu} against other attacks. Its performance against \ac{jsm} is better than \ac{kd}+\ac{bu} on CIFAR dataset while, for other datasets, it has comparable performance to \ac{kd}+\ac{bu}. It uses \acl{lid} to estimate the distance distribution of the input sample to its $k$-neighbors to assess the space-filling capability of the region surrounding that input sample.  It is clear that distance-based approaches are effective in detecting \acp{ae} but not in high dimensional data, as in \image. Besides, \ac{lid} needs not noisy training data to accurately train the detector to identify the boundaries between clean and adversarial inputs. That is why it is not effective in attacks with very small perturbations. The main limitation of \ac{lid} detector is it has high \ac{fpr} when trained using some attacks, like \ac{pgd} and \ac{cw} attacks, on all datasets except MNIST.

\textbf{Black box attacks}. The detector  effectively detects \ac{st} attacks and it is effective against \ac{sa} on CIFAR and \image. For \ac{pa} attack, \ac{lid} has acceptable detection rate for CIFAR dataset only. Moreover, for \hop attack, the detector is effective only on CIFAR dataset, while it has medium performance, around 60\%, on other datasets. That is because \hop sends the attack to the boundary and makes it difficult for \ac{lid} to estimate the distance distribution.

\textbf{Gray box attacks}. \ac{lid} detector is resistant against transferable features of the attacks on MNIST dataset. For the CIFAR dataset, the detector is not effective against \ac{bim} and \ac{pgd} attacks. Moreover, the detector has comparable performance to white box attacks except for \ac{bim} and \ac{pgd}-$L_\infty$. Finally, the detector has poor performance against gray box attacks because of the same reasons discussed above in white box attacks. 

\subsection{\ac{nss}~\cite{kherchouche2020detection}}
\textbf{White box attacks}. \ac{nss} has shown great impact in image quality assessment to estimate the artifact. For \acp{ae} detection, \ac{nss} based detectors show promised results. The work in~\cite{kherchouche2020detection} used \ac{nss} features to train the detector using only 1000 \ac{pgd} \acp{ae}. In our experiment, we trained the detector using the whole testing data of successful attacks. The detector shows great performance on MNIST except against \ac{cw} attacks. On the other hand, it has poor performance against \ac{pgd}-$L_1$, \ac{pgd}-$L_2$, \ac{cw}, \ac{df} and \ac{jsm}. It seems that the \ac{nss} features of \ac{pgd}-$L_\infty$ generalize well for other attacks, especially other \ac{pgd}-norms. We believe that the detector will give great results if other \acp{ae} are included in the training process. \ac{nss} detector has low \ac{fpr} except for \image \ dataset because the training data has many noise images.

\textbf{Black box attacks}. The \ac{pgd}-based \ac{nss} features do not generalize well against the black box attacks for some attacks of specific dataset. For instance, the detector is effective only against \ac{sa} attacks on MNIST and \image, and it is effective only against \ac{st} on SVHN.

\textbf{Gray box attacks}. Like white box attacks, the detector has comparable performance with gray box attacks with little improvement. Hence, one way to enhance \ac{nss} based detector is to include \acp{ae} generated from different models and to include some noisy samples to be trained as original samples.

\begin{table*}[!htb]
\caption{Detection rates (DR\%) and false positive rate (FPR\%) for the detectors against tested black box attacks($\epsilon$) on MNIST. Top 3 are colored with \textcolor{red}{red}, \textcolor{blue}{blue} and \textcolor{customgreen}{green},  respectively. }
\label{tab:black_mnist}
\resizebox{\linewidth}{!}{
\begin{tabular}{|P{0.1\linewidth}|M{0.039\linewidth}|M{0.039\linewidth}||M{0.039\linewidth}|M{0.039\linewidth}||M{0.039\linewidth}|M{0.039\linewidth}|||M{0.039\linewidth}|M{0.039\linewidth}||M{0.039\linewidth}|M{0.039\linewidth}||M{0.039\linewidth}|M{0.039\linewidth}||M{0.039\linewidth}|M{0.039\linewidth}||M{0.039\linewidth}|M{0.039\linewidth}|}
\hline
     \multirow{3}{*}{Attack}& \multicolumn{6}{c|||}{Supervised Detectors} & \multicolumn{10}{c|}{Unsupervised Detectors}\\ \cline{2-17}
     &\multicolumn{2}{c||}{\ac{kd}+\ac{bu}~\cite{feinman2017detecting}}& \multicolumn{2}{c||}{\ac{lid}~\cite{ma2018characterizing}} & \multicolumn{2}{c|||}{\ac{nss}~\cite{kherchouche2020detection}}& \multicolumn{2}{c||}{\acs{fs}~\cite{xu2017feature}}& \multicolumn{2}{c||}{MagNet~\cite{meng2017magnet}} & \multicolumn{2}{c||}{\ac{dnr}~\cite{sotgiu2020deep}} & \multicolumn{2}{c||}{\ac{sfad}~\cite{aldahdooh2021selective}} & \multicolumn{2}{c|}{\ac{nic}~\cite{ma2019nic}} \\ \cline{2-17}
     &DR & \ac{fpr} & DR & \ac{fpr} & DR & \ac{fpr} & DR & \ac{fpr} & DR & \ac{fpr} & DR & \ac{fpr} & DR & \ac{fpr} & DR & \ac{fpr} \\ \hline \hline
     \ac{sa}& 53.53&0.24&42.78&0.03&87.64&0.0&\textcolor{red}{99.96}&5.27&\textcolor{blue}{99.93}&0.2&81.27&10.01&98.85&10.79&\textcolor{customgreen}{99.68}&10.12\\ \hline
     \hop& 61.82&0.57&61.52&2.22&99.88&0.0&\textcolor{blue}{99.98}&5.27&98.32&0.2&59.98&10.01&\textcolor{customgreen}{99.91}&10.79&\textcolor{red}{100}&10.12\\ \hline
     \ac{st}& 47.94&0.5&\textcolor{customgreen}{93.81}&0.64&12.26&0.0&77.49&5.27&1.61&0.2&88.0&10.01&\textcolor{blue}{97.61}&10.79&\textcolor{red}{99.83}&10.12\\ \hline
     \color{red}\ac{pa}& 91.51&1.83&68.7&2.87&15.83&0&\textcolor{red}{100}&5.27&74.51&0.2&51.49&10.1&\textcolor{blue}{100}&10.79&\textcolor{customgreen}{99.9}&10.12\\ \hline
     \textbf{Average}& \textbf{63.7}&0.79&\textbf{66.7}&1.44&\textbf{53.9}&0&\textbf{\textcolor{customgreen}{94.36}}&5.27&\textbf{68.59}&0.2&\textbf{70.19}&10.01&\textbf{\textcolor{blue}{99.09}}&10.79&\textbf{\textcolor{red}{99.85}}&10.12\\ \hline
\end{tabular}
}
\end{table*}

\begin{table*}[!htb]
\caption{Detection rates (DR\%) and false positive rate (FPR\%) for the detectors against tested black box attacks($\epsilon$) on CIFAR-10. Top 3 are colored with \textcolor{red}{red}, \textcolor{blue}{blue} and \textcolor{customgreen}{green},  respectively. }
\label{tab:black_cifar}
\resizebox{\linewidth}{!}{
\begin{tabular}{|P{0.1\linewidth}|M{0.039\linewidth}|M{0.039\linewidth}||M{0.039\linewidth}|M{0.039\linewidth}||M{0.039\linewidth}|M{0.039\linewidth}|||M{0.039\linewidth}|M{0.039\linewidth}||M{0.039\linewidth}|M{0.039\linewidth}||M{0.039\linewidth}|M{0.039\linewidth}||M{0.039\linewidth}|M{0.039\linewidth}||M{0.039\linewidth}|M{0.039\linewidth}|}
\hline
     \multirow{3}{*}{Attack}& \multicolumn{6}{c|||}{Supervised Detectors} & \multicolumn{10}{c|}{Unsupervised Detectors}\\ \cline{2-17}
     &\multicolumn{2}{c||}{\ac{kd}+\ac{bu}~\cite{feinman2017detecting}}& \multicolumn{2}{c||}{\ac{lid}~\cite{ma2018characterizing}} & \multicolumn{2}{c|||}{\ac{nss}~\cite{kherchouche2020detection}}& \multicolumn{2}{c||}{\acs{fs}~\cite{xu2017feature}}& \multicolumn{2}{c||}{MagNet~\cite{meng2017magnet}} & \multicolumn{2}{c||}{\ac{dnr}~\cite{sotgiu2020deep}} & \multicolumn{2}{c||}{\ac{sfad}~\cite{aldahdooh2021selective}} & \multicolumn{2}{c|}{\ac{nic}~\cite{ma2019nic}} \\ \cline{2-17}
     &DR & \ac{fpr} & DR & \ac{fpr} & DR & \ac{fpr} & DR & \ac{fpr} & DR & \ac{fpr} & DR & \ac{fpr} & DR & \ac{fpr} & DR & \ac{fpr} \\ \hline \hline
     \ac{sa}& 0.0&0.0&\textcolor{customgreen}{85.76}&4.72&1.49&6.56&17.82&5.07&\textcolor{red}{94.04}&0.77&52.86&10.01&\textcolor{blue}{93.91}&10.9&61.88&10.08\\ \hline
     \hop& 28.03&7.19&\textcolor{blue}{88.34}&11.18&21.42&6.56&\textcolor{customgreen}{84.16}&5.07&0.58&0.77&38.81&10.01&\textcolor{red}{95.57}&10.9&67.53&10.08\\ \hline
     \ac{st}& 44.15&3.1&\textcolor{red}{94.23}&5.27&31.73&6.56&22.46&5.07&2.32&0.77&\textcolor{customgreen}{56.20}&10.01&\textcolor{blue}{92.9}&10.9&48.77&10.08\\ \hline
     \color{red}\ac{pa}& 61.94&8.65&\textcolor{blue}{86.85}&7.27&36.56&6.56&\textcolor{customgreen}{67.21}&5.07&0.9&0.77&34.2&10.1&\textcolor{red}{89.76}&10.9&54.44&10.08\\ \hline
     \textbf{Average}& \textbf{33.53}&4.74&\textbf{\textcolor{blue}{88.8}}&7.11&\textbf{22.8}&6.56&\textbf{47.91}&5.07&\textbf{24.46}&0.77&\textbf{45.52}&10.01&\textbf{\textcolor{red}{93.04}}&10.9&\textbf{\textcolor{customgreen}{58.16}}&10.08\\ \hline
\end{tabular}
}
\end{table*}

\begin{table*}[!htb]
\caption{Detection rates (DR\%) and false positive rate (FPR\%) for the detectors against tested black box attacks($\epsilon$) on SVHN. Top 3 are colored with \textcolor{red}{red}, \textcolor{blue}{blue} and \textcolor{customgreen}{green},  respectively. }
\label{tab:black_svhn}
\resizebox{\linewidth}{!}{
\begin{tabular}{|P{0.1\linewidth}|M{0.039\linewidth}|M{0.039\linewidth}||M{0.039\linewidth}|M{0.039\linewidth}||M{0.039\linewidth}|M{0.039\linewidth}|||M{0.039\linewidth}|M{0.039\linewidth}||M{0.039\linewidth}|M{0.039\linewidth}||M{0.039\linewidth}|M{0.039\linewidth}||M{0.039\linewidth}|M{0.039\linewidth}||M{0.039\linewidth}|M{0.039\linewidth}|}
\hline
     \multirow{3}{*}{Attack}& \multicolumn{6}{c|||}{Supervised Detectors} & \multicolumn{10}{c|}{Unsupervised Detectors}\\ \cline{2-17}
     &\multicolumn{2}{c||}{\ac{kd}+\ac{bu}~\cite{feinman2017detecting}}& \multicolumn{2}{c||}{\ac{lid}~\cite{ma2018characterizing}} & \multicolumn{2}{c|||}{\ac{nss}~\cite{kherchouche2020detection}}& \multicolumn{2}{c||}{\acs{fs}~\cite{xu2017feature}}& \multicolumn{2}{c||}{MagNet~\cite{meng2017magnet}} & \multicolumn{2}{c||}{\ac{dnr}~\cite{sotgiu2020deep}} & \multicolumn{2}{c||}{\ac{sfad}~\cite{aldahdooh2021selective}} & \multicolumn{2}{c|}{\ac{nic}~\cite{ma2019nic}} \\ \cline{2-17}
     &DR & \ac{fpr} & DR & \ac{fpr} & DR & \ac{fpr} & DR & \ac{fpr} & DR & \ac{fpr} & DR & \ac{fpr} & DR & \ac{fpr} & DR & \ac{fpr} \\ \hline \hline
     \ac{sa}& 70.02&7.45&52.01&5.95&33.36&0.54&\textcolor{customgreen}{74.76}&5.1&28.32&0.49&45.56&10.0&\textcolor{red}{93.23}&11.02&\textcolor{blue}{78.59}&9.99\\ \hline
     \hop& 84.85&5.72&58.26&10.6&57.59&0.54&\textcolor{blue}{94.42}&5.1&6.13&0.49&34.57&10.0&\textcolor{red}{96.47}&11.02&\textcolor{customgreen}{86.33}&9.99\\ \hline
     \ac{st}& 73.21&7.72&90.12&1.82&\textcolor{red}{99.89}&0.54&32.63&5.1&14.56&0.49&73.41&10.0&\textcolor{blue}{96.81}&11.02&\textcolor{customgreen}{94.4}&9.99\\ \hline
     \color{red}\ac{pa}& 80.35&6.92&71.39&6.34&\textcolor{customgreen}{81.32}&0.54&\textcolor{red}{98.76}&5.1&45.75&0.49&41.4&10.0&\textcolor{blue}{93.95}&11.02&59.87&9.99\\ \hline
     \textbf{Average}& \textcolor{customgreen}{\textbf{77.11}}&6.95&\textbf{67.95}&6.18&\textbf{68.04}&0.54&\textbf{75.14}&5.1&\textbf{23.69}&0.49&\textbf{48.74}&10&\textcolor{red}{\textbf{95.12}}&11.02&\textcolor{blue}{\textbf{79.80}}&9.99\\ \hline
\end{tabular}
}
\end{table*}

\begin{table*}[!htb]
\caption{Detection rates (DR\%) and false positive rate (FPR\%) for the detectors against tested black box attacks($\epsilon$) on \image. Top 3 are colored with \textcolor{red}{red}, \textcolor{blue}{blue} and \textcolor{customgreen}{green},  respectively. }
\label{tab:black_tiny}
\resizebox{\linewidth}{!}{
\begin{tabular}{|P{0.1\linewidth}|M{0.039\linewidth}|M{0.039\linewidth}||M{0.039\linewidth}|M{0.039\linewidth}||M{0.039\linewidth}|M{0.039\linewidth}|||M{0.039\linewidth}|M{0.039\linewidth}||M{0.039\linewidth}|M{0.039\linewidth}||M{0.039\linewidth}|M{0.039\linewidth}||M{0.039\linewidth}|M{0.039\linewidth}||M{0.039\linewidth}|M{0.039\linewidth}|}
\hline
     \multirow{3}{*}{Attack}& \multicolumn{6}{c|||}{Supervised Detectors} & \multicolumn{10}{c|}{Unsupervised Detectors}\\ \cline{2-17}
     &\multicolumn{2}{c||}{\ac{kd}+\ac{bu}~\cite{feinman2017detecting}}& \multicolumn{2}{c||}{\ac{lid}~\cite{ma2018characterizing}} & \multicolumn{2}{c|||}{\ac{nss}~\cite{kherchouche2020detection}}& \multicolumn{2}{c||}{\acs{fs}~\cite{xu2017feature}}& \multicolumn{2}{c||}{MagNet~\cite{meng2017magnet}} & \multicolumn{2}{c||}{\ac{dnr}~\cite{sotgiu2020deep}} & \multicolumn{2}{c||}{\ac{sfad}~\cite{aldahdooh2021selective}} & \multicolumn{2}{c|}{\ac{nic}~\cite{ma2019nic}} \\ \cline{2-17}
     &DR & \ac{fpr} & DR & \ac{fpr} & DR & \ac{fpr} & DR & \ac{fpr} & DR & \ac{fpr} & DR & \ac{fpr} & DR & \ac{fpr} & DR & \ac{fpr} \\ \hline \hline
     \ac{sa}& 33.11&5.13&\textcolor{customgreen}{89.25}&4.47&\textcolor{blue}{95.03}&21.81&25.72&5.33&81.71&0.9&-&-&75.69&16.38&\textcolor{red}{100}&10.09\\ \hline
     \hop& 0.0&0.0&0.0&0.0&\textcolor{customgreen}{60.19}&21.81&48.92&5.33&8.42&0.9&-&-&\textcolor{blue}{69.17}&16.38&\textcolor{red}{100}&10.09\\ \hline
     \ac{st}& 13.94&1.03&\textcolor{blue}{74.11}&24.08&0.06&21.81&25.5&5.33&0.26&0.9&-&-&\textcolor{customgreen}{73.42}&16.38&\textcolor{red}{100}&10.09\\ \hline
     \color{red}\ac{pa}& 0&0&0&0&\textcolor{customgreen}{17.22}&21.81&10.95&5.33&0.74&0.9&-&-&\textcolor{blue}{77.22}&16.38&\textcolor{red}{100}&10.09\\ \hline
     \textbf{Average}& \textbf{11.76}&1.54&\textbf{40.84}&7.14&\textcolor{customgreen}{\textbf{43.13}}&21.81&\textbf{27.77}&5.33&\textbf{22.78}&0.9&\textbf{-}&-&\textcolor{blue}{\textbf{73.88}}&16.38&\textcolor{red}{\textbf{100}}&10.09\\ \hline
\end{tabular}
}
\end{table*}

\subsection{\ac{fs}~\cite{xu2017feature}}
\textbf{White box attacks}. \ac{fs} is one of the promising techniques to apply if we found an effective squeezer that doesn't affect the baseline classifier accuracy and highly gives different confidence for \acp{ae}. It shows effective performance for MNIST dataset, but obtained medium to poor performance on other datasets. These results are consistent with the original paper results. The implemented squeezer is designed for small distortion only, hence it will not succeed against \acp{ae} of large distortion regardless of the $\epsilon$ value if it is small or not. The \ac{fpr} of the \ac{fs} detector is around 5\%, which is acceptable compared to other detectors.

\textbf{Black box attacks}. \ac{fs} detector is shown to be effective against \hop attack except for \image \ dataset, and not effective against other black box attacks for the same reasons discussed with white box attacks. 

\textbf{Gray box attacks}. \ac{fs} detector is an unsupervised detector, thus we expect comparable results with white box attacks. The little difference in performance is due to the threshold value calculation. We use a random subset of testing data to calculate it and we have to use enough amount of data to estimate the threshold. 

\subsection{MagNet~\cite{meng2017magnet}}
\textbf{White box attacks}. MagNet is a denoiser-based detection and defense method. We remind the reader, we report detection accuracy only and that why our results are not consistent with the original paper. MagNet detector works well if the \acp{ae} have high distortion within the $\epsilon$ range, which leads to high reconstruction error. This condition applied to MNIST and SVHN datasets. The small distortion of \acp{ae} of CIFAR and \image \ yield a small reconstruction error and are comparable to clean samples. This adversarial property doesn't stand against the reformer, i.e., the defense, that re-project the \acp{ae} into the training samples manifold and hence, are classified correctly. Thus, one method to improve the denoiser based detectors is to design a high quality denoiser for small distortion attacks and for $L_0$, $L_1$ and $L_2$ based attacks. One advantage of MagNet is that it has very low \ac{fpr} on all datasets, which is important in a defense method to prevent from processing the clean images. 

\textbf{Black box attacks}. For \ac{sa}, MagNet is shown to be effective except for SVHN dataset, while the detector has poor performance against other black box attacks. 

\textbf{Gray box attacks}. Like \ac{fs} detector, MagNet is an unsupervised detector and we expect comparable results with white box attacks. The little difference in performance is due the threshold value calculation since we use 5000 samples from training data to calculate the threshold.

\begin{table*}[!htb]
\caption{Detection rates (DR\%) for the detectors against tested gray box attacks($\epsilon$) on MNIST. \ac{fpr} is same as reported for white box attacks. Top 3 are colored with \textcolor{red}{red}, \textcolor{blue}{blue} and \textcolor{customgreen}{green},  respectively. }
\label{tab:gray_mnist}
\resizebox{\linewidth}{!}{%
\begin{tabular}{|P{0.12\linewidth}|M{0.1\linewidth}||M{0.07\linewidth}||M{0.08\linewidth}||M{0.08\linewidth}||M{0.09\linewidth}||M{0.08\linewidth}||M{0.085\linewidth}||M{0.07\linewidth}|}
\hline
     \multirow{2}{*}{Attack}& \multicolumn{3}{c||}{Supervised Detectors} & \multicolumn{5}{c|}{Unsupervised Detectors}\\ \cline{2-9}
     &\ac{kd}+\ac{bu}~\cite{feinman2017detecting}& \ac{lid}~\cite{ma2018characterizing} & \ac{nss}~\cite{kherchouche2020detection}& \ac{fs}~\cite{xu2017feature}& MagNet~\cite{meng2017magnet} & \ac{dnr}~\cite{sotgiu2020deep} & \ac{sfad}~\cite{aldahdooh2021selective} & \ac{nic}~\cite{ma2019nic} \\ \hline \hline
     \ac{fgsm}(32)& 86.41&78.96&\textcolor{red}{100.0}&97.6&\textcolor{blue}{100.0}&58.1&\textcolor{customgreen}{98.51}&-\\ \hline
     \ac{fgsm}(64)& 66.72&80.14&\textcolor{red}{100.0}&\textcolor{customgreen}{99.06}&\textcolor{blue}{100.0}&89.61&98.91&-\\ \hline
     \ac{fgsm}(80)& 62.89&76.41&\textcolor{red}{100.0}&99.31&\textcolor{blue}{100.0}&92.56&\textcolor{customgreen}{99.64}&-\\ \hline
     \ac{bim}(32)& 78.6&74.18&\textcolor{red}{100.0}&\textcolor{customgreen}{98.9}&\textcolor{blue}{100.0}&57.28&97.5&-\\ \hline
     \ac{bim}(64)& 23.77&53.38&\textcolor{red}{100.0}&\textcolor{customgreen}{97.79}&\textcolor{blue}{100.0}&56.22&77.23&-\\ \hline
     \ac{bim}(80)& 24.64&62.99&\textcolor{red}{100.0}&\textcolor{customgreen}{95.92}&\textcolor{blue}{100.0}&45.87&75.88&-\\\hline
     \ac{pgd}-$L_1$(10)& 78.53&74.69&\textcolor{customgreen}{86.91}&\textcolor{blue}{94.89}&14.83&47.29&\textcolor{red}{97.91}&-\\ \hline
     \ac{pgd}-$L_1$(15)& 70.18&69.52&\textcolor{red}{98.62}&\textcolor{blue}{95.73}&81.04&55.35&\textcolor{customgreen}{95.26}&-\\ \hline
     \ac{pgd}-$L_1$(20)& 61.51&56.63&\textcolor{red}{99.78}&\textcolor{customgreen}{93.65}&\textcolor{blue}{99.71}&57.54&91.16&-\\ \hline
     \ac{pgd}-$L_2$(1)& 74.63&70.99&\textcolor{customgreen}{90.78}&\textcolor{blue}{96.24}&53.01&49.94&\textcolor{red}{98.41}&-\\ \hline
     \ac{pgd}-$L_2$(1.5)& 0.03&46.41&\textcolor{red}{99.46}&\textcolor{blue}{96.72}&\textcolor{customgreen}{94.98}&58.31&94.39&-\\ \hline
     \ac{pgd}-$L_2$(2)& 0.13&33.36&\textcolor{red}{99.93}&\textcolor{customgreen}{92.01}&\textcolor{blue}{99.73}&54.95&85.68&-\\ \hline
     \ac{pgd}-$L_\infty$(32)& 78.57&73.42&\textcolor{red}{100.0}&\textcolor{customgreen}{98.91}&\textcolor{blue}{100.0}&57.13&97.51&-\\ \hline
     \ac{pgd}-$L_\infty$(64)& 23.93&53.12&\textcolor{red}{100.0}&\textcolor{customgreen}{97.87}&\textcolor{blue}{100.0}&56.04&77.1&-\\ \hline
     \ac{cw}-$L_\infty$& 25.92&62.67&3.78&\textcolor{blue}{97.12}&48.21&\textcolor{customgreen}{67.21}&\textcolor{red}{98.17}&-\\ \hline
     \ac{cw}-\ac{hca}(80)& 36.11&76.25&59.71&\textcolor{blue}{99.93}&\textcolor{red}{100.0}&77.69&\textcolor{customgreen}{96.16}&-\\ \hline
     \ac{cw}-\ac{hca}(128)& 91.16&99.73&1.38&\textcolor{blue}{100.0}&\textcolor{red}{100.0}&99.86&\textcolor{customgreen}{100.0}&-\\ \hline
     \ac{df}& 42.53&95.31&\textcolor{red}{99.99}&60.22&\textcolor{customgreen}{99.54}&98.25&\textcolor{blue}{99.68}&-\\ \hline
     \textbf{Average}& \textbf{51.46}&\textbf{68.79} &\textbf{85.57} &\textcolor{blue}{\textbf{95.10}}&\textcolor{customgreen}{\textbf{88.39}}&\textbf{65.51}&\textcolor{red}{\textbf{93.28}}&-\\ \hline
\end{tabular}
}
\end{table*}

\begin{table*}[!htb]
\caption{Detection rates (DR\%) for the detectors against tested gray box attacks($\epsilon$) on CIFAR-10. \ac{fpr} is same as reported for white box attacks. Top 3 are colored with \textcolor{red}{red}, \textcolor{blue}{blue} and \textcolor{customgreen}{green},  respectively. }
\label{tab:gray_cifar}
\resizebox{\linewidth}{!}{
\begin{tabular}{|P{0.12\linewidth}|M{0.1\linewidth}||M{0.07\linewidth}||M{0.08\linewidth}||M{0.08\linewidth}||M{0.09\linewidth}||M{0.08\linewidth}||M{0.085\linewidth}||M{0.07\linewidth}|}
\hline
     \multirow{2}{*}{Attack}& \multicolumn{3}{c||}{Supervised Detectors} & \multicolumn{5}{c|}{Unsupervised Detectors}\\ \cline{2-9}
     &\ac{kd}+\ac{bu}~\cite{feinman2017detecting}& \ac{lid}~\cite{ma2018characterizing} & \ac{nss}~\cite{kherchouche2020detection}& \ac{fs}~\cite{xu2017feature}& MagNet~\cite{meng2017magnet} & \ac{dnr}~\cite{sotgiu2020deep} & \ac{sfad}~\cite{aldahdooh2021selective} & \ac{nic}~\cite{ma2019nic} \\ \hline \hline
     \ac{fgsm}(8)& \textcolor{customgreen}{50.47}&49.4&\textcolor{red}{87.06}&42.18&0.36&31.94&\textcolor{blue}{77.73}&-\\ \hline
     \ac{fgsm}(16)& 42.21&\textcolor{customgreen}{70.72}&\textcolor{red}{99.87}&44.75&2.38&31.03&\textcolor{blue}{82.36}&-\\ \hline
     \ac{fgsm}(32)& 0.08&87.32&\textcolor{blue}{99.68}&37.28&\textcolor{red}{100.0}&28.5&\textcolor{customgreen}{92.95}&-\\ \hline
     \ac{bim}(8)& 2.71&3.8&\textcolor{customgreen}{26.93}&\textcolor{red}{54.16}&0.3&21.9&\textcolor{blue}{41.8}&-\\ \hline
     \ac{bim}(16)& 0.24&1.32&\textcolor{red}{77.76}&\textcolor{blue}{22.04}&0.58&11.05&\textcolor{customgreen}{12.58}&-\\ \hline
     \ac{pgd}-$L_1$(5)& \textcolor{blue}{68.29}&0.0&3.29&41.8&0.13&\textcolor{customgreen}{42.63}&\textcolor{red}{93.55}&-\\ \hline
     \ac{pgd}-$L_1$(10)& 18.1&25.04&4.86&\textcolor{blue}{52.35}&0.21&\textcolor{customgreen}{36.84}&\textcolor{red}{79.44}&-\\ \hline
     \ac{pgd}-$L_1$(15)& 14.68&15.24&7.61&\textcolor{blue}{60.28}&0.22&\textcolor{customgreen}{32.02}&\textcolor{red}{65.96}&-\\ \hline
     \ac{pgd}-$L_2$(0.25)& \textcolor{blue}{59.13}&15.81&4.89&\textcolor{customgreen}{52.19}&0.18&39.68&\textcolor{red}{81.62}&-\\ \hline
     \ac{pgd}-$L_2$(0.3125)& 0.71&17.77&5.54&\textcolor{blue}{56.83}&0.18&\textcolor{customgreen}{34.55}&\textcolor{red}{74.87}&-\\ \hline
     \ac{pgd}-$L_2$(0.5)& 11.32&12.82&9.78&\textcolor{red}{61.02}&0.27&\textcolor{customgreen}{28.47}&\textcolor{blue}{56.45}&-\\ \hline
     \ac{pgd}-$L_\infty$(8)& 1.87&2.78&\textcolor{red}{51.41}&\textcolor{blue}{46.35}&0.35&16.98&\textcolor{customgreen}{26.47}&-\\ \hline
     \ac{pgd}-$L_\infty$(16)& 0.26&0.84&\textcolor{red}{91.18}&\textcolor{blue}{14.95}&0.61&8.59&\textcolor{customgreen}{8.87}&-\\ \hline
     \ac{cw}-$L_\infty$& 29.1&\textcolor{blue}{62.65}&\textcolor{customgreen}{60.32}&29.61&38.73&47.05&\textcolor{red}{91.76}&-\\ \hline
     \ac{cw}-\ac{hca}(8)& \textcolor{customgreen}{47.71}&26.12&44.61&\textcolor{blue}{56.62}&0.25&37.02&\textcolor{red}{74.93}&-\\ \hline
     \ac{cw}-\ac{hca}(16)& 11.6&36.28&\textcolor{red}{70.57}&\textcolor{customgreen}{41.67}&0.39&30.76&\textcolor{blue}{55.11}&-\\ \hline
     \ac{df}& 83.74&\textcolor{customgreen}{88.39}&50.19&42.94&\textcolor{blue}{89.78}&19.52&\textcolor{red}{97.74}&-\\ \hline
     \textbf{Average}& \textbf{26.01}&\textbf{30.37}&\textcolor{blue}{\textbf{46.80}}&\textcolor{customgreen}{\textbf{44.53}}&\textbf{13.82}&\textbf{29.33}&\textcolor{red}{\textbf{65.54}}&-\\ \hline
\end{tabular}
}
\end{table*}

\begin{table*}[!htb]
\caption{Detection rates (DR\%) for the detectors against tested gray box attacks($\epsilon$) on SVHN. \ac{fpr} is same as reported for white box attacks. Top 3 are colored with \textcolor{red}{red}, \textcolor{blue}{blue} and \textcolor{customgreen}{green},  respectively. }
\label{tab:gray_svhn}
\resizebox{\linewidth}{!}{
\begin{tabular}{|P{0.12\linewidth}|M{0.1\linewidth}||M{0.07\linewidth}||M{0.08\linewidth}||M{0.08\linewidth}||M{0.09\linewidth}||M{0.08\linewidth}||M{0.085\linewidth}||M{0.07\linewidth}|}
\hline
     \multirow{2}{*}{Attack}& \multicolumn{3}{c||}{Supervised Detectors} & \multicolumn{5}{c|}{Unsupervised Detectors}\\ \cline{2-9}
     &\ac{kd}+\ac{bu}~\cite{feinman2017detecting}& \ac{lid}~\cite{ma2018characterizing} & \ac{nss}~\cite{kherchouche2020detection}& \ac{fs}~\cite{xu2017feature}& MagNet~\cite{meng2017magnet} & \ac{dnr}~\cite{sotgiu2020deep} & \ac{sfad}~\cite{aldahdooh2021selective} & \ac{nic}~\cite{ma2019nic} \\ \hline \hline
     \ac{fgsm}(8)& 48.47&\textcolor{customgreen}{78.15}&\textcolor{red}{99.17}&47.43&4.63&46.76&\textcolor{blue}{89.01}&-\\ \hline
     \ac{fgsm}(16)& 42.05&\textcolor{customgreen}{92.09}&\textcolor{red}{99.93}&53.23&13.04&53.62&\textcolor{blue}{93.43}&-\\ \hline
     \ac{bim}(8)& 3.87&22.6&\textcolor{customgreen}{31.97}&26.01&7.46&\textcolor{blue}{33.98}&\textcolor{red}{47.1}&-\\ \hline
     \ac{bim}(16)& 15.6&\textcolor{blue}{41.42}&\textcolor{red}{99.6}&5.26&\textcolor{customgreen}{28.48}&21.99&23.77&-\\ \hline
     \ac{pgd}-$L_1$(15)& 10.57&\textcolor{customgreen}{36.74}&0.31&\textcolor{blue}{41.48}&4.22&35.43&\textcolor{red}{58.63}&-\\ \hline
     \ac{pgd}-$L_1$(20)& 9.83&\textcolor{blue}{36.75}&0.37&31.47&7.24&\textcolor{customgreen}{34.25}&\textcolor{red}{49.37}&-\\ \hline
     \ac{pgd}-$L_1$(25)& 14.07&\textcolor{blue}{38.85}&0.42&26.45&10.62&\textcolor{customgreen}{33.56}&\textcolor{red}{43.34}&-\\ \hline
     \ac{pgd}-$L_2$(0.3125)& 31.27&26.57&0.24&\textcolor{blue}{50.07}&1.57&\textcolor{customgreen}{36.07}&\textcolor{red}{70.22}&-\\ \hline
     \ac{pgd}-$L_2$(0.5)& 8.09&\textcolor{blue}{37.54}&0.32&\textcolor{customgreen}{35.67}&4.86&34.21&\textcolor{red}{52.99}&-\\ \hline
     \ac{pgd}-$L_2$(1)& 20.37&\textcolor{red}{45.19}&1.01&14.91&17.05&\textcolor{customgreen}{31.44}&\textcolor{blue}{33.16}&-\\ \hline
     \ac{pgd}-$L_\infty$(8)& 10.09&\textcolor{customgreen}{34.73}&\textcolor{red}{97.0}&17.4&17.13&31.48&\textcolor{blue}{36.58}&-\\ \hline
     \ac{pgd}-$L_\infty$(16)& 16.45&\textcolor{blue}{42.35}&\textcolor{red}{99.96}&4.29&\textcolor{customgreen}{38.64}&19.47&21.84&-\\ \hline
     \ac{cw}-$L_\infty$& 46.33&\textcolor{customgreen}{53.9}&27.79&\textcolor{blue}{57.53}&14.28&53.27&\textcolor{red}{88.29}&-\\ \hline
     \ac{cw}-\ac{hca}(8)& 22.01&24.3&\textcolor{red}{82.77}&33.4&12.48&\textcolor{blue}{82.73}&\textcolor{customgreen}{56.58}&-\\ \hline
     \ac{cw}-\ac{hca}(16)& 18.33&\textcolor{customgreen}{39.37}&\textcolor{red}{93.99}&18.99&35.45&33.72&\textcolor{blue}{42.12}&-\\ \hline
     \ac{df}& 74.34&\textcolor{blue}{89.72}&\textcolor{customgreen}{82.11}&70.6&68.44&28.08&\textcolor{red}{93.66}&-\\ \hline
     \textbf{Average}& \textbf{24.48}&\textcolor{customgreen}{\textbf{46.27}}&\textcolor{blue}{\textbf{51.06}}&\textbf{33.39}&\textbf{17.85}&\textbf{38.13}&\textcolor{red}{\textbf{56.26}}&\textbf{-}\\ \hline
\end{tabular}
}
\end{table*}

\begin{table*}[!htb]
\caption{Detection rates (DR\%) for the detectors against tested gray box attacks($\epsilon$) on \image. \ac{fpr} is same as reported for white box attacks. Top 2 are colored with \textcolor{red}{red}, and \textcolor{blue}{blue},  respectively. }
\label{tab:gray_tiny}
\resizebox{\linewidth}{!}{
\begin{tabular}{|P{0.12\linewidth}|M{0.1\linewidth}||M{0.07\linewidth}||M{0.08\linewidth}||M{0.08\linewidth}||M{0.09\linewidth}||M{0.08\linewidth}||M{0.085\linewidth}||M{0.07\linewidth}|}
\hline
     \multirow{2}{*}{Attack}& \multicolumn{3}{c||}{Supervised Detectors} & \multicolumn{5}{c|}{Unsupervised Detectors}\\ \cline{2-9}
     &\ac{kd}+\ac{bu}~\cite{feinman2017detecting}& \ac{lid}~\cite{ma2018characterizing} & \ac{nss}~\cite{kherchouche2020detection}& \ac{fs}~\cite{xu2017feature}& MagNet~\cite{meng2017magnet} & \ac{dnr}~\cite{sotgiu2020deep} & \ac{sfad}~\cite{aldahdooh2021selective} & \ac{nic}~\cite{ma2019nic} \\ \hline \hline
     \ac{fgsm}(8)& 7.14&0.0&\textcolor{red}{87.16}&31.22&0.37&-&\textcolor{blue}{65.65}&-\\ \hline
     \ac{fgsm}(16)& 0.0&25.23&\textcolor{red}{98.23}&28.88&0.84&-&\textcolor{blue}{64.87}&-\\ \hline
     \ac{bim}(8)& 30.57&16.12&\textcolor{red}{62.85}&28.32&0.39&-&\textcolor{blue}{59.02}&-\\ \hline
     \ac{bim}(16)& 11.47&10.0&\textcolor{red}{82.07}&31.85&0.37&-&\textcolor{blue}{53.26}&-\\ \hline
     \ac{bim}(32)& 1.37&4.12&\textcolor{red}{91.45}&28.88&0.51&-&\textcolor{blue}{52.45}&-\\ \hline
     \ac{pgd}-$L_1$(15)& 0.0&0.0&\textcolor{blue}{17.22}&12.16&1.11&-&\textcolor{red}{81.11}&-\\ \hline
     \ac{pgd}-$L_1$(20)& 0.0&0.0&\textcolor{blue}{17.13}&16.98&1.2&-&\textcolor{red}{79.68}&-\\ \hline
     \ac{pgd}-$L_1$(25)& 0.0&\textcolor{blue}{25.71}&18.18&15.17&0.94&-&\textcolor{red}{78.06}&-\\ \hline
     \ac{pgd}-$L_2$(0.5)& 0.0&\textcolor{blue}{29.9}&18.11&23.68&0.62&-&\textcolor{red}{74.02}&-\\ \hline
     \ac{pgd}-$L_2$(1)& 13.45&\textcolor{blue}{25.4}&19.76&21.76&0.66&-&\textcolor{red}{65.38}&-\\ \hline
     \ac{pgd}-$L_\infty$(8)& 18.52&7.66&\textcolor{red}{68.66}&30.32&0.32&-&\textcolor{blue}{57.89}&-\\ \hline
     \ac{pgd}-$L_\infty$(16)& 2.07&4.16&\textcolor{red}{90.29}&29.61&0.36&-&\textcolor{blue}{55.19}&-\\ \hline
     \ac{cw}-$L_\infty$& 0.0&0.0&\textcolor{blue}{55.07}&20.85&11.52&-&\textcolor{red}{67.28}&-\\ \hline
     \ac{cw}-\ac{hca}(8)& 15.21&18.77&\textcolor{blue}{41.23}&24.16&0.61&-&\textcolor{red}{72.15}&-\\ \hline
     \ac{cw}-\ac{hca}(16)& 23.86&30.44&\textcolor{blue}{34.52}&29.66&0.12&-&\textcolor{red}{67.93}&-\\ \hline
     \textbf{Average}& \textbf{8.24}&\textbf{13.17}&\textcolor{blue}{\textbf{53.46}}&\textbf{24.90}&\textbf{1.33}&\textbf{-}&\textcolor{red}{\textbf{66.26}}&\textbf{-}\\ \hline
\end{tabular}
}
\end{table*}

\subsection{\ac{dnr}~\cite{sotgiu2020deep}}
\textbf{White box attacks}. This detector trains three \ac{svm} classifiers, each has one or more layers representative output, then transfers their outputs to train the last \ac{svm} classifier. It was believed that the confidence probability of \acp{ae} is less than of clean samples. The detector has limited success on MNIST and poor to medium performance on other datasets. We didn't test it in \image \ due to high complexity to train \ac{svm} classifiers. The \ac{fpr} of this method is flexible and it is set to be 10\%. 

\textbf{Black box attacks}. The results show that the detector is effective for MNIST dataset against \ac{sa} and \ac{st} attacks but not for \hop attacks due to attack ability to maintain the confidence probability of the \ac{ae} as high as of those of clean samples.

\textbf{Gray box attacks}. \ac{dnr} detector maintains its ability to detect the transferable attacks on all datasets against the tested attacks as compared to white box attacks.

\subsection{\ac{sfad}~\cite{aldahdooh2021selective}} \label{sec: sfad_results}
\textbf{White box attacks}. \ac{sfad} detector is basically an ensemble detection technique that combines an uncertainty method through the selective rejection, confidence probability, like \ac{dnr} and bi-modal mismatch detection. Hence, for MNIST it achieves best performance evaluation in general compared with the tested detectors. For other datasets, the detector is effective for \ac{fgsm}, \ac{cw}, \ac{df} and \ac{jsm} attacks, while it has poor to medium performance for iterative attacks including \ac{bim} and \ac{pgd}. The detector can be improved by tuning the different parameters of its ensemble detection to improve the detection rate and the \ac{fpr}.

\textbf{Black box attacks}. In general, the non-gradient based attacks, like black box attacks, sound to be easily detected using \ac{sfad} detector. The results show that the detector is effective to detect all the tested black box attacks. Its power comes from the employed confidence-based detection\linebreak method that relies on processing the features of the selective classifiers using autoencoders, up/down sampling and noise addition. 

\textbf{Gray box attacks}. The detector shows better performance in detecting gray box attacks compared to white box attacks on all datasets, except for \ac{pgd}-$L_\infty$ on CIFAR. Compared to other detectors, in general, \ac{sfad} \linebreak achieves better performance in detecting gray box attacks.
\subsection{\ac{nic}~\cite{ma2019nic}}
\textbf{White box attacks}. According to the original paper, the provenance invariant alone or the activation value invariant alone is not effective in detecting \acp{ae}. Hence, \ac{nic} combines two network invariants, the provenance channel and the activation value channel. Our results are consistent with the reported results in the original paper to the large extent. The little difference is due to the fact that our experiments didn't optimize the OneClassSVM classifiers' parameters because of \ac{nic} complexity issue. \ac{nic}, in general and relative to other detectors, achieves very high detection rate in most of the attacks, except for \ac{cw} and \ac{jsm} attacks on CIFAR and SVHN datasets. Small distribution of \ac{pgd}-$L_1$ and \ac{pgd}-$L_2$ based attacks sounds to be much harder than the high distortion \ac{ae} of the same attacks. That is due to the fact that some provenance and activation value channels of \acp{ae} are so close to the clean samples.

\textbf{Black box attacks}. Similar to \ac{sfad}, \ac{nic} is shown to be effective for black box attacks. \ac{sa}, \hop and \ac{st} attacks highly trigger the provenance and the activation value channels which make them easily detected using \ac{nic}.

\textbf{Gray box attacks}. Due to the \ac{nic} complexity, we didn't test it in a gray box attack scenario, but we expect its performance to be comparable with the white box attacks' results.

\begin{table*}[!htb]
\caption{Complexity(CM), overhead(OV) and inference time latency (INF) performance for each detector in 3-star ranking. $\bigstar$=low, $\bigstar\bigstar$=middle, $\bigstar\bigstar\bigstar$=high}
\label{tab:performance_cm_ov_inf}
\resizebox{\linewidth}{!}{
\begin{tabular}{|P{0.09\linewidth}|M{0.1\linewidth}||M{0.08\linewidth}||M{0.075\linewidth}||M{0.065\linewidth}||M{0.09\linewidth}||M{0.08\linewidth}||M{0.085\linewidth}||M{0.07\linewidth}|}
\hline
     \multirow{2}{*}{\shortstack{Performance\\Measure}}& \multicolumn{3}{c||}{Supervised Detectors} & \multicolumn{5}{c|}{Unsupervised Detectors}\\ \cline{2-9}
     &\ac{kd}+\ac{bu}~\cite{feinman2017detecting}& \ac{lid}~\cite{ma2018characterizing} & \ac{nss}~\cite{kherchouche2020detection}& \ac{fs}~\cite{xu2017feature}& MagNet~\cite{meng2017magnet} & \ac{dnr}~\cite{sotgiu2020deep} & \ac{sfad}~\cite{aldahdooh2021selective} & \ac{nic}~\cite{ma2019nic} \\ \hline \hline
     OV& $\bigstar$&$\bigstar$&$\bigstar$&$\bigstar$&$\bigstar\bigstar$&$\bigstar\bigstar\bigstar$&$\bigstar\bigstar\bigstar$&$\bigstar\bigstar\bigstar$\\ \hline
     CM& $\bigstar\bigstar\bigstar$&$\bigstar\bigstar\bigstar$&$\bigstar\bigstar$&$\bigstar$&$\bigstar\bigstar$&$\bigstar\bigstar\bigstar$&$\bigstar\bigstar$&$\bigstar\bigstar\bigstar$\\ \hline
     INF& $\bigstar$&$\bigstar$&$\bigstar$&$\bigstar\bigstar$&$\bigstar\bigstar$&$\bigstar$&$\bigstar$&$\bigstar\bigstar\bigstar$\\ \hline
\end{tabular}
}
\end{table*}

\subsection{Performance on high resolution dataset} \label{sec: high_resolution}
In this subsection, the efforts for testing the detection methods on ImageNet~\cite{imagenet_cvpr09}, as a high resolution dataset, will be discussed. Table \ref{tab:compare_stateoftheart} shows that few detectors were tested using ImageNet. ImageNet is a dataset that contains 14 million images annotated with 1000 classes. It is widely used in computer vision research. When it is used for training the neural networks, the images are downsampled to $224\times 224$, $299\times 299$, or $384\times 384$ to match the model's and computational requirements.


\textbf{\ac{lid}~\cite{ma2018characterizing}:} ImageNet was considered in the experiments of \cite{ma2019nic}. It was shown that it does not scale well on ImageNet white box and black box attacks. It achieves the detection rate 82\% on average. The main reason behind this is that ImageNet's images contain more noises, which makes it more difficult for \ac{lid} to identify the boundaries between clean images and adversarial images. Moreover, \ac{lid} has high \ac{fpr}, around 14.5\%.

\textbf{\ac{nss}~\cite{kherchouche2020detection}:} ImageNet was considered in the experiments of \cite{kherchouche2020detection}, and it showed a limited success for \ac{cw} attacks. It achieves a detection rate of 84\% with 6.2\% \ac{fpr}.

\textbf{\acs{fs}~\cite{xu2017feature}:} ImageNet was considered in the experiments of \cite{ma2019nic}. It was shown that \ac{fs} has similar low performance for \ac{fgsm} and \ac{bim} based attacks. It achieves 43\% 64\% of detection rates for \ac{fgsm} and \ac{bim} attacks, respectively.  

\textbf{MagNet~\cite{meng2017magnet}:} In \cite{liao2018defense}, it was shown that MagNet doesn't scale well for high resolution images and has high \ac{fpr}. Besides denoiser-based detectors requiring large computation power to be trained on large datasets, and  denoiser-based detectors have also be shown to not perform well against $L_0$-norm attack.

\textbf{\ac{dnr}~\cite{sotgiu2020deep} and \ac{sfad}~\cite{aldahdooh2021selective}:} \ac{sfad} is expected to perform well specially for black box attacks and for \ac{cw}, \ac{df}, and $L_0$ attacks, as discussed in Section \ref{sec: sfad_results}. The main drawback of these detectors is to train more 4 classifiers for the detection process, which is time consuming. 

\textbf{\ac{nic}~\cite{ma2019nic}:} It was shown in \cite{ma2019nic} that \ac{nic} performs well on ImageNet dataset but it has a high \ac{fpr}, 14.6\% on ResNet50, and has a high runtime overhead that reaches 28\% on ResNet50.

\subsection{Other performance evaluations}
Performance measures related to complexity (CM), overhead (OV) and inference time latency (INF) is very important, but it is application dependent. For  \ac{kd}+\ac{bu}~\cite{feinman2017detecting} and  \ac{lid}~\cite{ma2018characterizing} detectors, to train them with known attacks is time consuming taking into consideration that \ac{lid} requires more time. Their overhead is very small, since we need to save only the classifier parameters, while the inference time is very small. \ac{nss}~\cite{kherchouche2020detection} has middle complexity since it is trained only with \ac{pgd}-based \acp{ae}, and have very small overhead due to saving \ac{svm} classifier parameters, and have no latency in the inference time since feature extraction process can be done in parallel with the prediction process of the baseline classifier. For \ac{fs}~\cite{xu2017feature}, the training complexity is very low and has no overhead, while it has middle inference time due to generating squeezed images. MagNet~\cite{meng2017magnet} has middle complexity and overhead due to denoiser training and its parameters saving, while it has small latency due to the detection process coming before the reformer and the baseline prediction processes. \ac{dnr}~\cite{sotgiu2020deep} and \ac{sfad}~\cite{aldahdooh2021selective} have high and middle complexity, respectively. Both have high overhead due to classifiers parameters saving, and both have no inference time latency. Finally, \ac{nic}~\cite{ma2019nic} has high complexity, overhead and latency than other detectors due to per-layer classifiers training and parameters saving. Table \ref{tab:performance_cm_ov_inf} shows the estimate of 3-star rank for each performance measure per detector.

\subsection{Supervised vs. Unsupervised Detection}
In this subsection, we discuss the effectiveness of supervised and unsupervised detection methods with respect to the performance criteria that are discussed in Section \ref{sec: detection_methods}.
\begin{itemize}
\item \textbf{Efficiency:} If the number of adversarial attacks algorithms is limited, it would be more efficient to build supervised detection models. Unfortunately, this is not the case, hence, putting more efforts to build unsupervised detection models is more effective since it has the power to model the normal samples distributions and features space, as in \ac{lid} and \ac{sfad}, and to detect samples that lie out of this distribution and feature space. That will be more effective in order to detect unknown attacks, especially the black box attacks. 

\item\textbf{Overhead:} The main price that the unsupervised detection models pay is the overhead. Most of the unsupervised detection models learn extra models to help the baseline model to detect the \acp{ae}. The extra models need an additional storage space to be stored in which might not be applicable in some devices and systems. \ac{lid}, for instance, models the distribution of normal samples for each layer. Moreover, \ac{dnr} and \ac{sfad} model the features space of some layers, which requires additional storage space.

\item\textbf{Complexity:} There is no preference with respect to the complexity issues. It highly depends on the complexity of implemented algorithms. For instance, \ac{fs} doesn't have any training and depends on the predicted scores of the transformed samples. On the other hand, \ac{nss}, MagNet, and \ac{sfad} have modest complexity if they are compared with \ac{nic} and \ac{lid}. The source of complexity in \ac{nic} came from the training models for each layer, while the complexity source in \ac{lid} came from the features extraction processes that are required for training.

\item\textbf{Inference time:} Detection techniques have access to the baseline model outputs at different layers, which facilitates working in parallel with the baseline models. Supervised detection methods take advantage of such parallelism and don't compromise inference time constraints, while some unsupervised detection methods, such as \ac{fs}, MagNet, and \ac{nic} compromise the inference time constraints. \ac{fs} has first to apply many transformation methods before running the detection. MagNet has first to apply the detection and then run the baseline model for the prediction. \ac{nic} has to wait until all layers run the corresponding detection model and then combine all model results to run the final prediction. This extra latency is not suitable for real time applications. 

\item\textbf{False positive rate:} There is no preference for any category with respect to false positive rate values. It highly depends on the implemented algorithm. For instance \ac{lid}, \ac{nss}, and \ac{sfad} have high \ac{fpr} for \image~dataset. 
\end{itemize}

\subsection{Content, \ac{cnn}, and Detectors related discussion}
\begin{figure}[!t]
    \begin{center}
        \includegraphics[width=0.9\linewidth, keepaspectratio]{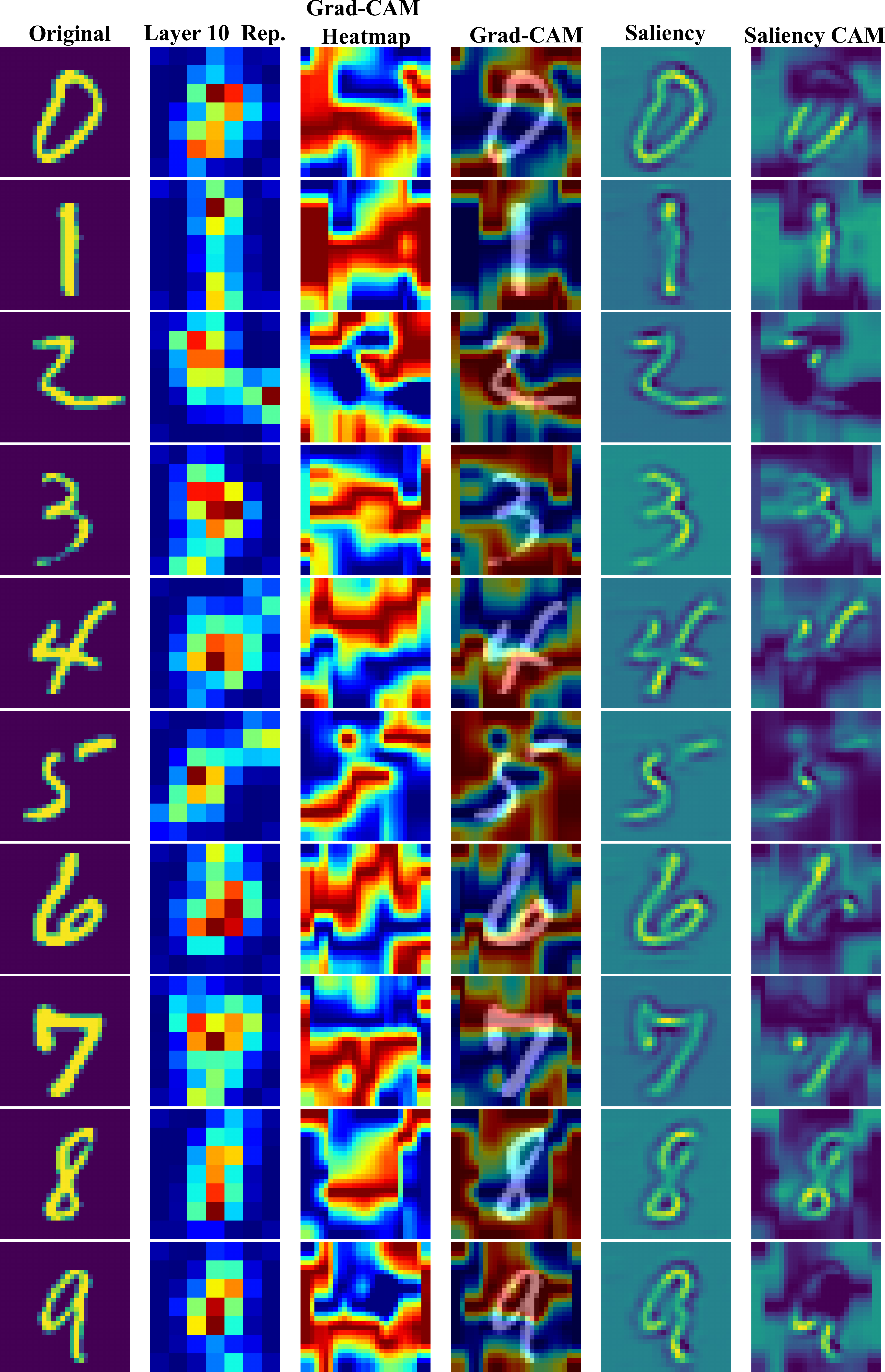}\vspace{-3mm}
    \end{center}
    \caption{MNIST dataset original samples visualization. First column is the original/clean image sample. The second column is the representative output of the $10^{th}$ layer of baseline classifier. The third column is the heatmap generated with \ac{cam} technique. Fourth column is a combination of the heatmap and the corresponding original image. The fifth column is the Saliency map generated with Back-propagation technique while the last column is the combination of the saliency and the \ac{cam} heatmap.} \vspace{-3mm}
    \label{fig:orig_cnn_mnist}
\end{figure}

\begin{figure}[!t]
    \begin{center}
        \includegraphics[width=0.9\linewidth, keepaspectratio]{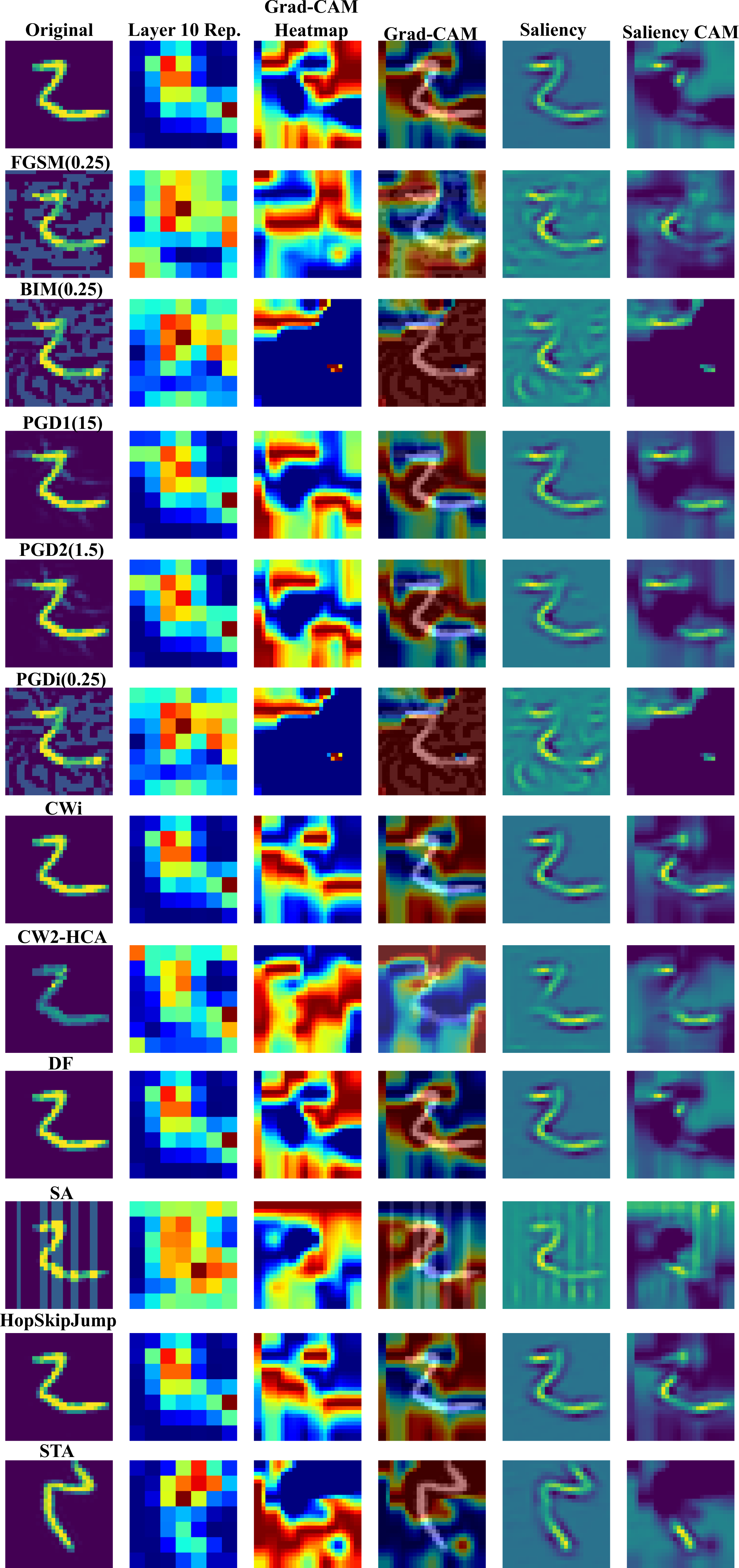}\vspace{-3mm}
    \end{center}
    \caption{MNIST dataset original and adversarial samples visualization. First column is the original/clean or attacked image sample. The second column is the representative output of the $10^{th}$ layer of baseline classifier. The third column is the heatmap generated with \ac{cam} technique. Fourth column is a combination of the heatmap and the corresponding original or attacked image. The fifth column is the Saliency map generated with Back-propagation technique while the last column is the combination of the saliency and the \ac{cam} heatmap.} \vspace{-3mm}
    \label{fig:adv_cnn_mnist}
\end{figure}

\begin{figure}[!t]
    \begin{center}
        \includegraphics[width=0.9\linewidth, keepaspectratio]{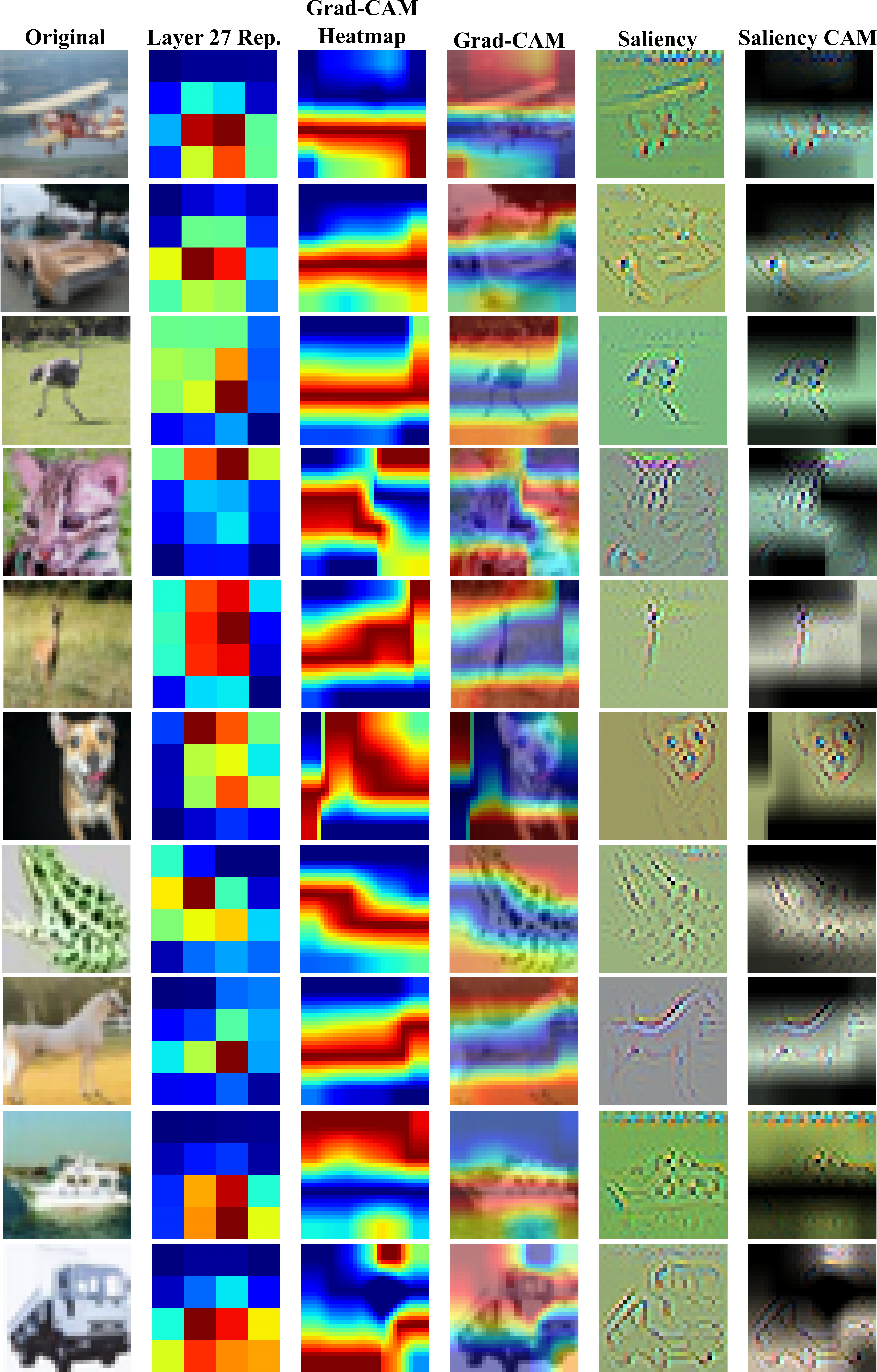}\vspace{-3mm}
    \end{center}
    \caption{CIFAR dataset original samples visualization. First column is the original/clean image sample. The second column is the representative output of the $27^{th}$ layer of baseline classifier. The third column is the heatmap generated with \ac{cam} technique. Fourth column is a combination of the heatmap and the corresponding original image. The fifth column is the Saliency map generated with Back-propagation technique while the last column is the combination of the saliency and the \ac{cam} heatmap.} \vspace{-3mm}
    \label{fig:orig_cnn_cifar}
\end{figure}

\begin{figure}[!t]
    \begin{center}
        \includegraphics[width=0.9\linewidth, keepaspectratio]{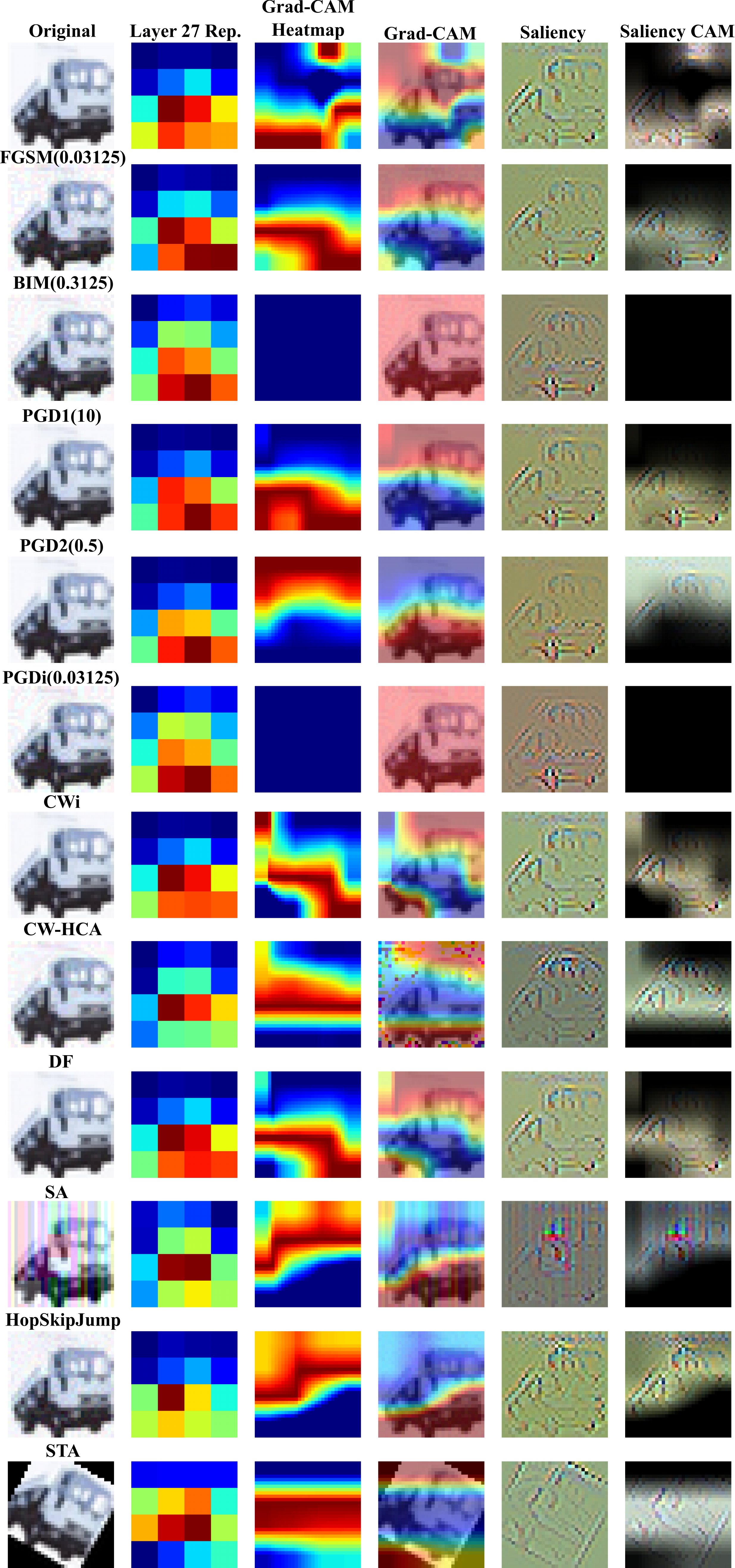}\vspace{-3mm}
    \end{center}
    \caption{CIFAR dataset original and adversarial samples visualization. First column is the original/clean or attacked image sample. The second column is the representative output of the $27^{th}$ layer of baseline classifier. The third column is the heatmap generated with \ac{cam} technique. Fourth column is a combination of the heatmap and the corresponding original or attacked image. The fifth column is the Saliency map generated with Back-propagation technique while the last column is the combination of the saliency and the \ac{cam} heatmap.} \vspace{-3mm}
    \label{fig:adv_cnn_cifar}
\end{figure}

In the adversarial detection systems, we have 4 main players, content of clean images, content of attacked image, the baseline classifier and the detection method in play. In this section, we provide our observations with respect to each player. We follow our demonstration using visualisations that are generated using \ac{cam} \cite{selvaraju2016grad} and back-propagation based saliency \cite{springenberg2014striving}. \ac{cam}  uses the gradients of any target class flowing into the final convolutional layer to produce a coarse localization map highlighting the important regions in the image for predicting the class. Back-propagation based saliency is a variant of the deconvolution approach for visualizing features learned by \acp{cnn}. Figures \ref{fig:orig_cnn_mnist}, \ref{fig:adv_cnn_mnist}, \ref{fig:orig_cnn_cifar} and \ref{fig:adv_cnn_cifar} show different visualization methods for MNIST and CIFAR datasets. First column is the original/clean or attacked image sample. MNIST images displayed with `viridis' colormap. The second column is the representative output of a specified layer of baseline classifier. The third column is the heatmap generated with \ac{cam} technique. Fourth column is a combination of the heatmap and the corresponding original or attacked image. The fifth column is the saliency map generated with back-propagation technique, while the last column is the combination of the saliency and the \ac{cam} heatmap. Many observations can be concluded from these figures: 
\begin{enumerate}[leftmargin=0cm,itemindent=.5cm,labelwidth=\itemindent,labelsep=0cm,align=left]
    \item The saliency regions of MNIST dataset are more\linebreak restricted to the number regions that cover the whole image, while the saliency regions of CIFAR dataset span beyond the target object, i.e., the texture around the target object. Thus, 1) this will maximize the probability of distracting the \ac{cnn} model by giving importance to non-relevant regions, and 2) any small perturbation added to such images will highly affect its saliency and, as a consequence, the \ac{cnn} will target another prediction class.
    \item Looking at the \ac{cam} and guided \ac{cam}, we can notice that the \ac{cnn} does not necessarily use saliency regions of clean images in predicting the correct class. This happens due to the use of complex/over-parametrized classifiers to solve not complex tasks~\cite{madry2017towards}. Hence, \acp{cnn} are vulnerable to small perturbation that will cause higher loss than of clean samples. This can be confirmed as well by looking at the \ac{cam} and the guided \ac{cam} of the \acp{ae} and notice that different small perturbations from different attacks yield different \ac{cam} and guided \ac{cam}. 
    \item Most of \acp{ae} detectors solutions rely on that the representative \ac{cnn} layers output of adversarial input is significantly different from clean input, which is a true assumption. For MNIST, it was easy for most of the detectors to detect the \acp{ae}, while for other datasets, detectors are not able to detect \acp{ae} effectively?  That might be due to 1) dataset has noisy samples, 2) \acp{cnn} either very complex or very over-parametrized, 3) the behaviors of the \ac{cnn} with respect to the content itself of clean and adversarial inputs, for instance, in the case where the guided \ac{cam}, i.e., important regions, of the \ac{ae} is slightly changed from the clean one, the detector work became much harder.
\end{enumerate}

\section{Challenges, future perspectives, and conclusion} \label{sec: challenges}
\subsection{Challenges and future perspectives}

The problem of adversarial examples is not yet solved. Figure \ref{fig:summary} show that most of the detectors are not robust against new/unknown attacks. Moreover, findings from \cite{carlini2017adversarial,athalye2018obfuscated} show that the defenses and the detectors are vulnerable to the carefully designed adversarial perturbations. Hence, we can conclude that, to date, there are no robust defenses and detectors and more investigations are required to identify the features of the \acp{ae}. Such features will facilitate the design of robust defense and detection techniques. Moreover, researchers are facing many questions and challenges that are still open. Here we highlight them:

\textbf{Supervised or Unsupervised detection?} This is a confusing question. Supervised detection methods have, in general, better performance evaluation due to the detector capability to learn from labeled clean and \acp{ae} training data, but this is much restricted to known attacks and cannot be generalized to all kinds of attacks. Relatively, unsupervised detection method is much more flexible since it relies only on the clean data. The main two challenges of the unsupervised approach are 1) tuning its hyper-parameters is very challenging since it cannot be generalized to all models and datasets. 2) finding discriminating features for clean data that are not sensitive to noise and data processing like compression, for instance. As a recommendation, we need to focus on unsupervised detection methods. As discussed in Section \ref{sec: detection_methods}, each unsupervised method proposed a way to detect the \acp{ae} and here is another confusing question, Which one to use? Denoiser based is much more effective if the denoiser has a good estimation of the training data and it does not require manual parameter tuning. Similar to denoiser based approaches, \ac{fs} is highly dependent on the squeezer quality. These two approaches are not computationally expensive but have some inference test time latency. The statistical methods are effective in estimating training data distribution but it is not effective in large-scale applications with large amount of data and classes. It is time consuming in the training phase and has no inference test time latency. Most auxiliary models and \ac{nic} approaches depend on the underlying techniques and they require a lot of hyper-parameters tuning to be suitable for different \ac{dl} based applications. Consequently, we leave the choice of the model to the application requirement.

\textbf{Generalization.} The wide view of generalization is the capability of the detector to detect white, black, gray (transferability) box attacks, and counter-counter attacks. The reported performance in Section \ref{sec: exp_setting} shows that we still need more research efforts to push detectors towards the generalization.

\textbf{Lightweight detection.} Lightweight detector is a detector that 1) has very small overhead, 2) has no inference time latency and 3) is not time consuming in training phase (some industries don't care about that). According to the reported results of the reviewed detector, one of these factors is compromised. Thus, detectors that trade-off between these factors are highly recommended.

\textbf{Ensemble Detection.} It is believed that ensemble techniques can boost one detection technique. For instance,\linebreak \ac{sfad} model used ensemble detection methods,\linebreak selective/uncertainty prediction, softmax based prediction\linebreak and bi-match prediction. These techniques are jointly integrated and give promised results with a little price to \ac{fpr}. Hence, ensemble detection is highly recommended without compromising the \ac{fpr}.

\textbf{Meet the defense.} As discussed in the Section \ref{sec: introduction}, defense techniques, especially robust classification techniques, try to correctly classify input samples, whether attacked or not. Providing defense and detection jointly is an added value to the deployed model since it makes the adversary work much harder. In this approach, the adversary tries to generate \ac{ae} with more perturbation to fool the defense, but in this case, it will be easy for the detector to detect it.

\textbf{Training data Matters?} Yes, the training data plays an important role. Firstly, not noisy training data helps a lot in understanding and recognizing \ac{ae} features and properties in supervised and unsupervised detection methods. Secondly, findings in \cite{schmidt2018adversarially} show that adversarially robust generalization requires more data. Hence, enough data helps in building \ac{cnn} and detectors that generalize well. Finally, the designed \ac{cnn} should be suitable and not complex with respect to the training data and the classification task.

\textbf{Consider high resolution data.} As discussed in Section \ref{sec: high_resolution}, detectors face a huge challenge when they are tested on high resolution data such as ImageNet. They might be not effective, have high \ac{fpr}, have high overhead, or/and are time consuming. Designing effective detectors for high resolution data is highly required.

\subsection{Conclusion} \label{sec: conclusion}
This paper reviewed the detection methods of evasion attacks for neural networks classifiers of image classification task. We firstly categorized the \ac{ae} detection methods into supervised and unsupervised methods. Then, each category is subdivided into statistical, auxiliary models, network invariant, feature squeezing, denoiser-based and object-based methods. Secondly, we demonstrated the performance evaluation of eight state-of-the-art algorithms experimentally in terms of detection rates, false positive rates, complexity, overhead and test time inference latency. We showed, as well, the impact of the content and the \ac{cnn} with respect to detection methods. Finally, we highlighted the fact that the tested algorithms lack generalization and that more research efforts should be made in this research direction. 

\section*{Acknowledgement}
The project is funded by both R\'egion Bretagne (Brittany region), France, and direction g\'en\'erale de l'armement (DGA).

{\small
\bibliographystyle{unsrt}
\bibliography{ref}

\begin{thebibliography}{100}

\bibitem{krizhevsky2012imagenet}
Alex Krizhevsky, Ilya Sutskever, and Geoffrey~E Hinton.
\newblock Imagenet classification with deep convolutional neural networks.
\newblock In {\em Advances in neural information processing systems}, pages
  1097--1105, 2012.

\bibitem{simonyan2014very}
Karen Simonyan and Andrew Zisserman.
\newblock Very deep convolutional networks for large-scale image recognition.
\newblock In Yoshua Bengio and Yann LeCun, editors, {\em 3rd International
  Conference on Learning Representations, {ICLR} 2015, San Diego, CA, USA, May
  7-9, 2015, Conference Track Proceedings}, 2015.

\bibitem{ren2015faster}
Shaoqing Ren, Kaiming He, Ross Girshick, and Jian Sun.
\newblock Faster r-cnn: Towards real-time object detection with region proposal
  networks.
\newblock In {\em Advances in neural information processing systems}, pages
  91--99, 2015.

\bibitem{long2015fully}
Jonathan Long, Evan Shelhamer, and Trevor Darrell.
\newblock Fully convolutional networks for semantic segmentation.
\newblock In {\em Proceedings of the IEEE conference on computer vision and
  pattern recognition}, pages 3431--3440, 2015.

\bibitem{bertinetto2016fully}
Luca Bertinetto, Jack Valmadre, Joao~F Henriques, Andrea Vedaldi, and Philip~HS
  Torr.
\newblock Fully-convolutional siamese networks for object tracking.
\newblock In {\em European conference on computer vision}, pages 850--865.
  Springer, 2016.

\bibitem{danelljan2017eco}
Martin Danelljan, Goutam Bhat, Fahad Shahbaz~Khan, and Michael Felsberg.
\newblock Eco: Efficient convolution operators for tracking.
\newblock In {\em Proceedings of the IEEE conference on computer vision and
  pattern recognition}, pages 6638--6646, 2017.

\bibitem{ker2017deep}
Justin Ker, Lipo Wang, Jai Rao, and Tchoyoson Lim.
\newblock Deep learning applications in medical image analysis.
\newblock {\em IEEE Access}, 6:9375--9389, 2017.

\bibitem{bahdanau2014neural}
Dzmitry Bahdanau, Kyunghyun Cho, and Yoshua Bengio.
\newblock Neural machine translation by jointly learning to align and
  translate.
\newblock In Yoshua Bengio and Yann LeCun, editors, {\em 3rd International
  Conference on Learning Representations, {ICLR} 2015, San Diego, CA, USA, May
  7-9, 2015, Conference Track Proceedings}, 2015.

\bibitem{hannun2014deep}
Awni~Y. Hannun, Carl Case, Jared Casper, Bryan Catanzaro, Greg Diamos, Erich
  Elsen, Ryan Prenger, Sanjeev Satheesh, Shubho Sengupta, Adam Coates, and
  Andrew~Y. Ng.
\newblock Deep speech: Scaling up end-to-end speech recognition.
\newblock {\em CoRR}, abs/1412.5567, 2014.

\bibitem{kurakin2016adversarial}
Alexey Kurakin, Ian Goodfellow, and Samy Bengio.
\newblock Adversarial examples in the physical world.
\newblock {\em ICLR Workshop}, 2017.

\bibitem{evtimov2017robust}
Ivan Evtimov, Kevin Eykholt, Earlence Fernandes, Tadayoshi Kohno, Bo~Li, Atul
  Prakash, Amir Rahmati, and Dawn Song.
\newblock Robust physical-world attacks on machine learning models.
\newblock {\em CoRR}, abs/1707.08945, 2017.

\bibitem{gu2017badnets}
Tianyu Gu, Brendan Dolan{-}Gavitt, and Siddharth Garg.
\newblock Badnets: Identifying vulnerabilities in the machine learning model
  supply chain.
\newblock {\em CoRR}, abs/1708.06733, 2017.

\bibitem{papernot2017practical}
Nicolas Papernot, Patrick McDaniel, Ian Goodfellow, Somesh Jha, Z~Berkay Celik,
  and Ananthram Swami.
\newblock Practical black-box attacks against machine learning.
\newblock In {\em Proceedings of the 2017 ACM on Asia conference on computer
  and communications security}, pages 506--519, 2017.

\bibitem{melis2017deep}
Marco Melis, Ambra Demontis, Battista Biggio, Gavin Brown, Giorgio Fumera, and
  Fabio Roli.
\newblock Is deep learning safe for robot vision? adversarial examples against
  the icub humanoid.
\newblock In {\em Proceedings of the IEEE International Conference on Computer
  Vision Workshops}, pages 751--759, 2017.

\bibitem{o2015introduction}
Keiron O'Shea and Ryan Nash.
\newblock An introduction to convolutional neural networks.
\newblock {\em arXiv preprint arXiv:1511.08458}, 2015.

\bibitem{he2016deep}
Kaiming He, Xiangyu Zhang, Shaoqing Ren, and Jian Sun.
\newblock Deep residual learning for image recognition.
\newblock In {\em Proceedings of the IEEE conference on computer vision and
  pattern recognition}, pages 770--778, 2016.

\bibitem{szegedy2016rethinking}
Christian Szegedy, Vincent Vanhoucke, Sergey Ioffe, Jonathon Shlens, and
  Zbigniew Wojna.
\newblock Rethinking the inception architecture for computer vision.
\newblock In {\em 2016 {IEEE} Conference on Computer Vision and Pattern
  Recognition, {CVPR} 2016, Las Vegas, NV, USA, June 27-30, 2016}, pages
  2818--2826. {IEEE} Computer Society, 2016.

\bibitem{howard2017mobilenets}
Andrew~G. Howard, Menglong Zhu, Bo~Chen, Dmitry Kalenichenko, Weijun Wang,
  Tobias Weyand, Marco Andreetto, and Hartwig Adam.
\newblock Mobilenets: Efficient convolutional neural networks for mobile vision
  applications.
\newblock {\em CoRR}, abs/1704.04861, 2017.

\bibitem{lecun1998gradient}
Yann LeCun, L{\'e}on Bottou, Yoshua Bengio, and Patrick Haffner.
\newblock Gradient-based learning applied to document recognition.
\newblock {\em Proceedings of the IEEE}, 86(11):2278--2324, 1998.

\bibitem{krizhevsky2009learning}
A.~Krizhevsky and G.~Hinton.
\newblock Learning multiple layers of features from tiny images.
\newblock {\em Master's thesis, Department of Computer Science, University of
  Toronto}, 2009.

\bibitem{netzer2011reading}
Yuval Netzer, Tao Wang, Adam Coates, Alessandro Bissacco, Bo~Wu, and Andrew~Y.
  Ng.
\newblock Reading digits in natural images with unsupervised feature learning.
\newblock In {\em NIPS Workshop on Deep Learning and Unsupervised Feature
  Learning 2011}, Granada, Spain, 2011.

\bibitem{yao2015tiny}
Leon Yao and John Miller.
\newblock Tiny imagenet classification with convolutional neural networks.
\newblock {\em CS 231N}, 2(5):8, 2015.

\bibitem{imagenet_cvpr09}
J.~Deng, W.~Dong, R.~Socher, L.-J. Li, K.~Li, and L.~Fei-Fei.
\newblock {ImageNet: A Large-Scale Hierarchical Image Database}.
\newblock In {\em CVPR09}, 2009.

\bibitem{girshick2014rich}
Ross Girshick, Jeff Donahue, Trevor Darrell, and Jitendra Malik.
\newblock Rich feature hierarchies for accurate object detection and semantic
  segmentation.
\newblock In {\em Proceedings of the IEEE conference on computer vision and
  pattern recognition}, pages 580--587, 2014.

\bibitem{girshick2015fast}
Ross~B. Girshick.
\newblock Fast {R-CNN}.
\newblock In {\em 2015 {IEEE} International Conference on Computer Vision,
  {ICCV} 2015, Santiago, Chile, December 7-13, 2015}, pages 1440--1448. {IEEE}
  Computer Society, 2015.

\bibitem{redmon2016you}
Joseph Redmon, Santosh~Kumar Divvala, Ross~B. Girshick, and Ali Farhadi.
\newblock You only look once: Unified, real-time object detection.
\newblock In {\em 2016 {IEEE} Conference on Computer Vision and Pattern
  Recognition, {CVPR} 2016, Las Vegas, NV, USA, June 27-30, 2016}, pages
  779--788. {IEEE} Computer Society, 2016.

\bibitem{devlin2019bert}
Jacob Devlin, Ming{-}Wei Chang, Kenton Lee, and Kristina Toutanova.
\newblock {BERT:} pre-training of deep bidirectional transformers for language
  understanding.
\newblock In {\em Proceedings of the 2019 Conference of the North American
  Chapter of the Association for Computational Linguistics: Human Language
  Technologies, {NAACL-HLT} 2019, Minneapolis, MN, USA, June 2-7, 2019, Volume
  1 (Long and Short Papers)}, pages 4171--4186. Association for Computational
  Linguistics, 2019.

\bibitem{yang2020xlnet}
Zhilin Yang, Zihang Dai, Yiming Yang, Jaime~G. Carbonell, Ruslan Salakhutdinov,
  and Quoc~V. Le.
\newblock Xlnet: Generalized autoregressive pretraining for language
  understanding.
\newblock In {\em Advances in Neural Information Processing Systems 32: Annual
  Conference on Neural Information Processing Systems 2019, NeurIPS 2019,
  December 8-14, 2019, Vancouver, BC, Canada}, pages 5754--5764, 2019.

\bibitem{lan2020albert}
Zhenzhong Lan, Mingda Chen, Sebastian Goodman, Kevin Gimpel, Piyush Sharma, and
  Radu Soricut.
\newblock {ALBERT:} {A} lite {BERT} for self-supervised learning of language
  representations.
\newblock In {\em 8th International Conference on Learning Representations,
  {ICLR} 2020, Addis Ababa, Ethiopia, April 26-30, 2020}. OpenReview.net, 2020.

\bibitem{PITROPAKIS2019100199}
Nikolaos Pitropakis, Emmanouil Panaousis, Thanassis Giannetsos, Eleftherios
  Anastasiadis, and George Loukas.
\newblock A taxonomy and survey of attacks against machine learning.
\newblock {\em Computer Science Review}, 34:100199, 2019.

\bibitem{li2020backdoor}
Yiming Li, Baoyuan Wu, Yong Jiang, Zhifeng Li, and Shu-Tao Xia.
\newblock Backdoor learning: A survey.
\newblock {\em arXiv preprint arXiv:2007.08745}, 2020.

\bibitem{szegedy2013intriguing}
Christian Szegedy, Wojciech Zaremba, Ilya Sutskever, Joan Bruna, Dumitru Erhan,
  Ian~J. Goodfellow, and Rob Fergus.
\newblock Intriguing properties of neural networks.
\newblock In Yoshua Bengio and Yann LeCun, editors, {\em 2nd International
  Conference on Learning Representations, {ICLR} 2014, Banff, AB, Canada, April
  14-16, 2014, Conference Track Proceedings}, 2014.

\bibitem{carlini2017adversarial}
Nicholas Carlini and David Wagner.
\newblock Adversarial examples are not easily detected: Bypassing ten detection
  methods.
\newblock In {\em Proceedings of the 10th ACM Workshop on Artificial
  Intelligence and Security}, pages 3--14, 2017.

\bibitem{ilyas2019adversarial}
Andrew Ilyas, Shibani Santurkar, Dimitris Tsipras, Logan Engstrom, Brandon
  Tran, and Aleksander Madry.
\newblock Adversarial examples are not bugs, they are features.
\newblock In {\em Advances in Neural Information Processing Systems}, pages
  125--136, 2019.

\bibitem{akhtar2018threat}
Naveed Akhtar and Ajmal Mian.
\newblock Threat of adversarial attacks on deep learning in computer vision: A
  survey.
\newblock {\em IEEE Access}, 6:14410--14430, 2018.

\bibitem{hao2020adversarial}
Han Xu Yao~Ma Hao-Chen, Liu~Debayan Deb, Hui Liu Ji-Liang~Tang Anil, and
  K~Jain.
\newblock Adversarial attacks and defenses in images, graphs and text: A
  review.
\newblock {\em International Journal of Automation and Computing},
  17(2):151--178, 2020.

\bibitem{goodfellow2014explaining}
Ian~J. Goodfellow, Jonathon Shlens, and Christian Szegedy.
\newblock Explaining and harnessing adversarial examples.
\newblock In Yoshua Bengio and Yann LeCun, editors, {\em 3rd International
  Conference on Learning Representations, {ICLR} 2015, San Diego, CA, USA, May
  7-9, 2015, Conference Track Proceedings}, 2015.

\bibitem{moosavi2016deepfool}
Seyed-Mohsen Moosavi-Dezfooli, Alhussein Fawzi, and Pascal Frossard.
\newblock Deepfool: a simple and accurate method to fool deep neural networks.
\newblock In {\em Proceedings of the IEEE conference on computer vision and
  pattern recognition}, pages 2574--2582, 2016.

\bibitem{carlini2017towards}
Nicholas Carlini and David Wagner.
\newblock Towards evaluating the robustness of neural networks.
\newblock In {\em 2017 ieee symposium on security and privacy (sp)}, pages
  39--57. IEEE, 2017.

\bibitem{madry2017towards}
Aleksander Madry, Aleksandar Makelov, Ludwig Schmidt, Dimitris Tsipras, and
  Adrian Vladu.
\newblock Towards deep learning models resistant to adversarial attacks.
\newblock In {\em 6th International Conference on Learning Representations,
  {ICLR} 2018, Vancouver, BC, Canada, April 30 - May 3, 2018, Conference Track
  Proceedings}. OpenReview.net, 2018.

\bibitem{papernot2016transferability}
Nicolas Papernot, Patrick~D. McDaniel, and Ian~J. Goodfellow.
\newblock Transferability in machine learning: from phenomena to black-box
  attacks using adversarial samples.
\newblock {\em CoRR}, abs/1605.07277, 2016.

\bibitem{chen2017zoo}
Pin-Yu Chen, Huan Zhang, Yash Sharma, Jinfeng Yi, and Cho-Jui Hsieh.
\newblock Zoo: Zeroth order optimization based black-box attacks to deep neural
  networks without training substitute models.
\newblock In {\em Proceedings of the 10th ACM Workshop on Artificial
  Intelligence and Security}, pages 15--26, 2017.

\bibitem{engstrom2019exploring}
Logan Engstrom, Brandon Tran, Dimitris Tsipras, Ludwig Schmidt, and Aleksander
  Madry.
\newblock Exploring the landscape of spatial robustness.
\newblock In {\em International Conference on Machine Learning}, pages
  1802--1811, 2019.

\bibitem{su2019one}
Jiawei Su, Danilo~Vasconcellos Vargas, and Kouichi Sakurai.
\newblock One pixel attack for fooling deep neural networks.
\newblock {\em IEEE Transactions on Evolutionary Computation}, 23(5):828--841,
  2019.

\bibitem{kotyan2019adversarial}
Shashank Kotyan and Danilo Vasconcellos~Vargas.
\newblock Adversarial robustness assessment: Why both $l_0$ and $l_\infty$
  attacks are necessary.
\newblock {\em arXiv e-prints}, pages arXiv--1906, 2019.

\bibitem{xie2017adversarial}
Cihang Xie, Jianyu Wang, Zhishuai Zhang, Yuyin Zhou, Lingxi Xie, and Alan~L.
  Yuille.
\newblock Adversarial examples for semantic segmentation and object detection.
\newblock In {\em {IEEE} International Conference on Computer Vision, {ICCV}
  2017, Venice, Italy, October 22-29, 2017}, pages 1378--1387. {IEEE} Computer
  Society, 2017.

\bibitem{lu2017no}
Jiajun Lu, Hussein Sibai, Evan Fabry, and David~A. Forsyth.
\newblock {NO} need to worry about adversarial examples in object detection in
  autonomous vehicles.
\newblock {\em CoRR}, abs/1707.03501, 2017.

\bibitem{zhang2020adversarial}
Wei~Emma Zhang, Quan~Z Sheng, Ahoud Alhazmi, and Chenliang Li.
\newblock Adversarial attacks on deep-learning models in natural language
  processing: A survey.
\newblock {\em ACM Transactions on Intelligent Systems and Technology (TIST)},
  11(3):1--41, 2020.

\bibitem{sun2020advbert}
Lichao Sun, Kazuma Hashimoto, Wenpeng Yin, Akari Asai, Jia Li, Philip~S. Yu,
  and Caiming Xiong.
\newblock Adv-bert: {BERT} is not robust on misspellings! generating nature
  adversarial samples on {BERT}.
\newblock {\em CoRR}, abs/2003.04985, 2020.

\bibitem{li2019universal}
Di~Li, Danilo~Vasconcellos Vargas, and Kouichi Sakurai.
\newblock Universal rules for fooling deep neural networks based text
  classification.
\newblock In {\em {IEEE} Congress on Evolutionary Computation, {CEC} 2019,
  Wellington, New Zealand, June 10-13, 2019}, pages 2221--2228. {IEEE}, 2019.

\bibitem{wang2020adversarial}
Donghua Wang, Rangding Wang, Li~Dong, Diqun Yan, Xueyuan Zhang, and Yongkang
  Gong.
\newblock Adversarial examples attack and countermeasure for speech recognition
  system: A survey.
\newblock In {\em International Conference on Security and Privacy in Digital
  Economy}, pages 443--468. Springer, 2020.

\bibitem{ren2021adversarial}
Huali Ren, Teng Huang, and Hongyang Yan.
\newblock Adversarial examples: attacks and defenses in the physical world.
\newblock {\em International Journal of Machine Learning and Cybernetics},
  pages 1--12, 2021.

\bibitem{dasgupta2019survey}
Prithviraj Dasgupta and Joseph Collins.
\newblock A survey of game theoretic approaches for adversarial machine
  learning in cybersecurity tasks.
\newblock {\em AI Magazine}, 40(2):31--43, 2019.

\bibitem{finlayson2018adversarial}
Samuel~G. Finlayson, Isaac~S. Kohane, and Andrew~L. Beam.
\newblock Adversarial attacks against medical deep learning systems.
\newblock {\em CoRR}, abs/1804.05296, 2018.

\bibitem{xie2020smooth}
Cihang Xie, Mingxing Tan, Boqing Gong, Alan~L. Yuille, and Quoc~V. Le.
\newblock Smooth adversarial training.
\newblock {\em CoRR}, abs/2006.14536, 2020.

\bibitem{tramer2017ensemble}
Florian Tram{\`{e}}r, Alexey Kurakin, Nicolas Papernot, Ian~J. Goodfellow, Dan
  Boneh, and Patrick~D. McDaniel.
\newblock Ensemble adversarial training: Attacks and defenses.
\newblock In {\em 6th International Conference on Learning Representations,
  {ICLR} 2018, Vancouver, BC, Canada, April 30 - May 3, 2018, Conference Track
  Proceedings}. OpenReview.net, 2018.

\bibitem{xie2019feature}
Cihang Xie, Yuxin Wu, Laurens van~der Maaten, Alan~L Yuille, and Kaiming He.
\newblock Feature denoising for improving adversarial robustness.
\newblock In {\em Proceedings of the IEEE Conference on Computer Vision and
  Pattern Recognition}, pages 501--509, 2019.

\bibitem{borkar2020defending}
Tejas Borkar, Felix Heide, and Lina Karam.
\newblock Defending against universal attacks through selective feature
  regeneration.
\newblock In {\em Proceedings of the IEEE/CVF Conference on Computer Vision and
  Pattern Recognition}, pages 709--719, 2020.

\bibitem{liao2018defense}
Fangzhou Liao, Ming Liang, Yinpeng Dong, Tianyu Pang, Xiaolin Hu, and Jun Zhu.
\newblock Defense against adversarial attacks using high-level representation
  guided denoiser.
\newblock In {\em Proceedings of the IEEE Conference on Computer Vision and
  Pattern Recognition}, pages 1778--1787, 2018.

\bibitem{bakhti2019ddsa}
Yassine Bakhti, Sid~Ahmed Fezza, Wassim Hamidouche, and Olivier D{\'e}forges.
\newblock Ddsa: a defense against adversarial attacks using deep denoising
  sparse autoencoder.
\newblock {\em IEEE Access}, 7:160397--160407, 2019.

\bibitem{mustafa2019image}
Aamir Mustafa, Salman~H Khan, Munawar Hayat, Jianbing Shen, and Ling Shao.
\newblock Image super-resolution as a defense against adversarial attacks.
\newblock {\em IEEE Transactions on Image Processing}, 29:1711--1724, 2019.

\bibitem{prakash2018deflecting}
Aaditya Prakash, Nick Moran, Solomon Garber, Antonella DiLillo, and James
  Storer.
\newblock Deflecting adversarial attacks with pixel deflection.
\newblock In {\em Proceedings of the IEEE conference on computer vision and
  pattern recognition}, pages 8571--8580, 2018.

\bibitem{papernot2016distillation}
Nicolas Papernot, Patrick McDaniel, Xi~Wu, Somesh Jha, and Ananthram Swami.
\newblock Distillation as a defense to adversarial perturbations against deep
  neural networks.
\newblock In {\em 2016 IEEE Symposium on Security and Privacy (SP)}, pages
  582--597. IEEE, 2016.

\bibitem{gu2014towards}
Shixiang Gu and Luca Rigazio.
\newblock Towards deep neural network architectures robust to adversarial
  examples.
\newblock In Yoshua Bengio and Yann LeCun, editors, {\em 3rd International
  Conference on Learning Representations, {ICLR} 2015, San Diego, CA, USA, May
  7-9, 2015, Workshop Track Proceedings}, 2015.

\bibitem{nayebi2017biologically}
Aran Nayebi and Surya Ganguli.
\newblock Biologically inspired protection of deep networks from adversarial
  attacks.
\newblock {\em CoRR}, abs/1703.09202, 2017.

\bibitem{grosse2017statistical}
Kathrin Grosse, Praveen Manoharan, Nicolas Papernot, Michael Backes, and
  Patrick~D. McDaniel.
\newblock On the (statistical) detection of adversarial examples.
\newblock {\em CoRR}, abs/1702.06280, 2017.

\bibitem{meng2017magnet}
Dongyu Meng and Hao Chen.
\newblock Magnet: a two-pronged defense against adversarial examples.
\newblock In {\em Proceedings of the 2017 ACM SIGSAC conference on computer and
  communications security}, pages 135--147, 2017.

\bibitem{xu2017feature}
Weilin Xu, David Evans, and Yanjun Qi.
\newblock Feature squeezing: Detecting adversarial examples in deep neural
  networks.
\newblock In {\em 25th Annual Network and Distributed System Security
  Symposium, {NDSS} 2018, San Diego, California, USA, February 18-21, 2018}.
  The Internet Society, 2018.

\bibitem{grosse2016adversarial}
Kathrin Grosse, Nicolas Papernot, Praveen Manoharan, Michael Backes, and
  Patrick~D. McDaniel.
\newblock Adversarial perturbations against deep neural networks for malware
  classification.
\newblock {\em CoRR}, abs/1606.04435, 2016.

\bibitem{ma2019nic}
Shiqing Ma and Yingqi Liu.
\newblock Nic: Detecting adversarial samples with neural network invariant
  checking.
\newblock In {\em Proceedings of the 26th Network and Distributed System
  Security Symposium (NDSS 2019)}, 2019.

\bibitem{ortiz2020optimism}
Guillermo Ortiz-Jiménez, Apostolos Modas, Seyed-Mohsen Moosavi-Dezfooli, and
  Pascal Frossard.
\newblock Optimism in the face of adversity: Understanding and improving deep
  learning through adversarial robustness.
\newblock {\em Proceedings of the IEEE}, 109(5):635--659, 2021.

\bibitem{yuan2019adversarial}
Xiaoyong Yuan, Pan He, Qile Zhu, and Xiaolin Li.
\newblock Adversarial examples: Attacks and defenses for deep learning.
\newblock {\em IEEE transactions on neural networks and learning systems},
  30(9):2805--2824, 2019.

\bibitem{chakraborty2018adversarial}
Anirban Chakraborty, Manaar Alam, Vishal Dey, Anupam Chattopadhyay, and Debdeep
  Mukhopadhyay.
\newblock Adversarial attacks and defences: {A} survey.
\newblock {\em CoRR}, abs/1810.00069, 2018.

\bibitem{wang2019security}
Xianmin Wang, Jing Li, Xiaohui Kuang, Yu-an Tan, and Jin Li.
\newblock The security of machine learning in an adversarial setting: A survey.
\newblock {\em Journal of Parallel and Distributed Computing}, 130:12--23,
  2019.

\bibitem{machado2020adversarial}
Gabriel~Resende Machado, Eug{\^{e}}nio Silva, and Ronaldo~Ribeiro Goldschmidt.
\newblock Adversarial machine learning in image classification: {A} survey
  towards the defender's perspective.
\newblock {\em CoRR}, abs/2009.03728, 2020.

\bibitem{bulusu2020anomalous}
Saikiran Bulusu, Bhavya Kailkhura, Bo~Li, Pramod~K Varshney, and Dawn Song.
\newblock Anomalous example detection in deep learning: A survey.
\newblock {\em IEEE Access}, 8:132330--132347, 2020.

\bibitem{miller2019not}
David Miller, Yujia Wang, and George Kesidis.
\newblock When not to classify: Anomaly detection of attacks (ada) on dnn
  classifiers at test time.
\newblock {\em Neural computation}, 31(8):1624--1670, 2019.

\bibitem{miller2020adversarial}
David~J Miller, Zhen Xiang, and George Kesidis.
\newblock Adversarial learning targeting deep neural network classification: A
  comprehensive review of defenses against attacks.
\newblock {\em Proceedings of the IEEE}, 108(3):402--433, 2020.

\bibitem{serban2020adversarial}
Alex Serban, Erik Poll, and Joost Visser.
\newblock Adversarial examples on object recognition: A comprehensive survey.
\newblock {\em ACM Computing Surveys (CSUR)}, 53(3):1--38, 2020.

\bibitem{biggio2013evasion}
Battista Biggio, Igino Corona, Davide Maiorca, Blaine Nelson, Nedim
  {\v{S}}rndi{\'c}, Pavel Laskov, Giorgio Giacinto, and Fabio Roli.
\newblock Evasion attacks against machine learning at test time.
\newblock In {\em Joint European conference on machine learning and knowledge
  discovery in databases}, pages 387--402. Springer, 2013.

\bibitem{liu2016delving}
Yanpei Liu, Xinyun Chen, Chang Liu, and Dawn Song.
\newblock Delving into transferable adversarial examples and black-box attacks.
\newblock In {\em 5th International Conference on Learning Representations,
  {ICLR} 2017, Toulon, France, April 24-26, 2017, Conference Track
  Proceedings}. OpenReview.net, 2017.

\bibitem{biggio2014pattern}
Battista Biggio, Giorgio Fumera, and Fabio Roli.
\newblock Pattern recognition systems under attack: Design issues and research
  challenges.
\newblock {\em International Journal of Pattern Recognition and Artificial
  Intelligence}, 28(07):1460002, 2014.

\bibitem{biggio2014security}
Battista Biggio, Igino Corona, Blaine Nelson, Benjamin~IP Rubinstein, Davide
  Maiorca, Giorgio Fumera, Giorgio Giacinto, and Fabio Roli.
\newblock Security evaluation of support vector machines in adversarial
  environments.
\newblock In {\em Support Vector Machines Applications}, pages 105--153.
  Springer, 2014.

\bibitem{biggio2018wild}
Battista Biggio and Fabio Roli.
\newblock Wild patterns: Ten years after the rise of adversarial machine
  learning.
\newblock {\em Pattern Recognition}, 84:317--331, 2018.

\bibitem{liu1989limited}
Dong~C Liu and Jorge Nocedal.
\newblock On the limited memory bfgs method for large scale optimization.
\newblock {\em Mathematical programming}, 45(1-3):503--528, 1989.

\bibitem{croce2020reliable}
Francesco Croce and Matthias Hein.
\newblock Reliable evaluation of adversarial robustness with an ensemble of
  diverse parameter-free attacks.
\newblock In {\em Proceedings of the 37th International Conference on Machine
  Learning, {ICML} 2020, 13-18 July 2020, Virtual Event}, volume 119 of {\em
  Proceedings of Machine Learning Research}, pages 2206--2216. {PMLR}, 2020.

\bibitem{moosavi2017universal}
Seyed-Mohsen Moosavi-Dezfooli, Alhussein Fawzi, Omar Fawzi, and Pascal
  Frossard.
\newblock Universal adversarial perturbations.
\newblock In {\em Proceedings of the IEEE conference on computer vision and
  pattern recognition}, pages 1765--1773, 2017.

\bibitem{papernot2016limitations}
Nicolas Papernot, Patrick McDaniel, Somesh Jha, Matt Fredrikson, Z~Berkay
  Celik, and Ananthram Swami.
\newblock The limitations of deep learning in adversarial settings.
\newblock In {\em 2016 IEEE European symposium on security and privacy
  (EuroS\&P)}, pages 372--387. IEEE, 2016.

\bibitem{sabour2015adversarial}
Sara Sabour, Yanshuai Cao, Fartash Faghri, and David~J. Fleet.
\newblock Adversarial manipulation of deep representations.
\newblock In Yoshua Bengio and Yann LeCun, editors, {\em 4th International
  Conference on Learning Representations, {ICLR} 2016, San Juan, Puerto Rico,
  May 2-4, 2016, Conference Track Proceedings}, 2016.

\bibitem{baluja2017adversarial}
Shumeet Baluja and Ian Fischer.
\newblock Adversarial transformation networks: Learning to generate adversarial
  examples.
\newblock {\em CoRR}, abs/1703.09387, 2017.

\bibitem{athalye2018synthesizing}
Anish Athalye, Logan Engstrom, Andrew Ilyas, and Kevin Kwok.
\newblock Synthesizing robust adversarial examples.
\newblock In {\em International conference on machine learning}, pages
  284--293. PMLR, 2018.

\bibitem{athalye2018obfuscated}
Anish Athalye, Nicholas Carlini, and David~A. Wagner.
\newblock Obfuscated gradients give a false sense of security: Circumventing
  defenses to adversarial examples.
\newblock In Jennifer~G. Dy and Andreas Krause, editors, {\em Proceedings of
  the 35th International Conference on Machine Learning, {ICML} 2018,
  Stockholmsm{\"{a}}ssan, Stockholm, Sweden, July 10-15, 2018}, volume~80 of
  {\em Proceedings of Machine Learning Research}, pages 274--283. {PMLR}, 2018.

\bibitem{andriushchenko2020square}
Maksym Andriushchenko, Francesco Croce, Nicolas Flammarion, and Matthias Hein.
\newblock Square attack: {A} query-efficient black-box adversarial attack via
  random search.
\newblock In Andrea Vedaldi, Horst Bischof, Thomas Brox, and Jan{-}Michael
  Frahm, editors, {\em Computer Vision - {ECCV} 2020 - 16th European
  Conference, Glasgow, UK, August 23-28, 2020, Proceedings, Part {XXIII}},
  volume 12368 of {\em Lecture Notes in Computer Science}, pages 484--501.
  Springer, 2020.

\bibitem{brendel2017decision}
Wieland Brendel, Jonas Rauber, and Matthias Bethge.
\newblock Decision-based adversarial attacks: Reliable attacks against
  black-box machine learning models.
\newblock In {\em 6th International Conference on Learning Representations,
  {ICLR} 2018, Vancouver, BC, Canada, April 30 - May 3, 2018, Conference Track
  Proceedings}. OpenReview.net, 2018.

\bibitem{chen2020hopskipjumpattack}
Jianbo Chen, Michael~I Jordan, and Martin~J Wainwright.
\newblock Hopskipjumpattack: A query-efficient decision-based attack.
\newblock In {\em 2020 ieee symposium on security and privacy (sp)}, pages
  1277--1294. IEEE, 2020.

\bibitem{sarkar2017upset}
Sayantan Sarkar, Ankan Bansal, Upal Mahbub, and Rama Chellappa.
\newblock {UPSET} and {ANGRI} : Breaking high performance image classifiers.
\newblock {\em CoRR}, abs/1707.01159, 2017.

\bibitem{nguyen2015deep}
Anh Nguyen, Jason Yosinski, and Jeff Clune.
\newblock Deep neural networks are easily fooled: High confidence predictions
  for unrecognizable images.
\newblock In {\em Proceedings of the IEEE conference on computer vision and
  pattern recognition}, pages 427--436, 2015.

\bibitem{carlini2016defensive}
Nicholas Carlini and David~A. Wagner.
\newblock Defensive distillation is not robust to adversarial examples.
\newblock {\em CoRR}, abs/1607.04311, 2016.

\bibitem{lu2017safetynet}
Jiajun Lu, Theerasit Issaranon, and David Forsyth.
\newblock Safetynet: Detecting and rejecting adversarial examples robustly.
\newblock In {\em Proceedings of the IEEE International Conference on Computer
  Vision}, pages 446--454, 2017.

\bibitem{pertigkiozoglou2018detecting}
Stefanos Pertigkiozoglou and Petros Maragos.
\newblock Detecting adversarial examples in convolutional neural networks.
\newblock {\em CoRR}, abs/1812.03303, 2018.

\bibitem{carrara2018adversarial}
Fabio Carrara, Rudy Becarelli, Roberto Caldelli, Fabrizio Falchi, and Giuseppe
  Amato.
\newblock Adversarial examples detection in features distance spaces.
\newblock In {\em Proceedings of the European Conference on Computer Vision
  (ECCV)}, pages 0--0, 2018.

\bibitem{metzen2017detecting}
Jan~Hendrik Metzen, Tim Genewein, Volker Fischer, and Bastian Bischoff.
\newblock On detecting adversarial perturbations.
\newblock In {\em 5th International Conference on Learning Representations,
  {ICLR} 2017, Toulon, France, April 24-26, 2017, Conference Track
  Proceedings}. OpenReview.net, 2017.

\bibitem{eniser2020raid}
Hasan~Ferit Eniser, Maria Christakis, and Valentin W{\"{u}}stholz.
\newblock {RAID:} randomized adversarial-input detection for neural networks.
\newblock {\em CoRR}, abs/2002.02776, 2020.

\bibitem{feinman2017detecting}
Reuben Feinman, Ryan~R. Curtin, Saurabh Shintre, and Andrew~B. Gardner.
\newblock Detecting adversarial samples from artifacts.
\newblock {\em CoRR}, abs/1703.00410, 2017.

\bibitem{smith2018understanding}
Lewis Smith and Yarin Gal.
\newblock Understanding measures of uncertainty for adversarial example
  detection.
\newblock In Amir Globerson and Ricardo Silva, editors, {\em Proceedings of the
  Thirty-Fourth Conference on Uncertainty in Artificial Intelligence, {UAI}
  2018, Monterey, California, USA, August 6-10, 2018}, pages 560--569. {AUAI}
  Press, 2018.

\bibitem{hendrycks2016baseline}
Dan Hendrycks and Kevin Gimpel.
\newblock A baseline for detecting misclassified and out-of-distribution
  examples in neural networks.
\newblock In {\em 5th International Conference on Learning Representations,
  {ICLR} 2017, Toulon, France, April 24-26, 2017, Conference Track
  Proceedings}. OpenReview.net, 2017.

\bibitem{aigrain2019detecting}
Jonathan Aigrain and Marcin Detyniecki.
\newblock Detecting adversarial examples and other misclassifications in neural
  networks by introspection.
\newblock {\em CoRR}, abs/1905.09186, 2019.

\bibitem{monteiro2019generalizable}
Jo{\~a}o Monteiro, Isabela Albuquerque, Zahid Akhtar, and Tiago~H Falk.
\newblock Generalizable adversarial examples detection based on bi-model
  decision mismatch.
\newblock In {\em 2019 IEEE International Conference on Systems, Man and
  Cybernetics (SMC)}, pages 2839--2844. IEEE, 2019.

\bibitem{gong2017adversarial}
Zhitao Gong, Wenlu Wang, and Wei{-}Shinn Ku.
\newblock Adversarial and clean data are not twins.
\newblock {\em CoRR}, abs/1704.04960, 2017.

\bibitem{hosseini2017blocking}
Hossein Hosseini, Yize Chen, Sreeram Kannan, Baosen Zhang, and Radha
  Poovendran.
\newblock Blocking transferability of adversarial examples in black-box
  learning systems.
\newblock {\em CoRR}, abs/1703.04318, 2017.

\bibitem{lust2020gran}
Julia Lust and Alexandru~Paul Condurache.
\newblock Gran: An efficient gradient-norm based detector for adversarial and
  misclassified examples.
\newblock In {\em 28th European Symposium on Artificial Neural Networks,
  Computational Intelligence and Machine Learning, {ESANN} 2020, Bruges,
  Belgium, October 2-4, 2020}, pages 7--12, 2020.

\bibitem{kherchouche2020detection}
Anouar Kherchouche, Sid~Ahmed Fezza, Wassim Hamidouche, and Olivier
  D{\'e}forges.
\newblock Detection of adversarial examples in deep neural networks with
  natural scene statistics.
\newblock In {\em 2020 International Joint Conference on Neural Networks
  (IJCNN)}, pages 1--7. IEEE, 2020.

\bibitem{zuo2020exploiting}
Fei Zuo and Qiang Zeng.
\newblock Exploiting the sensitivity of {L2} adversarial examples to
  erase-and-restore.
\newblock In Jiannong Cao, Man~Ho Au, Zhiqiang Lin, and Moti Yung, editors,
  {\em {ASIA} {CCS} '21: {ACM} Asia Conference on Computer and Communications
  Security, Virtual Event, Hong Kong, June 7-11, 2021}, pages 40--51. {ACM},
  2021.

\bibitem{li2017adversarial}
Xin Li and Fuxin Li.
\newblock Adversarial examples detection in deep networks with convolutional
  filter statistics.
\newblock In {\em Proceedings of the IEEE International Conference on Computer
  Vision}, pages 5764--5772, 2017.

\bibitem{ma2018characterizing}
Xingjun Ma, Bo~Li, Yisen Wang, Sarah~M. Erfani, Sudanthi N.~R. Wijewickrema,
  Grant Schoenebeck, Dawn Song, Michael~E. Houle, and James Bailey.
\newblock Characterizing adversarial subspaces using local intrinsic
  dimensionality.
\newblock In {\em 6th International Conference on Learning Representations,
  {ICLR} 2018, Vancouver, BC, Canada, April 30 - May 3, 2018, Conference Track
  Proceedings}. OpenReview.net, 2018.

\bibitem{cohen2020detecting}
Gilad Cohen, Guillermo Sapiro, and Raja Giryes.
\newblock Detecting adversarial samples using influence functions and nearest
  neighbors.
\newblock In {\em Proceedings of the IEEE/CVF Conference on Computer Vision and
  Pattern Recognition}, pages 14453--14462, 2020.

\bibitem{lee2018simple}
Kimin Lee, Kibok Lee, Honglak Lee, and Jinwoo Shin.
\newblock A simple unified framework for detecting out-of-distribution samples
  and adversarial attacks.
\newblock In {\em Advances in Neural Information Processing Systems}, pages
  7167--7177, 2018.

\bibitem{freitas2020unmask}
Scott Freitas, Shang{-}Tse Chen, Zijie~J. Wang, and Duen~Horng Chau.
\newblock Unmask: Adversarial detection and defense through robust feature
  alignment.
\newblock In {\em {IEEE} International Conference on Big Data, Big Data 2020,
  Atlanta, GA, USA, December 10-13, 2020}, pages 1081--1088. {IEEE}, 2020.

\bibitem{song2017pixeldefend}
Yang Song, Taesup Kim, Sebastian Nowozin, Stefano Ermon, and Nate Kushman.
\newblock Pixeldefend: Leveraging generative models to understand and defend
  against adversarial examples.
\newblock In {\em 6th International Conference on Learning Representations,
  {ICLR} 2018, Vancouver, BC, Canada, April 30 - May 3, 2018, Conference Track
  Proceedings}. OpenReview.net, 2018.

\bibitem{hendrycks2016early}
Dan Hendrycks and Kevin Gimpel.
\newblock Early methods for detecting adversarial images.
\newblock In {\em 5th International Conference on Learning Representations,
  {ICLR} 2017, Toulon, France, April 24-26, 2017, Workshop Track Proceedings}.
  OpenReview.net, 2017.

\bibitem{zheng2018robust}
Zhihao Zheng and Pengyu Hong.
\newblock Robust detection of adversarial attacks by modeling the intrinsic
  properties of deep neural networks.
\newblock In {\em Advances in Neural Information Processing Systems}, pages
  7913--7922, 2018.

\bibitem{liang2017detecting}
Bin Liang, Hongcheng Li, Miaoqiang Su, Xirong Li, Wenchang Shi, and Xiaofeng
  Wang.
\newblock Detecting adversarial image examples in deep neural networks with
  adaptive noise reduction.
\newblock {\em {IEEE} Trans. Dependable Secur. Comput.}, 18(1):72--85, 2021.

\bibitem{carrara2017detecting}
Fabio Carrara, Fabrizio Falchi, Roberto Caldelli, Giuseppe Amato, Roberta
  Fumarola, and Rudy Becarelli.
\newblock Detecting adversarial example attacks to deep neural networks.
\newblock In {\em Proceedings of the 15th International Workshop on
  Content-Based Multimedia Indexing}, pages 1--7, 2017.

\bibitem{pang2018towards}
Tianyu Pang, Chao Du, Yinpeng Dong, and Jun Zhu.
\newblock Towards robust detection of adversarial examples.
\newblock In {\em Advances in Neural Information Processing Systems}, pages
  4579--4589, 2018.

\bibitem{sheikholeslami2019minimum}
Fatemeh Sheikholeslami, Swayambhoo Jain, and Georgios~B. Giannakis.
\newblock Minimum uncertainty based detection of adversaries in deep neural
  networks.
\newblock In {\em Information Theory and Applications Workshop, {ITA} 2020, San
  Diego, CA, USA, February 2-7, 2020}, pages 1--16. {IEEE}, 2020.

\bibitem{sotgiu2020deep}
Angelo Sotgiu, Ambra Demontis, Marco Melis, Battista Biggio, Giorgio Fumera,
  Xiaoyi Feng, and Fabio Roli.
\newblock Deep neural rejection against adversarial examples.
\newblock {\em EURASIP Journal on Information Security}, 2020:1--10, 2020.

\bibitem{aldahdooh2021selective}
Ahmed Aldahdooh, Wassim Hamidouche, and Olivier Déforges.
\newblock Revisiting model's uncertainty and confidences for adversarial
  example detection.
\newblock {\em arXiv preprint arXiv:2103.05354}, 2021.

\bibitem{carlini2017magnet}
Nicholas Carlini and David~A. Wagner.
\newblock Magnet and "efficient defenses against adversarial attacks" are not
  robust to adversarial examples.
\newblock {\em CoRR}, abs/1711.08478, 2017.

\bibitem{srivastava2014dropout}
Nitish Srivastava, Geoffrey Hinton, Alex Krizhevsky, Ilya Sutskever, and Ruslan
  Salakhutdinov.
\newblock Dropout: a simple way to prevent neural networks from overfitting.
\newblock {\em The journal of machine learning research}, 15(1):1929--1958,
  2014.

\bibitem{szegedy2015rethinking}
Christian Szegedy, Vincent Vanhoucke, Sergey Ioffe, Jonathon Shlens, and
  Zbigniew Wojna.
\newblock Rethinking the inception architecture for computer vision. 2015.
\newblock {\em arXiv preprint arXiv:1512.00567}, 2015.

\bibitem{mittal2012no}
Anish Mittal, Anush~Krishna Moorthy, and Alan~Conrad Bovik.
\newblock No-reference image quality assessment in the spatial domain.
\newblock {\em IEEE Transactions on image processing}, 21(12):4695--4708, 2012.

\bibitem{gretton2012kernel}
Arthur Gretton, Karsten~M Borgwardt, Malte~J Rasch, Bernhard Sch{\"o}lkopf, and
  Alexander Smola.
\newblock A kernel two-sample test.
\newblock {\em The Journal of Machine Learning Research}, 13(1):723--773, 2012.

\bibitem{mao2020learning}
Xiaofeng Mao, Yuefeng Chen, Yuhong Li, Yuan He, and Hui Xue.
\newblock Learning to characterize adversarial subspaces.
\newblock In {\em ICASSP 2020-2020 IEEE International Conference on Acoustics,
  Speech and Signal Processing (ICASSP)}, pages 2438--2442. IEEE, 2020.

\bibitem{vaswani2017attention}
Ashish Vaswani, Noam Shazeer, Niki Parmar, Jakob Uszkoreit, Llion Jones,
  Aidan~N. Gomez, Lukasz Kaiser, and Illia Polosukhin.
\newblock Attention is all you need.
\newblock In {\em Advances in Neural Information Processing Systems 30: Annual
  Conference on Neural Information Processing Systems 2017, December 4-9, 2017,
  Long Beach, CA, {USA}}, pages 5998--6008, 2017.

\bibitem{selective2019}
Yonatan Geifman and Ran El{-}Yaniv.
\newblock Selectivenet: {A} deep neural network with an integrated reject
  option.
\newblock {\em CoRR}, abs/1901.09192, 2019.

\bibitem{kullback1951information}
Solomon Kullback and Richard~A Leibler.
\newblock On information and sufficiency.
\newblock {\em The annals of mathematical statistics}, 22(1):79--86, 1951.

\bibitem{van2016conditional}
Aaron Van~den Oord, Nal Kalchbrenner, Lasse Espeholt, Oriol Vinyals, Alex
  Graves, et~al.
\newblock Conditional image generation with pixelcnn decoders.
\newblock {\em Advances in neural information processing systems},
  29:4790--4798, 2016.

\bibitem{huang2017densely}
Gao Huang, Zhuang Liu, Laurens Van Der~Maaten, and Kilian~Q Weinberger.
\newblock Densely connected convolutional networks.
\newblock In {\em Proceedings of the IEEE conference on computer vision and
  pattern recognition}, pages 4700--4708, 2017.

\bibitem{art2018}
Maria-Irina Nicolae, Mathieu Sinn, Minh~Ngoc Tran, Beat Buesser, Ambrish Rawat,
  Martin Wistuba, Valentina Zantedeschi, Nathalie Baracaldo, Bryant Chen, Heiko
  Ludwig, Ian~M. Molloy, and Ben Edwards.
\newblock Adversarial robustness toolbox v1.0.0, 2019.

\bibitem{he2016identity}
Kaiming He, Xiangyu Zhang, Shaoqing Ren, and Jian Sun.
\newblock Identity mappings in deep residual networks.
\newblock In {\em European conference on computer vision}, pages 630--645.
  Springer, 2016.

\bibitem{selvaraju2016grad}
Ramprasaath~R. Selvaraju, Michael Cogswell, Abhishek Das, Ramakrishna Vedantam,
  Devi Parikh, and Dhruv Batra.
\newblock Grad-cam: Visual explanations from deep networks via gradient-based
  localization.
\newblock In {\em {IEEE} International Conference on Computer Vision, {ICCV}
  2017, Venice, Italy, October 22-29, 2017}, pages 618--626. {IEEE} Computer
  Society, 2017.

\bibitem{springenberg2014striving}
Jost~Tobias Springenberg, Alexey Dosovitskiy, Thomas Brox, and Martin~A.
  Riedmiller.
\newblock Striving for simplicity: The all convolutional net.
\newblock In Yoshua Bengio and Yann LeCun, editors, {\em 3rd International
  Conference on Learning Representations, {ICLR} 2015, San Diego, CA, USA, May
  7-9, 2015, Workshop Track Proceedings}, 2015.

\bibitem{schmidt2018adversarially}
Ludwig Schmidt, Shibani Santurkar, Dimitris Tsipras, Kunal Talwar, and
  Aleksander Madry.
\newblock Adversarially robust generalization requires more data.
\newblock In {\em Advances in Neural Information Processing Systems 31: Annual
  Conference on Neural Information Processing Systems 2018, NeurIPS 2018,
  December 3-8, 2018, Montr{\'{e}}al, Canada}, pages 5019--5031, 2018.

\end{thebibliography}
}
\vfill
\pagebreak

\end{document}